\documentclass[]{bytedance_seed}

\usepackage[toc,page,header]{appendix}

\usepackage{minitoc}
\usepackage{solarized-light}
\usepackage{multirow}
\usepackage{graphicx}
\usepackage{array}
\usepackage{makecell}
\usepackage{framed}
\usepackage{hyperref}
\usepackage{amsmath} 
\usepackage[colorinlistoftodos]{todonotes}
\usepackage{longtable}
\usepackage{hhline}
\usepackage{fancyvrb}
\usepackage{fvextra}
\usepackage{CJKutf8}
\usepackage{multicol}
\usepackage{cleveref}
\usepackage{tablefootnote}
\usepackage{threeparttable}
\usepackage{tabularx}
\usepackage{mdframed}
\usepackage{subcaption}
\usepackage[usestackEOL]{stackengine}
\usepackage[numbers]{natbib}
\newcommand{\commentout}[1]{}
\renewcommand{\paragraph}[1]{\noindent\textbf{#1.}\hspace*{1em}}
\usepackage{enumitem}
\setlist[itemize]{leftmargin=15pt}

\RequirePackage{xspace}
\makeatletter
\DeclareRobustCommand\onedot{\futurelet\@let@token\@onedot}
\def\@onedot{\ifx\@let@token.\else.\null\fi\xspace}

\def\eg{\emph{e.g}\onedot} 
\def\ie{\emph{i.e}\onedot} 
 
\def\etc{\emph{etc}\onedot}

\def\etal{\emph{et al}\onedot}
\makeatother
\newcommand{\doubao}{Seed1.5-VL}
\newcommand{\vit}{{Seed-ViT}\xspace}

\title{ Seed1.5-VL Technical Report  }

\author{ByteDance Seed}

\contribution{See \hyperref[sec:contributions]{Contributions and Acknowledgments} section for a full author list.}

\abstract{
We present \doubao{}, a vision-language foundation model designed to advance general-purpose multimodal understanding and reasoning. 
\doubao{} is composed with a 532M-parameter vision encoder and a Mixture-of-Experts (MoE) LLM of 20B active parameters.
Despite its relatively compact architecture, it delivers strong performance across a wide spectrum of public VLM benchmarks and internal evaluation suites, achieving the state-of-the-art performance on 38 out of 60 public benchmarks. Moreover, in agent-centric tasks such as GUI control and gameplay, Seed1.5-VL outperforms leading multimodal systems, including OpenAI CUA and Claude 3.7. Beyond visual and video understanding, it also demonstrates strong reasoning abilities, making it particularly effective for multimodal reasoning challenges such as visual puzzles.
We believe these capabilities will empower broader applications across diverse tasks.
In this report, we mainly provide a comprehensive review of our experiences in building \doubao{} across model design, data construction, and training at various stages, hoping that this report can inspire further research.
\doubao{} is now accessible on \href{https://www.volcengine.com/}{Volcano Engine}\footnote{Model ID: doubao-1-5-thinking-vision-pro-250428}.
}

\date{\today}
\correspondence{\email{shiguang.sg@bytedance.com}}

\begin{document}
\maketitle

\newpage
\tableofcontents
\newpage

\section{Introduction}

\commentout{
This report introduces \doubao{}, our latest multimodal model developed at ByteDance Seed. We pretrain our model on trillions of multimodal tokens sourced across a variety of tasks (image, video, text, human-computer interaction, \etc) for the purpose of accumulating visual knowledge of the world and mastering basic visual capabilities, such as Optical Character Recognition (OCR) and visual grounding. In post-training, we further enhance model's reasoning capability by fine-tuning on Long Chain-of-Thought data and reinforcement learning incorporating human feedback and verifiable reward signals. 

We address the challenge of efficiently training large-scale multimodal models featuring disparate vision encoder and language model sizes. Our primary contributions include a novel hybrid parallel scheme tailored for this asymmetry and an effective vision token redistribution method for computational load balancing across GPUs. Additionally, a customized data loader is developed to reduce I/O overhead, particularly under 3D parallelism. These are combined with standard optimizations (\eg, kernel fusion, selective activation checkpointing, offloading) to maximize training throughput.

We evaluate \doubao{} on a comprehensive suite of both public and internal benchmarks, covering a wide spectrum of tasks, which includes visual reasoning, 2D grounding, counting, video understanding, computer use, etc. Specifically, we report results on a total of 51 public benchmarks, and our model achieves state-of-the-art results in 30 of them, most notably 14 out 18 in video domain and 5 out of 7 in computer use tasks. Besides driven by benchmark scores, we also build an internal chatbot and monitor model's real-world performance and its out-of-distribution (OOD) performance.

With such a strong foundation model, we are one step closer to building a generalist agent that can tackle more complex multi-step problems. ~\Cref{sec:agent_eval} demonstrates how \doubao{} is used to build smarter agents for computer use and gameplay advancing state-of-the-arts in those categories.

Despite demonstrating strong overall capabilities, \doubao{} features a relatively compact and efficient architecture. The model is composed of a vision encoder with 500 million parameters and a language model configured with 20 billion active parameters. This efficient design results in reduced inference costs and computational requirements. Such efficiency not only enhances accessibility for a broader audience through API deployment but also contributes to a more fluid and responsive user experience within interactive applications like the Doubao chatbot.

Access to the \doubao{} model will soon be available through two primary channels: interactive engagement through Doubao, our conversational AI service and programmatic access via the Volcano Engine API platform\footnote{https://www.volcengine.com/}. 
}

Vision-language models (VLMs) have emerged as a foundational paradigm for enabling general-purpose AI to perceive, reason, and act in open-ended virtual and physical environments. By aligning visual and textual modalities within a unified model, VLMs have rapidly advanced research frontiers in areas, such as multimodal reasoning~\cite{openai2023gpt4v,wang2024qwen2,team2025gemini}, image editing~\cite{gemini-exp,gpt4o-imagegen}, GUI agents~\cite{openaioperator,anthropic2025claude,qin2025ui}, autonomous driving~\cite{driveGPT4,DriveVLM,driveVLP}, and robotics~\cite{openvla,pi0.5_2025,figure2025helix}, while also powering real-world applications across education, healthcare, chatbots, and wearable devices.

However, despite substantial progress, current VLMs still fall short of human-level generality, particularly in tasks requiring 3D spatial understanding, object counting, imaginative visual inference, and interactive game play. These limitations highlight the inherent challenges in VLM development. Unlike large language models (LLMs), which benefit from abundant, high-quality textual corpora that capture a wide spectrum of human knowledge, VLMs lack access to equally rich and diverse vision-language annotations, especially for concepts grounded in low-level perceptual phenomena. Moreover, the heterogeneous nature of multimodal data introduces additional complexity in both training and inference, complicating data pipeline design, parallel training strategies, and evaluation protocols.

In this report, we share the efforts during the development of \doubao{}, our latest multimodal foundation model for vision-language understanding. To address the scarcity of high-quality annotations, we developed a suite of diversified data synthesis pipelines targeting key capabilities, including optical character recognition (OCR), visual grounding, counting, video understanding, and long-tail knowledge during pre-training, as well as visual puzzles and games during post-training. \doubao{} is pre-trained on trillions of multimodal tokens spanning diverse modalities, \ie, images, videos, text, and human-computer interaction data, to acquire broad visual knowledge and master core visual competencies. We also share the scaling behavior in the pre-training stage. In the post-training phase, we incorporate both human feedback and verifiable reward signals to further strengthen its general reasoning abilities.

We also address the challenge of efficiently training large-scale multimodal models with asymmetrical architecture, especially the imbalance between the vision encoder and the language model. Our contributions include (1) a novel \textit{hybrid parallelism} scheme optimized for this asymmetry and (2) a \textit{vision token redistribution strategy} to balance GPU workloads. In addition, we implement a customized data loader that minimizes I/O bottlenecks under 3D parallelism. These innovations, combined with standard system-level optimizations (\eg, kernel fusion, selective activation checkpointing, offloading), collectively enhance overall training throughput.

To establish a comprehensive understanding of the current landscape of VLM capabilities, thereby informing future research directions towards model improvements, we evaluate \doubao{} on an extensive suite of public and internal benchmarks, covering a wide range of tasks including visual reasoning, grounding, counting, video understanding, and computer usage. Specifically, we report results on 60 public benchmarks, where \doubao{} achieves state-of-the-art performance on 38 of them, including 21 out of 34 in vision-language benchmarks, 14 out of 19 in the video benchmarks, and 3 out of 7 in GUI agent tasks. Beyond benchmark performance, we also deploy \doubao{} within an internal chatbot system to monitor its real-world and out-of-distribution (OOD) performance in dynamic, interactive environments.

Despite its strong capabilities, \doubao{} maintains a compact and efficient architecture, featuring a 532-million-parameter vision encoder and a language model with 20 billion active parameters. This streamlined design reduces inference costs and computational demands, making the model well-suited for interactive applications. The efficiency of \doubao{} enhances accessibility for a broader user base via API services and contributes to a smoother user experience within the Doubao chatbot.
Access to \doubao{} will soon be available on the Volcano Engine API platform\footnote{\href{https://www.volcengine.com/}{https://www.volcengine.com}}.

The remainder of this report is organized as follows. We begin by presenting an overview of the model architecture and detailing the image and video encoding methods (\cref{sec:arch}). \Cref{sec:pretrain} describes the data curation strategies and the pre-training procedure, including initial findings on multimodal model scaling laws and metric prediction—a relatively underexplored area. \Cref{sec:posttrain} details the data and techniques employed during the post-training phase to enhance alignment with human preferences and improve reasoning capabilities. \Cref{sec:infra} elaborates on the necessary infrastructure innovations developed to enable scalable pre-training and post-training. Finally, \cref{sec:evaluation} presents comprehensive evaluation results on public benchmarks, showcases model capabilities via qualitative examples, discusses limitations of current multimodal models, and proposes directions for future research.
\section{Architecture}
\label{sec:arch}

\begin{figure*}[t]
    \centering
    \includegraphics[width=1.0\linewidth]{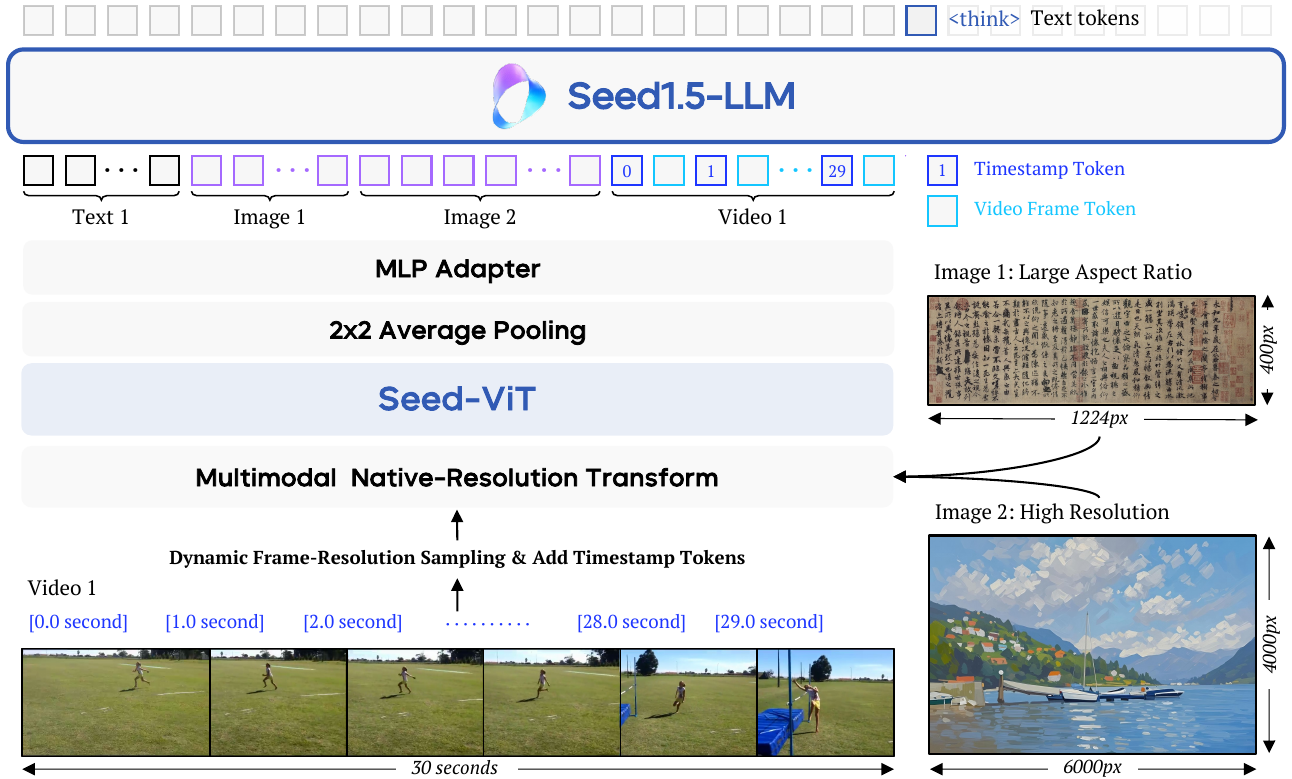}
    \caption{\textbf{The architecture of \doubao{}}. The proposed \doubao{} comprises three main components: (1) SeedViT to encode images and videos, (2) an MLP adapter to project visual features into multimodal tokens, and (3) a Large Language Model to process multimodal inputs. \doubao{} accepts images at various resolutions and processes them using a native-resolution transform to preserve maximum image detail. For video inputs, we propose the dynamic frame-resolution sampling strategy, which dynamically adjusts the sampling frame rate and resolution. Additionally, a timestamp token is added before each frame to enhance the model's temporal awareness.}
    \label{fig:model}
\end{figure*}

The architecture of \doubao{} consists of three components: a vision encoder, an MLP adapter, and a large language model (LLM). 
The vision encoder natively supports dynamic image resolutions and employs 2D RoPE~\cite{rope} for positional encoding, enabling flexible adaptation to images of arbitrary dimensions. 
To enhance computational efficiency, the architecture applies average pooling over adjacent 2×2 feature patches; a two-layer MLP subsequently processes these pooled features before being input to the LLM. 
Encoder-free architectures~\cite{fuyu8b,diao2024eve,team2024chameleon} are not considered, as the vision encoder provides efficient image compression, enabling high-resolution image representation with fewer tokens.
The overall architecture is shown in \cref{fig:model}.

\subsection{Vision Encoder}
\label{sec:ve}
Many contemporary Vision-Language Models (VLMs)~\cite{bai2025qwen2,wang2024qwen2,liu2023visual,grattafiori2024llama,team2023gemini,peng2023kosmos,abdin2024phi,hurst2024gpt,li2024llava,chen2024internvl,anthropic2025claude} commonly integrate pre-trained vision encoders designed for a fixed input resolution, typically square images. While this approach simplifies model architecture, it can inadvertently discard fine-grained visual information when processing high-resolution images, videos, or handling tasks requiring intricate detail such as OCR. Recent efforts, such as those in Qwen2-VL~\cite{wang2024qwen2} and InternVL-2.5~\cite{chen2024internvl}, have explored fine-tuning pre-trained vision encoders to accommodate dynamic-resolution inputs, offering a partial alleviation of this limitation. Nevertheless, these methods still largely depend on adapting existing fixed-resolution architectures and necessitate adjustments to position encodings (\eg, transitioning from 1D flatten position embedding to 2D RoPE~\cite{wang2024qwen2,chen2024internvl} or interpolation of 1D position embeddings to various shapes~\cite{oquab2023dinov2,tschannen2025siglip}), which may not fully retain visual details and precision post-adaptation. Furthermore, we incorporate video data into the pretraining phase to enable the model to learn not only spatial features from images but also spatial-temporal dynamics, thereby enhancing its capacity to process dynamic scenes and complex visual content.

Addressing the challenges posed by fixed-resolution processing, we developed Seed-ViT, a vision encoder specifically designed for native-resolution feature extraction. Based on the well-established Vision Transformer (ViT) architecture~\cite{dosovitskiy2020image}, Seed-ViT consists of 532 million parameters. It demonstrates strong capabilities in general visual perception across diverse domains. Notably, on zero-shot classification benchmarks, Seed-ViT attains performance comparable to models with substantially more parameters, such as InternVL-C (6 billion parameters), highlighting its efficiency. Further architectural details and our pretraining approach for Seed-ViT are provided in \cref{sec:vit_arch,sec:vit:pretrain}, respectively.

\subsubsection{Architecture}
\label{sec:vit_arch}

The architectural hyper-parameters of \vit can be found in \cref{tab:vit_arch}.
\renewcommand{\arraystretch}{1.2}
\begin{table}[ht]
    \centering
    \begin{tabular}{ccccccc}
    \toprule
         \textbf{Patch size}  & \textbf{Pos embed} & \textbf{Head dim} & \textbf{Num heads} & \textbf{Embed dim} & \textbf{MLP ratio} & \textbf{Depth}  \\
          \midrule
         14  & 2D RoPE & 64 & 20 & 1280 & 4.0 & 27 \\
    \bottomrule
    \end{tabular}
    \caption{The architectural hyperparameters of \vit.}
    \label{tab:vit_arch}
\end{table}
\renewcommand{\arraystretch}{1}

Our vision encoder is designed to accommodate input images of varying dimensions. Initially, input images undergo a pre-processing step involving bilinear interpolation to adjust their resolutions to the nearest multiple of $28\times28$ pixels. Subsequently, each image is segmented into a sequence of non-overlapping patches, each of $14\times14$ pixels. Following the approach outlined in NaViT~\cite{dehghani2023patch}, we concatenate patch sequences from multiple input images into a unified sequence. These raw patch sequences are then projected into tokens in the embedding space via a linear patch embedding layer, which are then fed into the transformer blocks. To ensure that tokens belonging to one image do not attend to tokens from other images within the batched sequence, we employ appropriate attention masks during the self-attention computations within the transformer blocks. Finally, a $2\times2$ average pooling operation is applied to the output patch embeddings before they are passed to the subsequent MLP adaptor and the LLM, as described above.

\subsubsection{ViT Pre-training Stage }
\label{sec:vit:pretrain}
\begin{table}[h]
\centering
\resizebox{0.88\linewidth}{!}{
\begin{tabular}{l|lll} \toprule
    \textbf{Categories}  & Unlabeled image & Image-text pairs & Video-audio-text tuples \\
    \midrule
    \textbf{Training samples} & 2.2B & 4.8B & 65M \\
    \textbf{Token percentages} & 4.0\% & 91.2\% & 4.8\% \\
    \textbf{Batch sizes} & 55,296 & 32,768  & 1,024 \\
    \textbf{LR warm up steps} & 1,692 & 2,000 & 12,800 \\
    \textbf{Maximum LR} & \(7.06\times 10^{-3}\) & \(1.0\times 10^{-4}\) & \(5.0\times 10^{-5}\) \\
    \textbf{Minimum LR} & \(1.05 \times 10^{-5} \) & \(1.2\times 10^{-6}\) & \(2.02 \times 10^{-7}\) \\
    \bottomrule
\end{tabular}
}
\caption{Training setup and hyperparameters used in the three ViT pre-training stages.}
\label{tab:vit_pretrain_config}
\end{table}
Our vision transformer, \vit{}, undergoes a dedicated pre-training pipeline before integration with the LLM. Guided by empirical evidence, we establish three key guidelines for our pre-training methodology:

\begin{itemize}
\item \textbf{Better Training Efficiency with ViT-pretraining.} Most successful VLMs~\cite{wang2024qwen2,bai2025qwen2,chen2024internvl} follow the setup of having a vision encoder (\eg, CLIP or SigLIP~\cite{zhai2023sigmoid}) and a few work~\cite{fuyu8b,diao2024unveiling} have attempted to remove vision encoder entirely and directly pass image patches in decoder-only LLMs but with mixed results. Beyer \etal~\cite{beyer2024paligemma} also concluded that encoder-free VLMs may be a promising future direction but still suffer in training efficiency. 
\item \textbf{Early Integration of Native-Resolution Modeling.} We prioritize the early introduction of native-resolution modeling within the pre-training pipeline. The architecture of \vit is maintained consistently throughout both the ViT pre-training and VLM stages. This ensures the prevention of performance degradation stemming from architectural modifications and eliminates the need for extensive fine-tuning to compensate for such discrepancies.
\item \textbf{Comprehensive Data Utilization.} The pre-training stage leverages the full spectrum of data intended for VLM training, encompassing unlabeled images, image-text pairs, and videos accompanied by visual and audio captions.
\end{itemize}

Based on the above guidelines, the ViT pre-training pipeline is divided into three stages: (i) Masked Image Modeling (MIM)~\cite{wei2022masked} with 2D RoPE, (ii) Native-Resolution Contrastive Learning, and (iii) Omni-modal Pre-training. Below, we provide more details of each stage.

\paragraph{MIM with 2D RoPE}
In the first stage, our goal is to enhance the visual perception ability on visual geometry and structure awareness by MIM. We leverage the EVA02-CLIP-E~\cite{fang2023eva} as the teacher model, and the student model is randomly initialized following the architecture defined in \cref{tab:vit_arch}. During training, we randomly mask out 75\% image patches and the corresponding RoPE embeddings and use the CLIP~\cite{radford2021learning} features produced by the teacher as reconstruction targets. This process is optimized by a simple cosine similarity loss between masked-out patches in the student's and teacher's outputs. We find that the discrepancy in visual position embeddings between student and teacher models does not harm the performance, as the teacher employs learnable positional embeddings while the student uses 2D RoPE. Instead, 2D RoPE empowers the student with robust native dynamic-resolution recognition. As we scale up this MIM process, the abilities of VLMs on chart/document understanding and OCR are significantly improved.

\paragraph{Native-Resolution Contrastive Learning} In the contrastive learning stage, the vision encoder is initialized with our MIM-trained student model, while the text encoder is initialized using the text encoder from EVA-02-CLIP-E. For each given image-text pair, we aggregate the extracted patch features from the vision encoder into a single 1280-dimensional image embedding using attention pooling. Alignment between the image and text embeddings is then achieved by jointly optimizing the SigLIP loss~\cite{zhai2023sigmoid} and the SuperClass loss~\cite{huang2024classification}.

\paragraph{Omni-modal Pre-training} 
This stage adopts the MiCo framework~\cite{zhang2024explore}, constructing aligned tuples consisting of video frames, audio, visual captions, and audio captions from video data. The ViT encodes both video frames and audio, while a separate text encoder processes captions. Through alignment of these embeddings, the ViT learns unified omni-modal representations. 
Despite consuming only 4.8\% of the token budget allocated for the entire ViT pre-training process, this stage significantly enhances the ViT's performance on image and video understanding tasks.

\Cref{tab:vit_pretrain_config} summarizes the training setup and hyperparameters used in each stage.

\subsection{Video Encoding}
Effectively encoding video, beyond static image representation, remains a core challenge. A model's ability to interpret temporal sequences, adapt to varying frame rates, and perceive absolute time is critical for understanding dynamic visual content. \doubao{} addresses these challenges by introducing \textbf{Dynamic Frame-Resolution Sampling}, a novel strategy that jointly optimizes sampling across both the temporal (frame) and spatial (resolution) dimensions to balance semantic richness and computational efficiency.

Under this Dynamic Frame-Resolution Sampling strategy, videos are processed as sequences of image frames. The temporal dimension is managed through dynamic frame sampling. Instead of a uniform rate, \doubao{} adjusts the frame sampling frequency based on content complexity and task requirements. The default sampling rate is set at 1 frame per second (FPS), suitable for capturing a general understanding of video content. For tasks~\cite{li2024mvbench,wang2024tarsierrecipestrainingevaluating} requiring detailed temporal information, the frame sampling rate is increased to 2 FPS. For tasks such as video counting~\cite{dwibedi2020counting} or motion tracking~\cite{hong2025motionbench}, the rate is increased to 5 FPS. To explicitly ground each frame within the video's timeline, we prepend timestamp tokens (\ie, \texttt{[1.5 second]})  to each frame. This explicit timing annotation substantially enhances the model’s temporal awareness and enables it to handle variable frame rates common in real-world scenarios effectively.

Considering computational constraints inherent in processing long video sequences, the spatial dimension of the sampling is governed by dynamically adjusting the resolution allocated to each selected frame, managed within a maximum budget of 81,920 tokens per video. The model dynamically adjusts spatial resolutions, assigning tokens per frame through a hierarchical allocation system offering six predefined levels: \{640, 512, 384, 256, 160, 128\}. This allows for a flexible trade-off, \ie, using higher resolution for fewer frames or lower resolution to accommodate more frames from longer videos. In cases where a video is exceptionally long and exceeds the maximum encoding length even when using the lowest token allocation (128 tokens per frame), a fallback mechanism is triggered. The model then reduces the total frame count through uniform sampling across the video. While this reduces temporal density, it ensures that the entire video is represented, balancing processing efficiency with the preservation of significant temporal information.

This flexible strategy allows \doubao{} to efficiently and accurately process varying video lengths and frame rates, maintaining essential temporal details crucial for diverse video understanding tasks.
\section{Pre-training}
\label{sec:pretrain}
This section describes the data curation process (\cref{sec:pretrain_data}) and training recipe (\cref{sec:pretrain_recipe}) used in the pre-training stage of \doubao{}. In \cref{sec:scaling_laws}, we present the scaling behavior of our model.

\subsection{Pre-training Data}
\label{sec:pretrain_data}
The \doubao{} pre-training corpus contains 3 trillion diverse, high-quality source tokens. This data is categorized based on target capabilities, with the curation process for each category detailed in the following subsections.
\subsubsection{Generic Image-Text Pairs \& Knowledge Data}
Web-sourced image-text pair data, including alt text, image captions, and surrounding text, is available at an unprecedented scale (billions of instances) and exhibits high diversity in both visual and textual concepts. However, this data is inherently noisy (\eg, irrelevant or inaccurate text) and often exhibits class imbalance.

To mitigate these challenges, we first employ a series of filtering techniques, including image-text similarity scoring (\eg, CLIP-score thresholding), image-based criteria (\eg, removal of undersized images or those with extreme aspect ratios), text-based criteria (\eg, filtering of excessively short or long text), deduplication strategies (\eg, exact and near-duplicate image removal), and URL/domain-based filtering.

Furthermore, the distribution of visual concepts within the raw image-text pairs adheres to a long-tail pattern. To empirically test this observation, we conduct a sandbox experiment using Biotrove \cite{yang2024biotrove}, a large-scale dataset for species classification containing 161.9 million images spanning 366,600 species. We train a 1.1 billion-active-parameter variant of our VLM using three distinct data distributions:
\begin{itemize}
    \item \textbf{Random-46M.} 46 million samples randomly selected from the training set.

    \item \textbf{Max1k-46M.} 46 million samples selected with a maximum of 1,000 samples per species, ensuring inclusion of rare species.

    \item \textbf{Max100-15M.} 15 million samples with a maximum of 100 samples per species, providing greater relative exposure to rare species.
\end{itemize}
We evaluate the models on two specially filtered test sets derived from the original dataset: Balanced10k (sampled from BioTrove-Balanced representing common species) and Rare2k (sampled from BioTrove-Unseen representing rare species). Our experiment shown in \cref{tab:distribution_experiment} indicates that the Random-46M configuration performs poorly on rare species recognition. In contrast, limiting the maximum samples per common species (Max1k-46M) significantly improves performance on rare species. Further restricting common species' representation (Max100-15M) enhances memorization of rare species but adversely affects common species recognition. Thus, effectively capturing visual knowledge requires maintaining diverse examples of common visual concepts while ensuring sufficient training iterations for rare visual concepts.

\begin{table}[ht]
\centering
\begin{tabular}{lcccc}
\toprule
 & Training tokens & Balanced10k & Rare2k & Average \\
\midrule
Random-46M (1 epoch)               & 12B & 78.92 & 10.46 & 44.69 \\
Max1k-46M (1 epoch)       & 12B & {\bf 79.17}          & 44.85 & 62.01 \\
Max100-15M (3 epochs)      & 12B & 60.31          & {\bf 89.41} & 74.86 \\
\bottomrule
\end{tabular}
\caption{Performance comparison on Balanced10k and Rare2k under three training data distributions, Random-46M, Max1k-46M, and Max100-15M. Evaluation was conducted using an open-ended Question Answering (QA) task, with responses automatically scored by a LLM judge. All models were trained with a fixed budget of 12 billion tokens.}
\label{tab:distribution_experiment}
\end{table}

To address the imbalance between common and rare visual knowledge acquisition from image-alt-text pairs, we propose a targeted pre-processing framework. Initially, this framework utilizes a precursor version of our VLM to automatically annotate the data with pertinent semantic domains (\eg, landmarks, food, commodities, biology) and associated named entities (\eg, product brands, species names). Named entities exhibiting low corpus frequency are identified as instances of rare visual knowledge. To mitigate data sparsity, we identify domains whose representation constitutes less than 50\% of the average domain frequency. Alt-texts corresponding to these underrepresented domains are subsequently duplicated. By merging this augmented subset, enriched with samples from less frequent domains, back into the original corpus, we achieve a more balanced distribution of visual concepts. This re-balancing is designed to enhance the visual knowledge learning component, crucial to our pre-training methodology.

\subsubsection{Optical Character Recognition (OCR)}

\begin{figure}[h]
\centering
\begin{subfigure}[t]{0.43\linewidth}
\centering
\includegraphics[height=4cm,keepaspectratio]{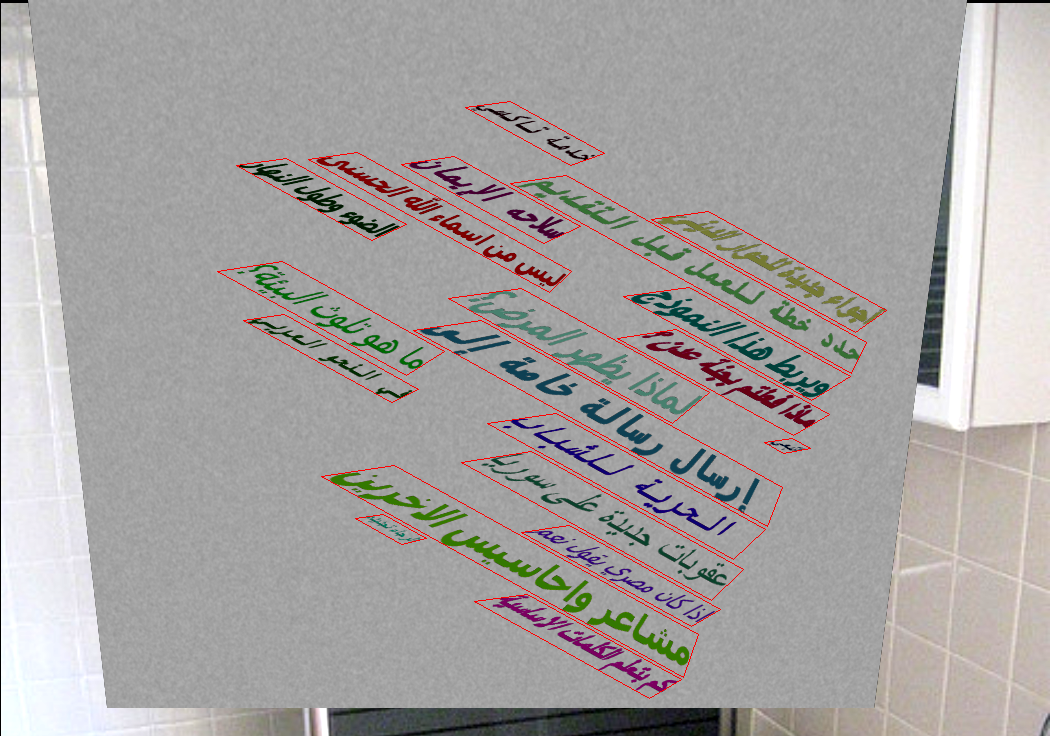}
\caption{}
\end{subfigure}
\begin{subfigure}[t]{0.43\linewidth}
\centering
\includegraphics[height=4cm,keepaspectratio]{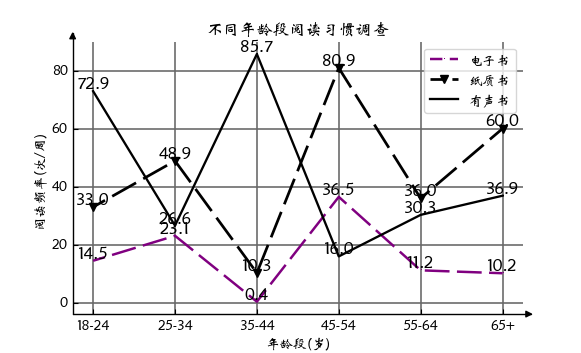}
\caption{}
\end{subfigure}
\begin{subfigure}[t]{0.2\linewidth}
\centering
\includegraphics[height=4cm,keepaspectratio]{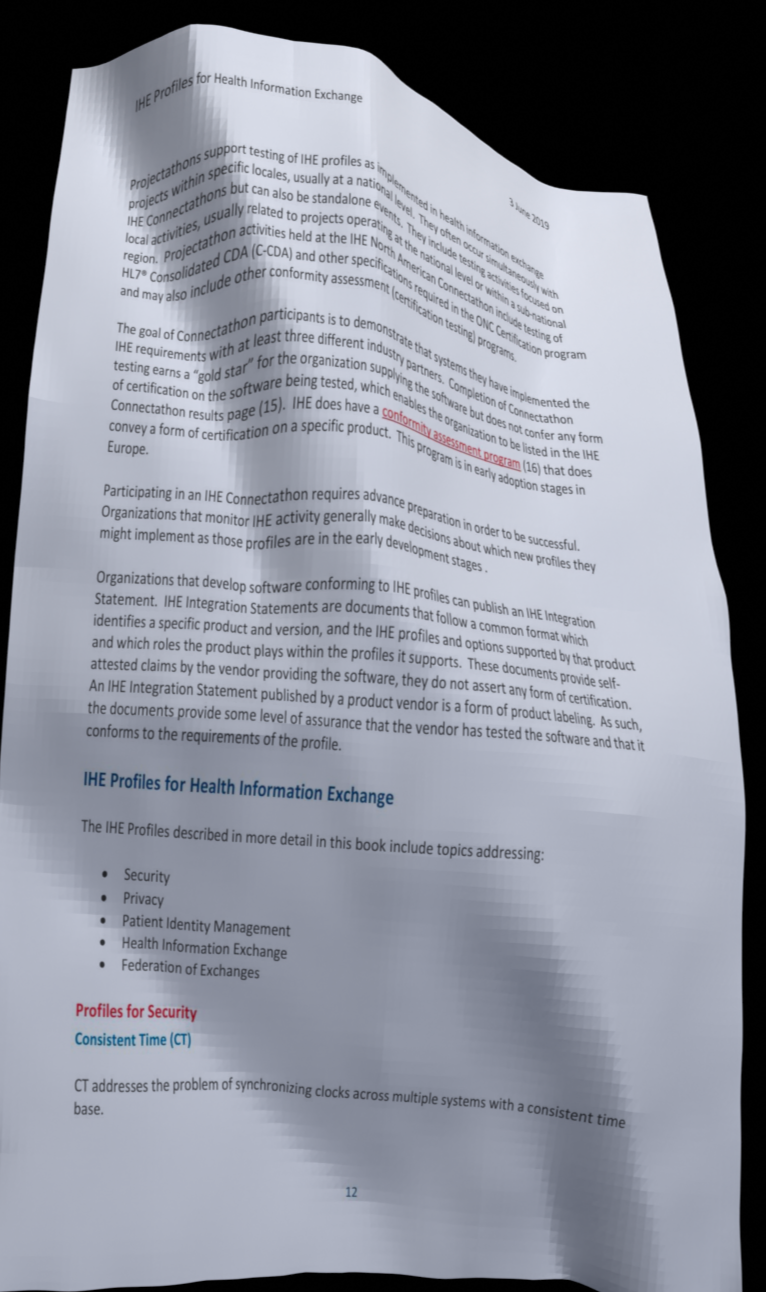}
\caption{}
\end{subfigure}
\begin{subfigure}[t]{0.66\linewidth}
\centering
\includegraphics[height=4cm,keepaspectratio]{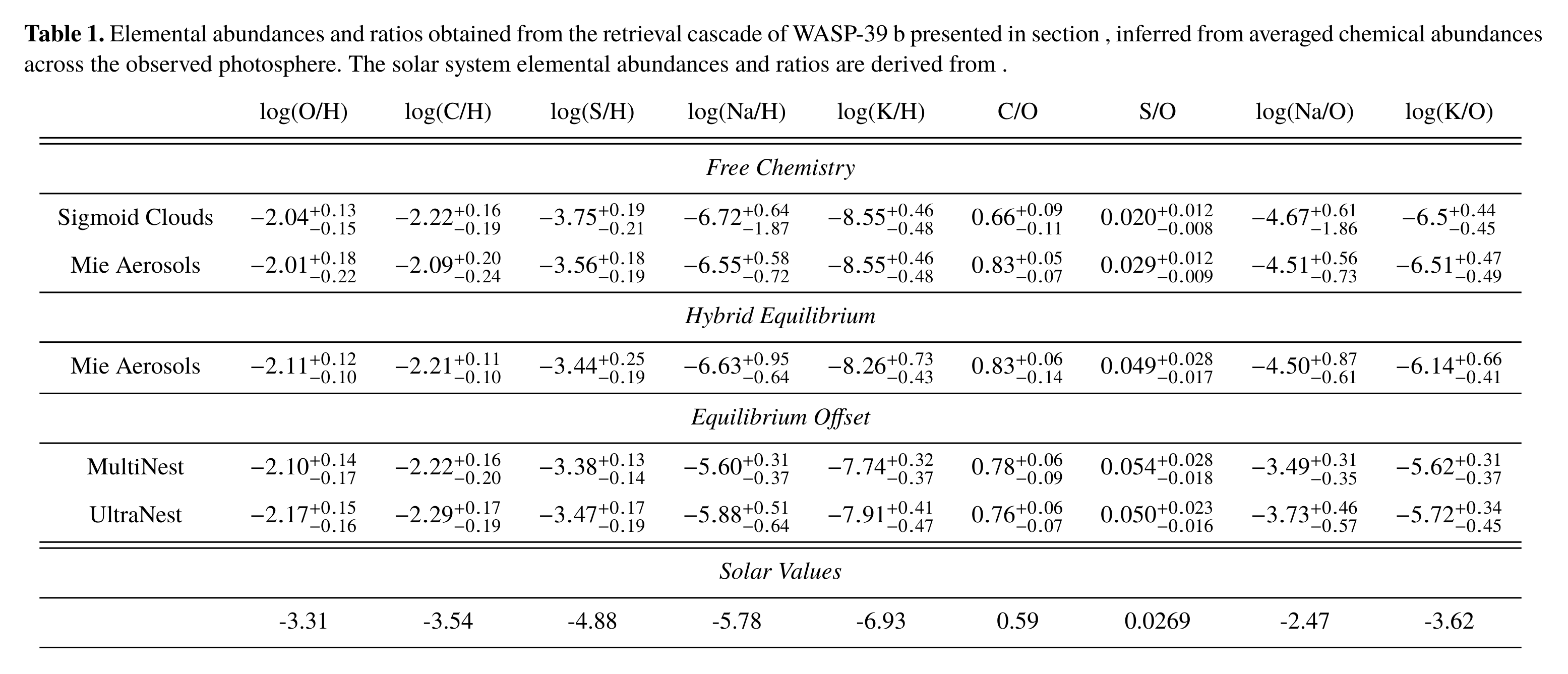}
\caption{}
\end{subfigure}
\caption{(a) An image generated by SynthDog and the corresponding textual annotations are organized in the following format: \texttt{<text>...</text><polygon>...</polygon>}; (b) The synthesized chart data includes two types of annotations: chart-to-text parsing and QA pairs; (c) The original document image undergoes transformations to simulate real-world distortions, such as perspective shifts, bends, and wrinkles. These augmentations enhance the model's robustness and improve its ability to recognize texts under diverse and challenging conditions; (d) An example of a QA pair generated for the above synthesized table image: {\it Question: What is the value of log(C/H) for Sigmoid Clouds? Give analytical steps. Answer: We look for the row labeled ``Sigmoid Clouds'' and the column labeled ``log(C/H)''. The value in that cell is $-2.22^{+0.16}_{-0.19}$}.}
\label{fig:ocr_data}
\end{figure}

To enhance the Optical Character Recognition (OCR) capabilities of the VLM, particularly for multilingual text, special symbols, and the analysis of structurally complex documents, as shown in \cref{fig:ocr_data}, we adopt large volumes of both annotated and synthetic data to train \doubao{}.

We build an in-house OCR training dataset containing over 1 billion samples, covering documents, scene text, tables, charts, and flowcharts. For document data, we collected a large volume of pages from various sources and applied our internal tools to extract content and layout information. Furthermore, we curated a diverse set of fonts, including artistic, handwritten, and non-Latin scripts, and subsequently synthesized over 200 million text-intensive images utilizing tools such as SynthDog~\cite{kim2022donut} and LaTeX (see \cref{fig:ocr_data}(a) for an example). To improve the model's robustness in understanding textual content within images, we apply various data augmentation techniques to the synthetic data, including blurring, the addition of $\text{moir}\acute{\text{e}}$ patterns, and image distortion. \Cref{fig:ocr_data}(c) illustrates an example of a document image after applying distortion-based augmentation.

Our chart dataset combines existing open-source datasets (e.g., FigureQA~\cite{kahou2017figureqa}) with newly generated synthetic data.
Synthetic charts were generated using both conventional tools (ECharts~\cite{li2018echarts}, Matplotlib~\cite{Hunter2007matplotlib}) and a novel LLM-based pipeline. In our pipeline, an LLM generates textual chart components (titles, legends, etc.), which are then transformed by an LLM into LaTeX or Python code for rendering (\cref{fig:ocr_data}(b)). Chart images were obtained via execution of this code. This multi-pronged approach resulted in a large-scale dataset exceeding 100 million chart examples.

For table data, we extract text in HTML, LaTeX, and Markdown formats from various sources, including web page HTML, GitHub README files, and LaTeX files from arXiv. Using this text, we render over 50 million table images, creating a comprehensive dataset for table parsing. This dataset enables our model to efficiently convert tables into formats such as HTML, LaTeX, and Markdown.

To further enhance the model's comprehension of textual content within images, we constructed a visual question answering (VQA) dataset to complement the structured image-text representations. Specifically, we employed a previous version of our VLM to generate question-answer pairs by conditioning on OCR outputs, chart content, table text, and the images themselves, utilizing a few-shot prompting approach. \Cref{fig:ocr_data}(d) gives an example of an input table image and the corresponding generated QA pair. Subsequently, we applied an internal LLM to filter the generated question-answer pairs, removing instances exhibiting low semantic relevance between the question and the answer. Our experiments indicate that the inclusion of this VQA dataset significantly improved the model's ability to understand textual information present in images.

\subsubsection{Visual Grounding \& Counting}
Object grounding, a fundamental capability for multimodal models, involves interpreting user instructions to identify and locate specific object regions within images. In this work, we employ two primary grounding representations for \doubao{}: bounding boxes and center points. Building upon this localization foundation, we extend \doubao{}'s capabilities to include object counting. Accordingly, our training strategy primarily utilizes three data types: bounding box annotations, point annotations, and counting data.

\paragraph{Bounding Box Data} Firstly, we adopt widely-used open-source datasets for generic object grounding, including Objects365~\cite{object365}, OpenImages~\cite{OpenImages}, and RefCOCO/+/g~\cite{refcoco,referit,refcocog}. Rather than directly incorporating those datasets for training, we filter low-quality samples of the open-source datasets and construct diverse grounding tasks. Specifically, we render all object bounding boxes for each category onto the images and adopt the previous version of our VLM to perform data inspection, which allows us to filter out samples with incorrect annotations, missing labels, or redundant annotations. Furthermore, we use these open-source datasets to construct diverse multi-task training data, including: (1) generic 2D grounding, (2) question answering about spatial relationships, and (3) question answering with visual prompts, which results in about 48 million samples and 41 billion tokens. Considering the limitations in the diversity of open-source grounding datasets in terms of both data domains and categories, we develop an efficient automatic annotation pipeline for generic multi-object grounding with large-scale image-text pairs. Specifically, we follow previous work~\cite{Cheng_2024_yolo} and extract noun phrases and entities from captions, and then adopt Grounding DINO~\cite{liu2024grounding,mmdetection} to annotate diverse open-vocabulary objects in web images.
We filter out low-quality annotations with CLIP~\cite{Radford2021LearningTV} and heuristic metrics, \eg, non-maximum suppression. The automatic annotation pipeline brings about 200 million samples and 200 billion tokens.

\paragraph{Point Data} Initially, we utilized the public data provided by PixMo-Points~\cite{deitke2024molmo}. Recognizing limitations in the diversity and quantity of the available PixMo data, we developed a dedicated pipeline for generating additional pointing data. This pipeline employs Molmo~\cite{deitke2024molmo} and CountGD~\cite{amini2024countgd} to annotate the center points of objects within a large collection of web images. Notably, CountGD proved particularly effective in annotating objects in dense image scenarios. Following annotation, low-quality data samples were filtered out, resulting in a final dataset comprising approximately 170 million instructions and 110 billion tokens.

\paragraph{Counting Data} We further sample from the aforementioned bounding box and point data to construct a counting dataset, containing approximately 8 million samples and 13 billion tokens. Specifically, we developed two variants: box-based counting and point-based counting, following a two-stage pipeline of 1) detection or pointing, then 2) generating counting results based on the numbers of the bounding boxes or points.

During training, we employ relative coordinates and normalize all coordinate values such that the output bounding boxes and points fall within the range \texttt{[0, 999]}, which enables \doubao{} to accurately predict corresponding bounding boxes and points irrespective of the input image resolution. We apply this normalization strategy to all data related, including Optical Character Recognition (OCR) and Graphical User Interfaces (GUI).

\subsubsection{3D Spatial Understanding} 
To enable the model's 3D spatial understanding ability from a single image, we construct data targeting the following three tasks: relative depth sorting, absolute depth estimation, and 3D grounding. To generate the \textbf{relative depth sorting} data, we employed DepthAnything V2~\cite{depth_anything_v2} to infer depth relationships among objects sampled from 2 million internet images. This process yielded a dataset component comprising 3.2 billion tokens associated with this task. In particular, we select the average depth of objects with a relative depth gap beyond 20\%. 

Data for \textbf{absolute depth estimation} was derived from publicly available datasets. For each entity identified by its semantic mask, we determined its absolute depth using the corresponding annotated depth map. This procedure resulted in 18 million instruction pairs (\eg, query/depth value) and contributed 28 billion tokens to our pre-training corpus. 

For \textbf{3D grounding} data, we utilized publicly available datasets from the internet. These datasets were then processed and reformulated into question-answering (QA) pairs. Specifically, our reformulation involved prompting for the 3D locations of objects belonging to a particular category. This process yielded a dataset of 770K instruction-following pairs, comprising 1.3 billion tokens.

\subsubsection{Video}
This part of data is used to improve the model's understanding of multi-frame time-series images in video. It comprises three primary categories. 
Firstly, general video understanding data, this portion encompasses a variety of tasks, including video captioning, video question answering, action recognition, action grounding, and multi-image understanding. Data are sourced from public datasets and internally collected video-caption pairs.
Secondly, we include several publicly available datasets for video temporal grounding and moment retrieval to enhance the model's temporal awareness. 
Specifically, \doubao{} directly predicts the start and end timestamps based on user prompts, with the default seconds format. Temporal grounding capability benefits complex reasoning tasks in videos.
Lastly, video streaming data is crucial for understanding dynamic and continuous video content. The data is drawn from various sources and structured into three main components:
\begin{itemize}
    \item \textbf{Interleaved Caption/QA Data.} First, we construct interleaved video text sequences either by directly captioning segmented video clips or by constructing multi-turn question-answer pairs in chronological order. These captions and QA pairs are inserted at the corresponding timestamps within the video to enhance real-time video understanding.
    \item \textbf{Proactive Reasoning Data.} Second, we reconstruct grounded video question answering and dense caption data into a frame-by-frame response format. This data requires the model to continuously monitor the video stream and proactively determine the appropriate timestamps to produce responses.
    \item \textbf{Realtime Commentary Data.} Third, we leverage naturally temporally synchronized video commentary data to provide fine-grained interleaving and alignment of video frames and texts. This formation enables the model to handle interruptions and dynamically update responses in real-time according to the video stream. 
\end{itemize}

Together, these datasets form a comprehensive foundation for effective video training.

\subsubsection{Science, Technology, Engineering, and Mathematics (STEM)} 
To enhance the model's reasoning capabilities during pre-training, we incorporated a diverse collection of problem-solving data across various STEM domains, obtained through both crawling and manual annotation. This effort culminated in the creation of comprehensive STEM datasets, structured around two primary components: \textbf{image comprehension data} and \textbf{problem-solving data}.

The \textbf{image comprehension data} comprises several subsets. We collected 3.2 million high-quality educational grounding samples across 300 categories within mathematics, physics, chemistry, and biology. Additionally, we synthesized 10 million structured tables with diverse formats, generated 4.5 million chemical structural diagrams, and produced 1.5 million synthetic coordinate system diagrams, including function plots and positional graphs. A specific subset, K12 Caption data, includes 100,000 human-annotated captions for educational images, 1 million visual question-answering (VQA) pairs, 1 million machine-generated captions using an automated pipeline, and hundreds of thousands of geometry-specific captions.

For the \textbf{problem-solving data} component, we processed over 100 million K12-level exercises through a rigorous cleaning and reformulation process. This was complemented by tens of millions of curated Chinese adult education problems and several million English-language image-associated questions.

The construction of these datasets employed hybrid acquisition strategies, integrating manual annotation, automated synthesis, and stringent quality control measures. This approach ensures multimodal coverage encompassing textual, visual, and diagrammatic representations across core STEM domains such as mathematics, physics, and chemistry.

\subsubsection{Graphical User Interface (GUI)}
For GUI data, we mainly include data curated from UI-TARS~\cite{qin2025ui,seed2025uitars15}. Specifically, to support robust GUI perception, grounding, and reasoning, we curated a large-scale dataset across web, app, and desktop environments. Each screenshot is paired with structured metadata---element type, bounding box, text, and depth---collected via automated parsing and human-assisted exploration. For \textbf{perception}, we constructed tasks including element description, dense captioning, and state transition captioning. These tasks teach the model to identify small UI components, understand overall layouts, and detect subtle visual changes across frames. Visual markers (Set-of-Mark) are also overlaid to strengthen spatial correspondence. For \textbf{grounding}, we train the model to predict element coordinates from textual descriptions. Bounding boxes are normalized across resolutions. For \textbf{reasoning}, we collect multi-step task trajectories, each annotated with observations, intermediate thoughts, and actions. This data, combining in-house and standardized open-source traces, enables the model to learn step-by-step planning, correction, and reflection. 

\subsection{Training Recipe}
\label{sec:pretrain_recipe}
\begin{table}
\centering
\begin{tabular}{>{\bfseries}l|lll} \toprule
    Stages & Stage 0 & Stage 1 & Stage 2 \\
    \midrule
    Training budget (tokens) & 16B & 3T & 240B \\
    Sequence length & 32,768 & 32,768 & 131,072 \\
    Trainable components & MLP adaptor & all & all \\
    Batch sizes (tokens) & 8.4M & 71M & 71M \\
    LR warmup steps & 100 & 500 & 0 \\
    Maximum LR & $2.52\times 10^{-4}$ & $5.22\times 10^{-5}$ & $5.22\times 10^{-6}$ \\
    Minimum LR & $4.50\times 10^{-5}$ & $5.22\times 10^{-6}$ & $5.22\times 10^{-6}$ \\
    \bottomrule
\end{tabular}
\caption{Training setup and hyperparameters in three pre-training stages.}
\label{tab:pretrain_stages}
\end{table}

Large multimodal models are typically trained either through joint multimodal learning from the start~\cite{hurst2024gpt,team2023gemini}, or via post-hoc adaptation after language model pre-training~\cite{wang2024qwen2,chen2024internvl}. Seed1.5-VL currently adopts the latter for flexible ablation and fast iterative development.

As delineated in \cref{sec:arch}, our proposed model comprises three primary modules: a vision encoder, an MLP adapter, and a language model. Prior to the VLM pre-training phase, the vision encoder undergoes an independent training procedure as detailed in \cref{sec:ve}. The language model is initialized from an internal pre-trained model with approximately 20 billion active parameters. This language model employs a decoder-only Mixture-of-Experts (MoE) architecture~\cite{shazeer2017sparsely} and has been trained on a large-scale corpus consisting of trillions of high-quality text-only tokens. Our VLM pre-training methodology is structured into three distinct stages, as summarized in \cref{tab:pretrain_stages}:
\begin{enumerate}
    \item In stage 0, we align the vision encoder with the language model by only training the MLP adapter while keeping the vision encoder and the language model frozen. Omitting this stage yields a slightly higher loss and worse performance. 
    \item In stage 1, all model parameters are trainable. This stage focuses on knowledge accumulation and mastering visual grounding and OCR capabilities of the model by training on a multimodal corpus of 3 trillion tokens, mainly composed of captions, interleaved image-text, visual grounding, and OCR data. Empirically, we found that adding a small amount of text-only tokens (\eg, 5\%) can maintain the model's language-only capabilities. Also, adding a small amount of instruction following data results in more reliable evaluation results, which allows us to decouple pre-training development from post-training's.
    \item In stage 2, we create a more balanced data mixture across different tasks, as well as adding data from new domains, such as video understanding, coding, and 3D spatial understanding.  In addition, we increase the sequence length from 32,768 to 131,072, which better accommodates modeling long dependencies in videos and complex reasoning problems. Same as in stage 1, all model parameters are trainable.
\end{enumerate}
We also experimented with an alternative training strategy, similar to approaches employed by~\cite{wang2024qwen2,chen2024internvl}, where in stage-0 both the MLP adaptor and the vision encoder are trained while the language model remains frozen. Empirical evaluation, however, demonstrated that our training recipe yields superior performance. We hypothesize that this difference may stem from the vision encoder attempting to compensate for potential inabilities within the frozen LLM, which could consequently compromise its perceptual capabilities.

We employ the AdamW optimizer~\cite{kingma2014adam} in all three stages' training with $\beta_1=0.9$, $\beta_2=0.95$, and a weight decay of 0.1. The bias and normalization parameters are omitted from the weight decay, and other training hyperparameters can be found in \cref{tab:pretrain_stages}. Stage-0 and stage-1 training follow a full cosine decay learning rate schedule, while the starting learning rate in stage 2 is equal to the ending learning rate from stage 1 and is kept constant throughout the training. In stage 2, we load the optimizer states from stage 1, so no learning rate warmup is used.

\subsection{Scaling Laws}
\label{sec:scaling_laws}

\begin{figure}[h]
  \centering
  \centering
  \begin{subfigure}[b]{0.32\textwidth}
    \includegraphics[width=\textwidth]{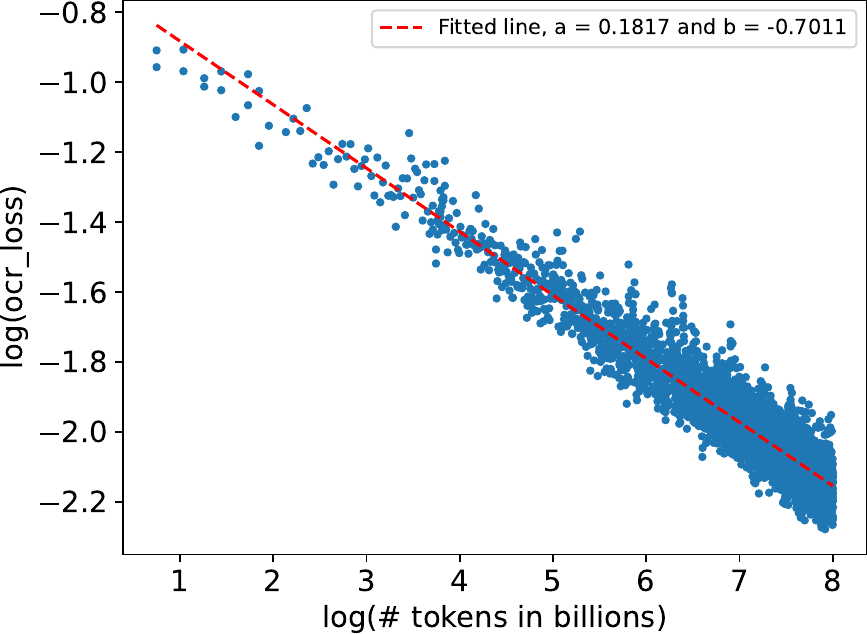}
    \caption{}
  \end{subfigure}
  \hfill
  \begin{subfigure}[b]{0.32\textwidth}
    \includegraphics[width=\textwidth]{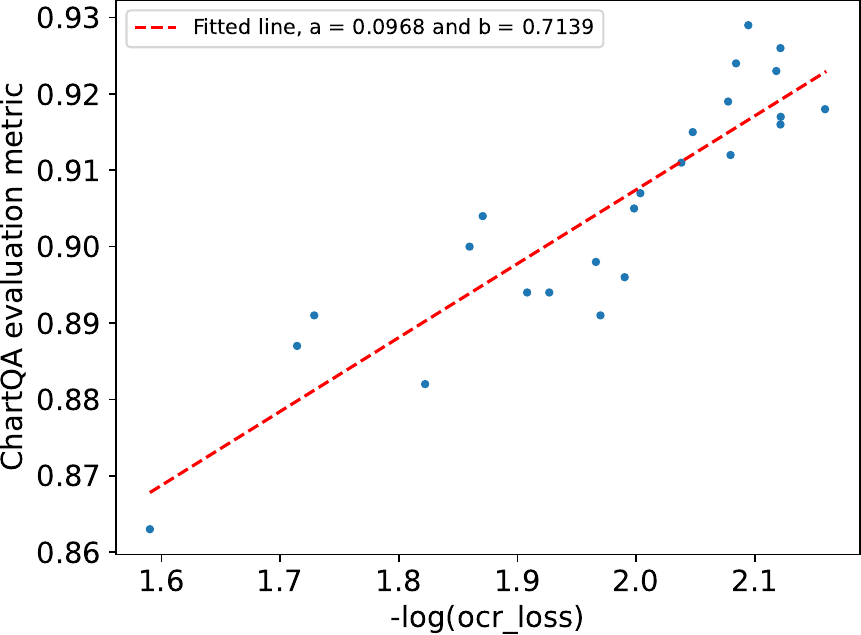}
    \caption{}
  \end{subfigure}
  \hfill
  \begin{subfigure}[b]{0.32\textwidth}
    \includegraphics[width=\textwidth]{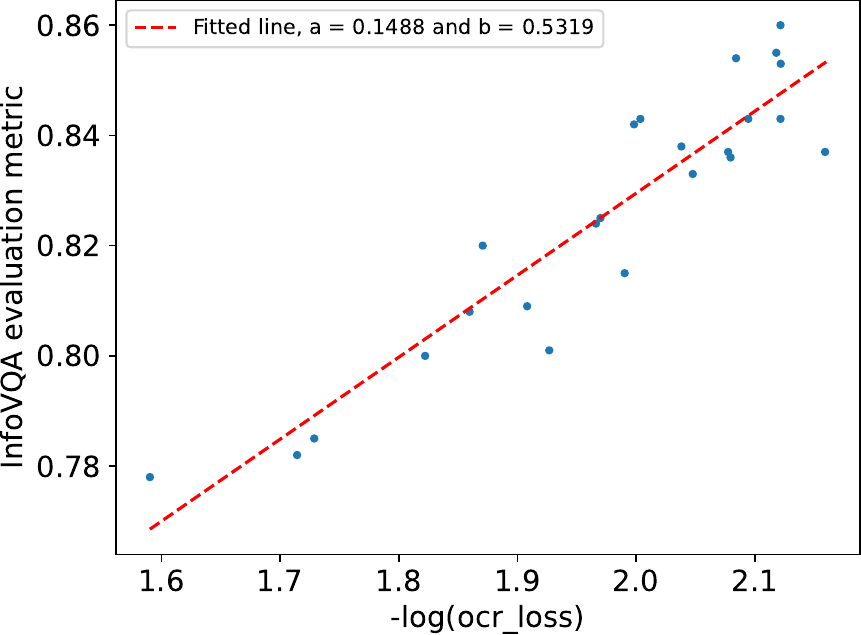}
    \caption{}
  \end{subfigure}
  \begin{subfigure}[b]{0.32\textwidth}
    \includegraphics[width=\textwidth]{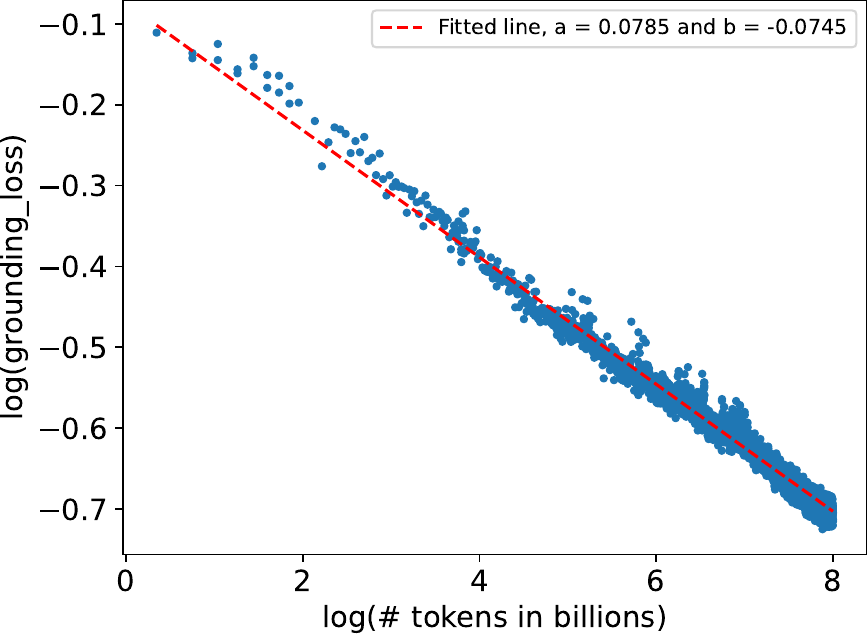}
    \caption{}
  \end{subfigure}
  \hfill
  \begin{subfigure}[b]{0.32\textwidth}
    \includegraphics[width=\textwidth]{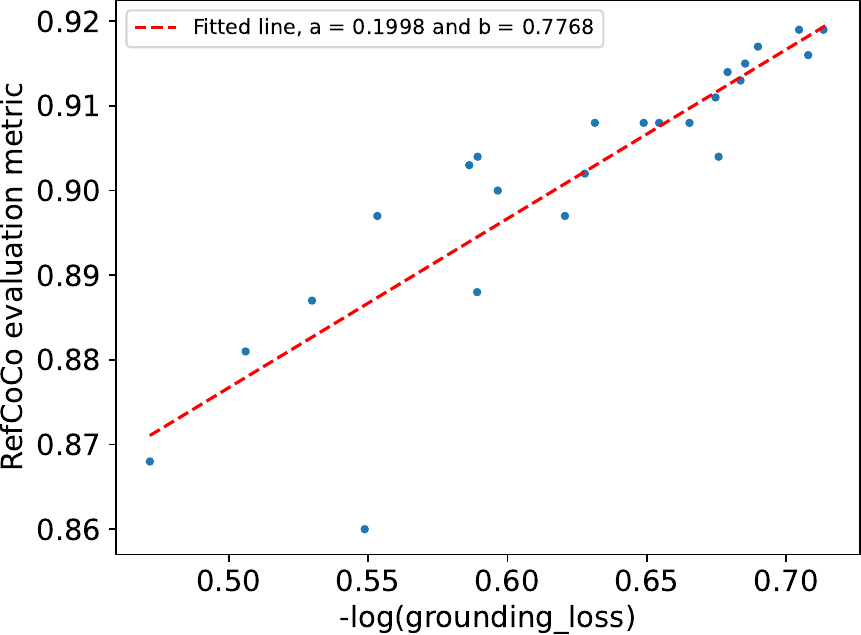}
    \caption{}
  \end{subfigure}
  \hfill
  \begin{subfigure}[b]{0.32\textwidth}
    \includegraphics[width=\textwidth]{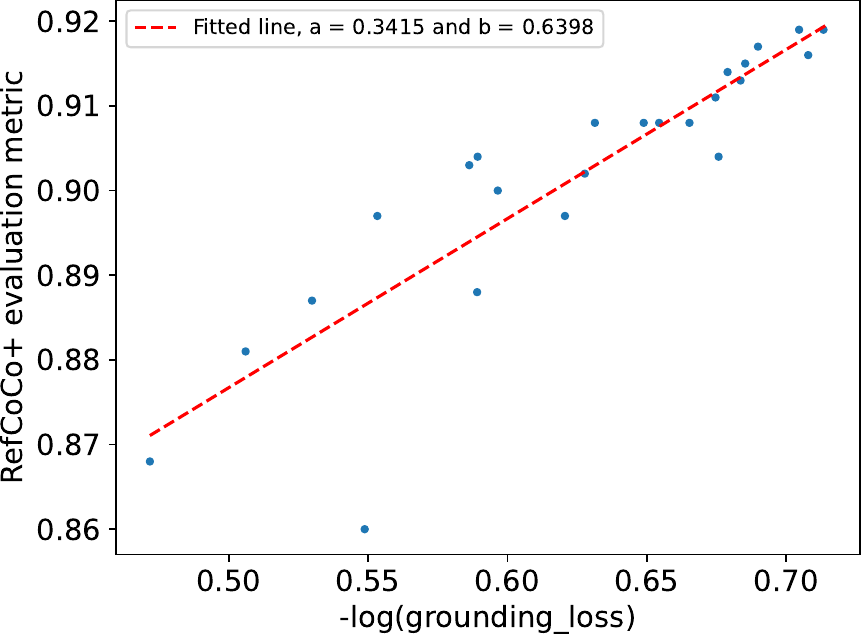}
    \caption{}
  \end{subfigure}
\caption{The relationship between the training loss of most sub-categories and training tokens obeys the power law~\cite{hoffmann2022training}. Also, the relationship between the training loss of a sub-category and the corresponding downstream evaluation metric appears to be log-linear (\eg, $\text{metric} \sim \log(\text{loss})$) within a local neighborhood. (a) The training loss of OCR related dataset as a function of training tokens; (b) Top-1 accuracy on ChartQA~\cite{masry2022chartqa} as a function of the training loss; (c) Top-1 accuracy on InfographicVQA~\cite{mathew2022infographicvqa} as a function of the training loss; (d) The training loss of grounding related dataset as a function of training tokens; (e) Precision@IoU=0.5 on RefCOCO~\cite{refcoco,referit} as a function of the training loss; (e) Precision@IoU=0.5 on RefCOCO+~\cite{refcoco,referit} as a function of the training loss. Note that the evaluation metrics displayed in this figure represent performance after pre-training and are therefore not directly comparable to the final results, which are achieved following reinforcement learning (RL) as detailed in Section~\ref{sec:evaluation}.}
\label{tab:scaling_laws}
\end{figure}

The pre-training of Vision-Language Models (VLMs) like \doubao{} differs fundamentally from the standard practice for Large Language Models (LLMs), which typically involves random initialization of all model parameters. In contrast, \doubao{} is built upon pre-trained components, including a vision encoder, an MLP adaptor, and a language model. This section focuses on understanding the scaling behavior of \doubao{} during the stage-1 phase of pre-training. Based on prior work on LLM scaling laws~\cite{henighan2020scaling,kaplan2020scaling,hoffmann2022training}, the average negative log-likelihood loss 
$L$ is modelled as a function of model parameters $N$ and training tokens $D$:
\begin{equation}
\hat{L} \sim \frac{A}{N^\alpha}+\frac{B}{D^\beta}.
\label{eq:llm_scaling_law}
\end{equation}
Given that our model architecture and thus the number of parameters are fixed during this stage, \cref{eq:llm_scaling_law} simplifies to a dependency primarily on the scale of the training data:
\begin{equation}
\hat{L} \sim \frac{B}{D^\beta}.
\end{equation}
To facilitate analysis, we examine this relationship in log-log space by taking the logarithm of both sides:
\begin{equation}
\log(\hat{L}) \sim \log(B)-\beta\log(D)=-a\log(D)+b.
\label{eq:loss_vs_tokens}
\end{equation}

We organized our pre-training dataset into distinct categories corresponding to specific capabilities (as detailed in \cref{sec:pretrain_data}). We observed that the training loss for the majority of these data sub-categories exhibits a clear adherence to the scaling relationship defined by \cref{eq:loss_vs_tokens}. As shown in \cref{tab:scaling_laws} (a) and (d), the training losses for OCR and grounding related datasets can be modeled as follows:
\begin{align*}
\log(\hat{L}_\text{ocr}) &\approx -0.1817\log(D)-0.7011 \\
\log(\hat{L}_\text{grounding}) &\approx -0.0785\log(D)-0.0745.
\end{align*}

Beyond the scaling laws of training loss, our analysis reveals that the training loss achieved on specific data sub-categories can serve as a predictor for performance on related downstream tasks. We find that the relationship between a sub-category's training loss and its corresponding downstream metric is approximately log-linear. However, it is important to note that such a log-linear relationship is likely sustainable only within a local neighborhood of performance values, as the range of typical evaluation metrics (\eg, accuracy, F1 score) is inherently bounded, usually between 0 and 1. As demonstrated in \cref{tab:scaling_laws} (b) and (c), the top-1 accuracies on the ChartQA and InfographicVQA datasets show a clear correlation with the logarithm of the OCR training loss ($\log(\text{loss}_{\text{OCR}})$), as captured by the following approximate linear models:
\begin{align*}
\text{Acc}_{\text{ChartQA}} &\approx -0.0968\log(\text{loss}_\text{ocr})+0.7139 \\
\text{Acc}_{\text{InfoVQA}} &\approx -0.1488\log(\text{loss}_\text{ocr})+0.5319
\end{align*}
Analogously, \cref{tab:scaling_laws} (e) and (f) detail the estimated relationship between the model's grounding loss during training and its performance on the RefCOCO evaluation benchmark. Performance prediction remains an active research area, and prior works have used a sigmoid function to model the relationship between LLM performance and loss~\cite{xiao2024densing,grattafiori2024llama} or compute~\cite{owen2024predictable}.
\section{Post-training}
\label{sec:posttrain}

\begin{figure*}[t]
    \centering
    \includegraphics[width=1.0\linewidth]{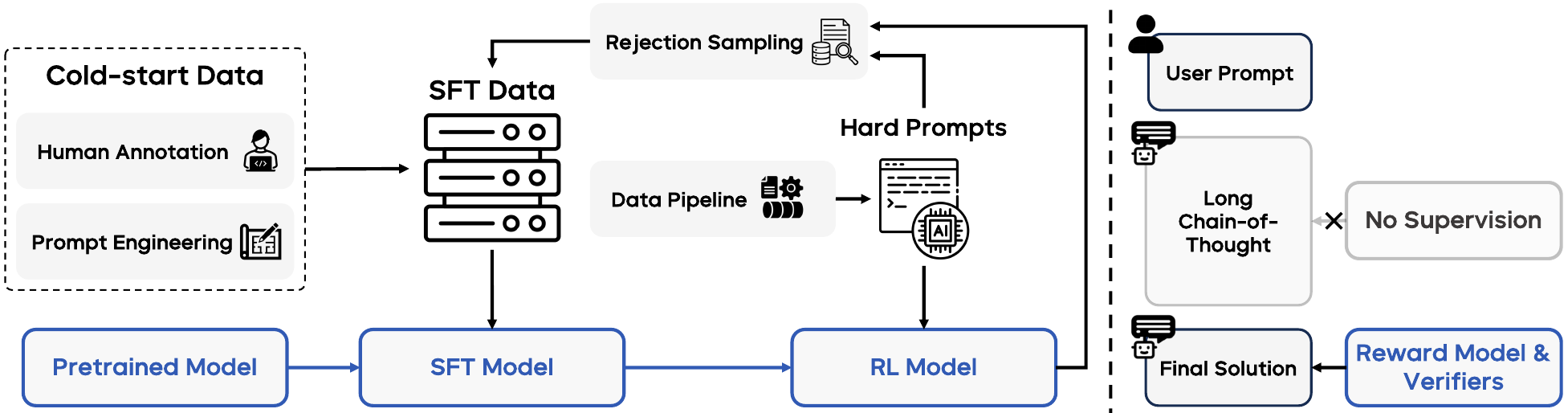}
    \caption{\textbf{The overview of post-training for \doubao{}}. The post-training for \doubao{} includes an iterative update combining rejection sampling and online reinforcement learning. We build a data pipeline including collection and curation of hard prompts for augmenting post-training data. A key aspect of our reinforcement learning implementation is that supervision, mediated by reward models and rule verifiers, is applied solely to the final generated output. We intentionally refrain from supervising the detailed chain-of-thought reasoning itself, a distinction highlighted in the illustration's right section.}
    \label{fig:posttrain}
\end{figure*}

The post-training stage equips \doubao{} with robust instruction-following and reasoning abilities through a combination of Supervised Fine-tuning (SFT) and Reinforcement Learning (RL). Depicted in \cref{fig:posttrain}, this begins with an SFT model trained on curated cold-start data. A crucial component is our data pipeline, continuously gathering hard and diverse prompts that feed into RL and improve SFT data via rejection sampling. Post-training proceeds iteratively: the SFT model is progressively enhanced by distilling the RL model's learnings on diverse prompts. This iterative refinement continues until the prompt pool is exhausted and performance metrics converge. Ultimately, this process yields \doubao{}, capable of generating both swift, succinct replies and in-depth responses featuring long Chain-of-Thought (LongCoT) reasoning~\cite{jaech2024openai}.
We discuss details of each component in the following subsections.

\subsection{Supervised Fine-tuning}

The Supervised Fine-tuning (SFT) stage is integral to equipping \doubao{} with foundational instruction-following and reasoning capabilities prior to reinforcement learning. Our SFT dataset comprises two primary components targeting distinct capabilities. The first component, General Instruction data, trains \doubao{} on diverse, complex instructions, emphasizing the generation of concise and accurate responses. The second, Long Chain-of-Thought (LongCoT) data, focuses on generating detailed, step-by-step reasoning. This data is generated via prompt engineering and rejection sampling (inspired by~\cite{touvron2023llama}), mainly using high-quality outputs from \doubao{}; specifics are detailed in \cref{sec:cold_start}. Besides, each data type is associated with a distinct system prompt, which allows users to dynamically toggle LongCoT reasoning during inference. The construction methodology for the SFT dataset and the specifics of \doubao{}'s SFT training regimen are further elaborated in \cref{sec:con_of_sft,sec:sft_training}, respectively.

\subsubsection{SFT Data Construction}\label{sec:con_of_sft}

In the initial phase of SFT data construction, we aimed to equip the model with the ability to address a broad spectrum of application scenarios. To this end, we developed a model capability taxonomy informed by the classification of traditional visual tasks and the empirical application requirements of vision-language models. Guided by this taxonomy, we utilized crowdsourcing to collect images from the internet and generate approximately 13,000 high-quality instruction-tuning data, each comprising a prompt and a corresponding response. These initial responses were designed to exhibit strong alignment with human preferences.

To further enhance the model's performance, we incorporated an additional 30,000 high-quality data samples sourced from the research community. These samples were curated from our carefully collected open-source repository containing approximately 1.5 million entries. Initially, we utilized a proprietary image-text embedding model to cluster the image-text pairs into task-specific categories. This clustering enabled targeted downsampling, ensuring the dataset preserved a high degree of diversity across various tasks. Subsequently, we leveraged our trained SFT model, aligned with human preferences, to perform multiple roll-outs on this sampled subset. The generated responses were filtered by LLM-as-a-judge~\cite{zheng2023judgingllmasajudgemtbenchchatbot}, which justifies the correctness of the model's generated responses with the original ground truth as reference. On this basis, we further adopted the Reward Model (\cref{sec:rm}) to screen out the responses that are most aligned with human preferences from the retained results, thus obtaining the final rejection sampling fine-tuning data~\cite{touvron2023llama}. Eventually, we compressed the amount of open-source data in the SFT data from 1.5 million to approximately 30,000 high-quality data. The other open-source data was used in the pre-training stage in advance. 

Building upon the enhanced capabilities acquired during pre-training, including complex chart understanding, STEM-related reasoning, grounding, and 3D perception, and video analysis, we iteratively increased the complexity of our fine-tuning data and instructions. This involved reducing the proportion of simple prompts readily solvable with individual capabilities and introducing more challenging questions that previously exposed limitations in the pre-trained model. Leveraging a self-instruct methodology~\cite{wang2022self}, we synthesized novel complex prompts and their corresponding model responses by combining multiple simpler prompts according to various logical structures. Responses generated through self-instruct and rejection sampling underwent a manual secondary verification process to identify and rectify errors. Compared to direct human annotation, this approach of refining model-generated responses significantly improves human annotation efficiency. Moreover, it enables the exclusion of data exceeding the model's current capacity, thereby mitigating the risk of hallucinations.

\subsubsection{Training Recipe}\label{sec:sft_training}
For the SFT stage, we assembled a concise and high-quality dataset comprising approximately 50,000 samples. This multimodal SFT data was integrated with an in-house text-only SFT dataset. Together with the Long Chain-of-Thought (LongCoT) SFT data, as described in \cref{sec:cold_start}, this combined corpus was used for training over two epochs. During SFT, the vision encoder's parameters were frozen, while all other model parameters remained trainable. The training was conducted with a sequence length of 131,072 tokens and a batch size equivalent to 16 times the sequence length. We utilized the AdamW optimizer~\cite{kingma2014adam} for training, with hyperparameters set to $\beta_1=0.9$, $\beta_2=0.95$, and a weight decay of 0.1. The training process included a warm-up phase spanning  $10\%$ of the total steps, after which the learning rate decayed from a peak value of $2\times10^{-5}$ to $2\times10^{-6}$ following a cosine decay schedule.

\subsection{Reinforcement Learning from Human Feedback}
To further boost both human evaluation performance and multimodal understanding capabilities, we conduct reinforcement learning from human feedback (RLHF)~\citep{ziegler2019fine}, which involves preference data collection, reward model training, and optimization with reinforcement algorithms.
\subsubsection{Preference Data}
To train the reward model, we collect list-wise multimodal preference datasets for reward modeling through human annotation and heuristic synthesis.

\paragraph{Human annotations} The human-annotated preference data involves comparing several candidate model responses using a 5-scale rating system. The prompts for generating preference data cover all general visual understanding abilities, and maintain a balanced scale across all abilities. We utilize the current top-performing in-house models to randomly sample responses through nucleus sampling~\cite{holtzman2019curious}. To ensure the diversity of responses, we apply filtering techniques—such as editing distance, semantic similarity, and length-balancing strategies—prior to selecting responses for human annotation. Beyond ranking the responses by quality, we instruct human annotators to select one model response that requires minimal editing to correct or improve its quality, which further compensates for the lack of diversity in the limited response sampling. Annotators are also tasked with identifying and highlighting issues within the responses—such as hallucinations, helpfulness, informativeness, etc.—and providing detailed explanations for these issues. To further enhance the efficiency of the annotation process, we employ the latest reward models to pre-annotate the rankings, offering initial guidance for human annotators. This approach not only streamlines the annotation workflow but also ensures more consistent and objective evaluations.

\paragraph{Synthetic data} While some recent approaches \cite{zhang2024critic, zhou2024aligning} have used deliberate error introduction to synthesize preference pairs, multiple studies \citep{amirloo2024understanding, ye2025painting, li2025devil} demonstrate that such synthetic data often fails to generalize effectively, as the reward model tends to learn the inherent patterns between edited and original responses. Instead, we aggregate a diverse set of multimodal prompts with clear ground-truths, while implementing format constraints such as ``Final Answer:''. For each prompt, we generate model responses $K$ times and use existing vision-language models to evaluate their correctness and adherence to format based on the ground-truth. Consequently, we establish list-wise preferences with clear rankings: correct responses with well-defined formats rank highest, followed by incorrect responses with well-defined formats, and lastly, incorrect responses that do not follow the format. Additionally, we follow FeedQuill~\citep{ye2025painting} to generate image captioning preference pairs, which helps in reducing hallucinations. All the synthetic preference data is refined using preference strength following ~\citep{wang2024secrets}.

\subsubsection{VLM as a Reward Model}\label{sec:rm}
We initialize the reward model with an instruction-tuned VLM. Then, following \citep{mahan2024generativerm, Shen2025ExploringDS} we prompt the model $\pi_{\phi}$ to act as a generative classifier that directly outputs answer indicator token $\hat{I}$ regarding the preference between two responses, $y_1$ and $y_2$, given the prompt $x$. This process can be formulated as $\hat{I} \sim \pi_{\phi}(I | x, y_1, y_2)$.

We find that this approach yields a more robust and superior reward model compared to traditional Bradley-Terry reward modeling~\cite{ouyang2022traininglanguagemodelsfollow} due to its direct handling of token probabilities and response comparisons. To mitigate the potential positional bias inherent in vision-language models~\citep{zheng2023judging}, we compute the probabilities for both possible orderings of the responses, i.e., both $(x, y_1, y_2)$ and $(x, y_2, y_1)$. This ensures that the model's preference judgment is fair and not affected by the order in which responses are presented.
Additionally, during training, we apply an iterative learning strategy to maintain the consistency of annotation principles as standards evolve. This strategy involves continuously updating the training data and annotation guidelines to reflect the most current and accurate criteria. By doing so, we ensure that the reward model remains reliable and adaptable to changing requirements. This approach helps in improving the generalization capability of the model and maintaining high-quality performance over time.

\subsubsection{Data Curation for Reinforcement Learning}
Our online reinforcement learning implementation employs a variant of the Proximal Policy Optimization (PPO) algorithm~\citep{xu2025unified}. In this approach, the reward signal is derived from the probability assigned by a reward model to the generated answer tokens. In addition, the ground truth response or the best-of-N responses from an SFT model are given as the reference answer to the reward model during PPO training.

Prompts utilized for RL training were derived from the preference dataset. It was observed that the coverage of the prompt distribution critically influences RL performance. Consequently, our data collection strategy aimed to mirror the distribution of the preference data. However, the collected prompts demonstrated significant heterogeneity in quality, characterized by highly skewed distributions across both task difficulty and ability categories. To address these issues, a multi-stage data refinement pipeline was implemented. Initially, a tagging model was trained to assign capability category labels to prompts, followed by stratified sampling to ensure a balanced representation across different ability categories. Subsequently, for each prompt, $K$ responses were generated using state-of-the-art internal models and evaluated using the most recent iteration of our reward model. A filtering criterion was applied based on the reward score variance: prompts where the difference between the maximum and mean reward across the K responses fell below a predefined threshold were excluded. This step ensures the retention of prompts for which the reward model exhibits significant discriminative capability. Finally, during the initial phases of RL training, prompts exhibiting rapid concurrent increase in both reward and KL divergence, indicative of lower task difficulty, were subject to downsampling.

\subsection{Reinforcement Learning with Verifiable Rewards}
In addition to human feedback, Reinforcement Learning with Verifiable Rewards (RLVR)~\cite{lambert2024t} emerges as an efficient training method for various tasks~\citep{guo2025deepseek,lambert2025tulu3pushingfrontiers}, such as mathematical reasoning and coding where we simply use answer matching or constraint verification to train the model, instead of leveraging model-based reward estimation. In this section, we design several visual tasks whose final solutions can be precisely verified by rules or external executors, which will later be incorporated into the RLVR training.

\subsubsection{Visual STEM}
STEM~(science, technology, engineering, and mathematics) questions usually have unique and verifiable answers, which are suitable for RLVR. We collect over one million problems with images in STEM fields, mostly on mathematics, from both open-sourced resources~\citep{luo2025ursa} and internal K-12 education collections.

To prepare the training data, multiple-choice questions were initially transformed into an open-ended format by removing the choices, thus forcing the model to generate the correct answer's content and preventing random guessing. Subsequently, difficult questions were selected via rejection sampling based on the performance of the SFT model. We carefully remove questions that can be answered by text only or text and captions, ensuring shortcuts on text or superficial visual elements will not be reinforced in RL. Specifically, 16 responses were generated per question, and questions achieving either 0\% or greater than 75\% accuracy with the SFT model were discarded. This filtering isolates challenging prompts ($0\% < \text{accuracy} \le 75\%$) appropriate for RLVR exploration while removing potentially erroneous or trivial questions. Lastly, a preamble instruction was prepended to prompts, instructing the model to format the final answer using designated LaTeX identifiers (\eg, \texttt{\textbackslash boxed\{answer\}}) to enable straightforward automated extraction.

Our STEM verifier transforms the predicted answers into a sympy expression and matches it with ground truths. To ensure the accuracy of our verifier, we also remove prompts that contain multiple questions or whose ground truths are complex phrases.

\subsubsection{Visual Perception and Reasoning}
Verifier feedback can also be collected from various visual tasks to enhance the perception and reasoning capabilities of VLMs. Here we present some early explorations on grounding, visual puzzles, and perception-related games.

\paragraph{Grounding} The grounding task aims to evaluate a model's ability to accurately associate (``ground'') textual descriptions with corresponding visual elements within an input image. For easier answer extraction, we add an instruction in the prompt to encourage the model to output the predicted bounding boxes enclosed between \texttt{<bbox>} and \texttt{</bbox>} tokens. The reward is computed as the intersection over union (IoU) between the predicted bounding box and the ground-truth one. We also optimize for pointing capability in a similar way and put the object' center point position between \texttt{<point>} and \texttt{</point>}.

\paragraph{Visual Instruction Following} Instruction-following capabilities can be improved with synthetic data and rule-based verifiers~\cite{dong2024selfplayexecutionfeedbackimproving,yao2023colliesystematicconstructionconstrained}. Following this idea, we synthesize diverse visual instructions whose outcomes can be verified by corresponding regular expressions to further enhance visual instruction-following capabilities.

\paragraph{Visual Puzzles \& Games} Visual puzzles are tasks that require the model to gather information from a visual scene and apply reasoning techniques such as abstract reasoning, inductive reasoning, and deductive reasoning. Similar to \cite{chia2024puzzlevqadiagnosingmultimodalreasoning,toh2025jumpingreasoningcurvetracking}, we synthesize over 20k visual puzzles and their corresponding solutions for RLVR. We carefully decontaminate our synthetic training data with existing visual puzzle benchmarks, such as PuzzleVQA~\cite{chia2024puzzlevqadiagnosingmultimodalreasoning}. We also involve puzzles in graph reasoning~\citep{wei2024gitagraphvisualtextual} and pattern identification. Similar to the STEM verifier, we prompt models to enclose final answers of puzzles in \texttt{\textbackslash boxed\{answer\}} and verify the prediction through a string matching algorithm.

\begin{figure}[ht]
\centering
\includegraphics[width=0.9\textwidth]{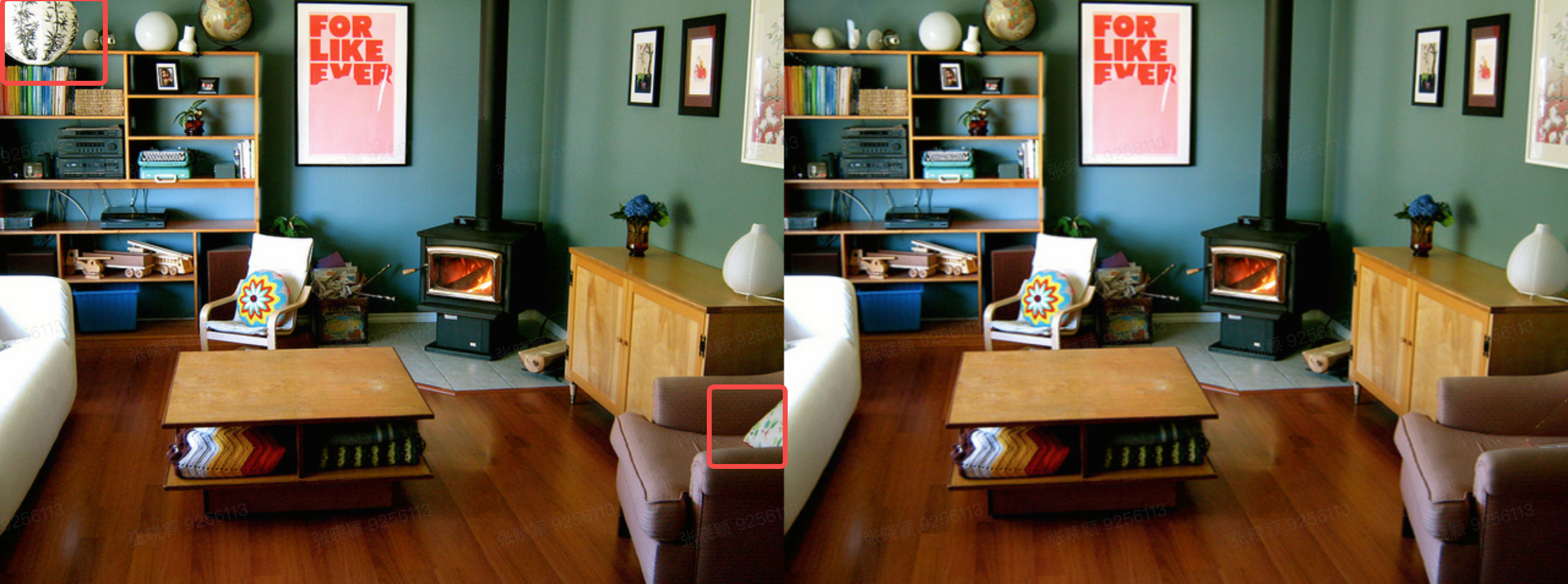}
\caption{An example of a synthesized image pair used for training the ``Spot the Differences'' game, with the differences highlighted by red boxes in the left image.}
\label{fig:spot_diff}
\end{figure}

Beyond generating natural language responses, we are exploring VLM output formats that enable direct interaction with or manipulation of image content, aiming to facilitate broader VLM applications through more intuitive and engaging interactions. Imagine, for example, AI-enhanced glasses overlaying a navigation route directly onto the user's view, rather than relying solely on text or speech---a potentially more intuitive approach. As an initial step towards developing these interactive capabilities, we focus on visual games, which are suitable testbeds because they require strong perceptual skills and have clearly verifiable outcomes indicating success. Specifically, we target the ``Spot the Differences'' game, tasking the model with identifying discrepancies between two images. Crucially, the model must not only explain these differences using natural language but also output bounding boxes that precisely localize the differing regions directly on the image. We train this capability using synthetically generated data employing two methods: (1) We take images from open-sourced datasets, randomly mask segments, use a diffusion model for inpainting (see \cref{fig:spot_diff} for an example), and then filter out pairs where the inpainted content is too similar to the original; (2) To ensure the model perceives subtle differences like line width or object size, we generate additional image pairs by systematically modifying SVG properties from open-sourced datasets.

\subsection{Hybrid Reinforcement Learning}
The \doubao{} model is trained utilizing a hybrid RL framework derived from a variant of the PPO algorithm. This framework incorporates a generative RM, as detailed in \cite{xu2025unifiedpairwiseframeworkrlhf}, and integrates several advancements and exploration techniques from recent RL research \cite{yue2025vapoefficientreliablereinforcement, shen2025exploringdatascalingtrends, yu2025dapoopensourcellmreinforcement, yuan2025whatspposcollapselongcot}. 
Specifically, our training is a combination of RLHF and RLVR. We present more detailed implementations as follows:

\paragraph{Format reward} We predefine a response format of \texttt{<think>\{thought\}</think>\{solution\}} to ensure models provide comprehensive thoughts before giving the final solution. We set rewards to zero if the model's responses do not comply with this format. We also apply penalties if responses fail to follow format requirements for different verifiers in various tasks.

\paragraph{Hybrid reward} Our training prompts are categorized into general and verifiable prompts based on tasks, rewarded with RM and the verifier, respectively. Prompts are randomly shuffled in each epoch. So, general and verifiable prompts are mixed in each batch. We truncate the thought and only keep the solution in response to the reward model. Therefore, RM will ignore the CoT thought and only focus on providing rewards for the final solution. Such modification can ease constraints on thoughts and encourage models to explore more effective CoT thoughts.

\paragraph{Shared critic} A single critic model architecture is employed to estimate the value function corresponding to both reward sources (\ie, the reward model and verifiers). This unified approach is viable due to both reward signals operating within the same normalized range of $[0,1]$. Specifically, the reward model inherently generates outputs within this interval, while the outcomes derived from all verifiers are explicitly scaled to conform to the same $[0,1]$ range. The critic model's parameters are initialized using the weights of the pre-trained reward model. Subsequently, the critic undergoes an initial warm-up phase consisting of 100 training steps, utilizing trajectory data (rollouts) generated by the SFT model.

\paragraph{KL coefficients} We employ distinct KL divergence coefficients for general and verifiable prompts. Specifically, a coefficient of $1\times10^{-5}$ is applied to general prompts, while a coefficient of 0 is used for verifiable prompts. The application of a small KL coefficient for general prompts serves to mitigate potential reward hacking. Conversely, training verifiable tasks without a KL divergence term facilitates greater exploratory capacity for the model.

\paragraph{Training recipe} The context length and max output length of hybrid RL training are 8,192 and 16,384, respectively. We sample 4,096 roll-outs in each episode. For training updates, we use a mini-batch size of 512 samples, performing 8 gradient steps per episode. PPO clip range for the training is 0.2. Learning rates for the actor and critic are $6\times 10^{-7}$ and $7.5\times 10^{-7}$, respectively.
The number of roll-outs is different for each prompt, as harder prompts need more comprehensive exploration.
We only sample once for each prompt rewarded by the reward model, while sampling 4 or 8 times for the counterpart rewarded by verifiers.
Noticeably, although we only train \doubao{} with LongCoT responses in the RL stage, we still witness a significant improvement in regular responses without extended reasoning.

\subsection{Iterative Update by Rejection Sampling Fine-tuning} \label{sec:cold_start}
In this work, we employ an iterative training strategy to enhance \doubao{} during the RL stage. The process commences with a cold-start SFT model for LongCoT, initially trained on a limited number of low-quality LongCoT samples generated via in-context prompting of the base model with a small set of hand-annotated examples. Observing that a stronger cold-start SFT naturally leads to a stronger final model after LongCoT RL, we adopt a rejection sampling fine-tuning approach to obtain an improved starting point. Specifically, following the release of each iteration of the LongCoT RL model, we gather additional challenging prompts through our data pipeline and evaluate the latest RL model on these prompts. Correctly answered responses are then collected, in the vein of rejection sampling, and incorporated into the data for the subsequent SFT release. The same verifiers used in the RL phase are utilized to confirm the correctness of these responses. Furthermore, we implement manually crafted regular expression-based filters to remove undesirable patterns such as infinite repetition, overthinking, and other linguistic artifacts. The current iteration of \doubao{} has undergone four such rounds of iteration, demonstrating consistent improvements, and this iterative refinement is expected to further enhance its performance.

\clearpage
\section{Training Infrastructure}
\label{sec:infra}

\subsection{Large-Scale Pre-training}
To accelerate and stabilize pretraining, we have developed a number of training optimizations, including hybrid parallelism, workload balancing, parallelism-aware data loading and robust training. We also apply high-performance attention kernels for context parallelism, selective activation checkpointing and offloading, kernel fusion, and fine-grained communication overlapping~\cite{chang2024flux,zhang2025comet}. The pretraining phase consumes 1.3 million GPU hours in total\footnote{For consistency, all computational costs mentioned in this report are normalized to GPU hours based on the H800.}.

\subsubsection{Hybrid Parallelism}
Training a VLM model faces unique challenges due to the heterogeneity of both the data, which consists of visual data and natural language data, and the model, which consists of a small vision encoder and a significantly larger language model. Existing training frameworks are primarily designed for sequential unimodal tasks and fall short in VLM training. They either treat the encoder as preprocessing for the LLM's data, or completely disaggregate the encoder from the LLM, leading to imbalanced workloads, prolonged device stalls and poor scalability.
To tackle these challenges, we develop a hybrid parallelism approach~\cite{feng2024optimus} that parallelizes the vision encoder and the language model differently. For the vision encoder and the MLP adaptor, we leverage ZeRO data parallelism~\cite{rajbhandari2020zero}, while for the language model, we use standard 4-D parallelism, which combines expert parallelism~\cite{shoeybi2019megatron,korthikanti2023reducing}, interleaved pipeline parallelism~\cite{huang2019gpipe,narayanan2021memory,narayanan2021efficient}, ZeRO-1 data parallelism~\cite{rajbhandari2020zero} and context parallelism~\cite{liu2023ring} for context extension. We separate the parallelism strategies for the encoder/adaptor and the LLM for efficiency and simplicity--it is challenging to integrate the encoder and the adaptor into 4-D parallelism without introducing pipeline-level imbalance. Our hybrid parallelism is simple and efficient, significantly accelerating training with minimal changes to model code.

\subsubsection{Workload Balancing}
Vision samples contain a varying number of images, causing computation imbalance among GPUs. We adopt a classical greedy algorithm to redistribute the vision data to achieve load balancing for the vision encoder and adaptor. Firstly, we sort the images in descending order according to their computation intensity, which is defined as the number of floating-point operations (FLOPS) needed to process each image. Secondly, we scan these images in the sorted order, and assign each image to the GPU with the lowest total computation intensity. Additionally, we leverage group-wise balancing to reduce data redistribution overhead. Instead of balancing  vision data across all GPUs, we divide them into evenly sized groups and only balance vision data within each group only. Empirically, we set the group size to 128-256 GPUs.

\subsubsection{Parallelism-Aware Data Loading}
To reduce multimodal data IO overhead, we have also built a parallelism-aware data loader. For example, GPUs within non-data-parallel groups are expected to consume the same set of training samples. Redundantly reading the same data from the distributed file system can significantly amplify data read and preprocessing overhead, slowing down microbatch readinesss. We address this problem using a parallelism-aware data loader. For example, only one GPU within a PP group loads the data while the other PP ranks receive the necessary metadata from it via broadcast. Additionally, since we use pure data parallelism for the vision encoder, each GPU only processes a portion of the loaded image data. We filter out unnecessary images before moving training batches to the GPU, reducing PCIe traffic. To hide these data broadcast and transfer costs, we use a prefetcher to ensure IO and computation fully overlap.

\subsubsection{Fault Tolerance}
To handle various hardware and software faults during training, we use the robust training framework MegaScale~\cite{jiang2024megascale} to achieve fault tolerance. Once the robust training framework detects a fault, it triggers the recovery process and resumes training from the last successful checkpoint. We leverage ByteCheckpoint~\cite{wan2024bytecheckpoint} for efficient checkpoint saving and resuming. 

\subsection{Post-Training Framework}
We conduct hybrid reinforcement learning with both human feedback (RLHF) and verifier feedback (RLVF) of \doubao{} on a verl-based~\cite{sheng2025hybridflow} framework. It combines a single-controller for managing inter-RL-role dataflow and multi-controllers for managing intra-RL-role data and model parallelism. Verifiers are deployed in process-based services to isolate potential verifier faults. This design greatly simplifies deployment and development for various experiments.
We use the same training system and optimization techniques as in the pretraining phase for efficient actor and critic updates, and vLLM~\cite{kwon2023efficientmemorymanagementlarge} for autoregressive generation of rollouts. Specifically, actor and critic training employs 3-D parallelism~\cite{shoeybi2019megatron,huang2019gpipe,narayanan2021memory, rajbhandari2020zero}; rollout generation and reward/reference model inference use replicas, each configured with tensor parallelism~\cite{seed2025seedthinking}. The RL phase of \doubao{} costs 60k GPU hours. The reward model is trained using the same framework as the \doubao{} pretraing phase, requiring 24k GPU hours. Post-training phases also leverage ByteCheckpoint~\cite{wan2024bytecheckpoint} for efficient checkpoint saving and resuming.
\newcommand{\gemini}{Gemini 2.5 Pro}
\newcommand{\openaio}{OpenAI o1}
\newcommand{\claude}{Claude 3.7 Sonnet}
\newcommand{\openaigpt}{OpenAI GPT-4o}
\newcommand{\qwen}{Qwen 2.5-VL 72B}
\newcommand{\openaicua}{OpenAI CUA}

\section{Evaluation}
\label{sec:evaluation}

This section is structured as follows. Quantitative results on public benchmarks are presented in \cref{sec:public_benchmark}, followed by an assessment of performance on agentic tasks in \cref{sec:agent_eval}. The design of our internal benchmark and a comparison of our model against industry-leading models are subsequently detailed in \cref{sec:eval:internal}. Model limitations are discussed in \cref{sec:limitations}. Qualitative examples are provided in \cref{sec:app:qualitative}, and comprehensive evaluation settings are described in \cref{app:eval_detail}.

\subsection{Public Benchmarks}
\label{sec:public_benchmark}

\subsubsection{Vision Encoder as a Zero-shot Classifier}
\label{sec:vision-encoder}
We evaluate \vit using zero-shot image classification benchmarks, including ImageNet-1K~\cite{deng2009imagenet}, ImageNet-V2~\cite{recht2019imagenet}, ImageNet-A~\cite{hendrycks2021natural}, ImageNet-R~\cite{hendrycks2021many}, ImageNet-S~\cite{wang2019learning}, and ObjectNet~\cite{barbu2019objectnet}. As detailed in \cref{tab:vit_eval}, \vit achieves an average zero-shot accuracy of 82.5 across these datasets, which is comparable to that of InternVL-C-6B~\cite{chen2024internvl}, despite the fact that the number of parameters of \vit is only 9\% of that of InternVL-C-6B. Impressively, compared to EVA-CLIP-18B, which has 30$\times$ more parameters, \vit achieves comparable accuracies on most of the ImageNet variants. Furthermore, compared to DFN-5B-CLIP-H/14++~\cite{fang2023data}, \vit demonstrates superior performance on ObjectNet (which contains images with challenging backgrounds, rotations, and viewpoints) and ImageNet-A (which contains natural adversarial examples), suggesting greater robustness of \vit to real-world variations.

\begin{table}[htbp]
    \centering
    
    \begin{tabular}{lc|cccc}
    \toprule
    \textbf{Models} & \textbf{Seed-ViT} & \textbf{OpenCLIP-G/14} & \textbf{DFN-5B-CLIP-H/14++} & \textbf{InternVL-C} & \textbf{EVA-CLIP-18B}\\
    \#Param    & \textbf{532M}   & 1.8B   & 632M & 6B  & 17.5B \\
    \midrule
    ImageNet-1K & 83.6  & 80.4  & 84.3  & 83.2  & 83.8 \\
    ImageNet-V2    & 77.6  & 73.6  & 78.3  & 77.3  & 77.9 \\
    ImageNet-A  & 85.5  & 69.3  & 79.6  & 83.8  & 87.3 \\
    ImageNet-R  & 95.2  & 92.8  & 94.9  & 95.7  & 95.7 \\
    ImageNet-S  & 74.1  & 69.9  & 73.6  & 74.3  & 74.7 \\
    ObjectNet   & 79.2  & 73.0  & 78.0  & 80.6  & 82.2 \\
    \midrule
    \textit{Avg.}        &  82.5  & 76.5  & 81.4  & 82.5  & 83.6 \\
    \bottomrule
    \end{tabular}
    \caption{Comparisons of pre-trained Seed-ViT (before integration with the LLM) and existing competitors with more parameters on the common zero-shot benchmarks.}
    \label{tab:vit_eval}
\end{table}

\subsubsection{Vision Task Evaluation}
We evaluated the performance of \doubao{} on a comprehensive suite of public image benchmarks, comparing it against several state-of-the-art multimodal models including \gemini{}~(0325 version), \openaio{}, \claude{}, \openaigpt{}, and \qwen{}.
We compare \doubao{} with \gemini{}~(Preview 03-25) instead of \gemini{}~(Preview 05-06) as \gemini{}~(Preview 03-25) shows stronger capabilities in open visual-language benchmarks~(81.7\textsubscript{Preview 03-25} v.s. 79.6\textsubscript{Preview 05-06} in MMMU)\footnote{\url{https://deepmind.google/technologies/gemini/pro/}}.
The evaluation covers capabilities ranging from multimodal reasoning and general visual question answering to document understanding, grounding, and spatial reasoning.
\Cref{tab:image_eval_combined} presents the detailed results, highlighting the highest score in bold and the second highest score underlined for each benchmark, except for FSC-147 and NYU-Depth V2 where lower is better. We report results for \doubao{} in both its standard `non-thinking' mode and an enhanced `thinking' mode, incorporating long chain-of-thought to improve reasoning.
\begin{table}[htbp]
\centering
\small
\begin{threeparttable}
\setlength{\tabcolsep}{1.8mm}{
\begin{tabular}{llll|lllll} \toprule
    \multirow[t]{3}{*}{\textbf{Capability}} & \multirow[t]{3}{*}{\textbf{Benchmark}} & \textbf{Seed} & \textbf{Seed} & \textbf{Gemini} & \textbf{OpenAI}  & \textbf{Claude} & \textbf{OpenAI} & \textbf{Qwen} \\
    & & \textbf{1.5-VL} & \textbf{1.5-VL} & \textbf{2.5 Pro} & \textbf{o1} & \textbf{3.7 Sonnet} & \textbf{GPT-4o} & \textbf{2.5-VL 72B} \\
    & & {\scriptsize thinking} & {\scriptsize non-thinking} & {\scriptsize thinking} & {\scriptsize thinking} & {\scriptsize thinking} & {\scriptsize non-thinking} & {\scriptsize non-thinking} \\
    \midrule
    \multirow{10}{*}{\shortstack[l]{{\scriptsize Multimodal}\\{\scriptsize reasoning}}} & MMMU & \underline{77.9} & 73.6 & {\bf 81.7} & 77.6 & $75.2^*$ & $70.7^*$ & 70.2 \\
    & MMMU-Pro & \underline{67.6} & 59.9 & $\mathbf{68.8}^{*}$ & $66.4^*$ & $50.1^*$ & $54.5^*$ & 51.1 \\
    & MathVision & \underline{68.7} & 65.5 & $\mathbf{73.3}^*$ & $63.2^*$ & $58.6^*$ & $31.2^*$ & 38.1 \\
    & OlympiadBench & \underline{65.0} & 60.4 & $\mathbf{69.8}^*$ & $48.5^*$ & $54.2^*$ & $25.9^*$ & $35.9$ \\
    & MathVista & {\bf 85.6} & \underline{83.0} & $82.7^*$ & 71.8 & $74.5^*$ & $63.8^*$ & 74.8 \\
    & V\textsuperscript{*} & \underline{89.0} & {\bf 89.5} & $79.1^*$ & $69.7^*$ & $86.4^*$ & $73.9^*$ & 86.4 \\
    & VLM are Blind & {\bf 92.1} & \underline{90.8} & $84.3^*$ & $57.0^*$ & $69.0^*$ & $50.4^*$ & 69 \\
    & ZeroBench~(main) & \underline{2} & 0 & $\mathbf{3}^*$ & $0^*$ & $\mathbf{3}^*$ & $0^*$ & 0 \\
    & ZeroBench~(sub) & {\bf 30.8} & \underline{29.0} & $26.0^*$ & $20.2^*$ & $20.4^*$ & $19.6^*$ & 13.0 \\
    & VisuLogic & {\bf 35.0} & \underline{33.0} & $31.0^*$ & $29.0^*$ & $24.8^*$ & $26.3^*$ & 28.0 \\
    \midrule
    \multirow{7}{*}{\shortstack[l]{{\scriptsize General}\\{\scriptsize visual}\\{\scriptsize question}\\{\scriptsize answering}}} & RealWorldQA & {\bf 78.4} & 77.0 & \underline{$78.0^*$} & $77.1^*$ & $67.8^*$ & $76.2^*$ & 75.7 \\
    & SimpleVQA & {\bf 63.4} & \underline{63.1} & $62.0^*$ & $58.8^*$ & $50.1^*$ & $52.4^*$ & 52.4 \\
    & MMStar & {\bf 77.8} & 76.2 & \underline{$77.5^*$} & $67.5^*$ & $68.8^*$ & $65.1^*$ & 70.8 \\
    & MMBench-en & \underline{89.9} & 88.0 & $\mathbf{90.1}^*$ & $83.8^*$ & $82.0^*$ & $84.3^*$ & 88.6 \\
    & MMBench-cn & \underline{89.1} & 88.1 & $\mathbf{89.7}^*$ & $81.3^*$ & $82.7^*$ & $82.0^*$ & 87.9 \\    
    & MMVP & \underline{69.3} & {\bf 70.7} & $\mathbf{70.7}^*$ & $-$\tnote{\dag} & $-$\tnote{\dag} & $\mathbf{70.7}^*$ & 66.7 \\
    & HallusionBench & \underline{60.3} & 60.0 & $\mathbf{63.7}^*$ & $55.6^*$ & $58.3^*$ & $56.2^*$ & 55.2 \\
    \midrule
    \multirow{8}{*}{\shortstack[l]{{\scriptsize Document}\\ {\scriptsize and chart}\\{\scriptsize understanding}}} & TextVQA & 81.8 & {\bf 84.2} & $76.8^*$ & $66.2^*$ & $62.4^*$ & $81.4^*$ & \underline{83.5} \\
    & AI2D & 87.3 & \underline{88.5} & $88.4^*$ & $79.5^*$ & $82.1^*$ & $84.9^*$ & {\bf 88.7} \\
    & ChartQA & \underline{89.1} & 87.4 & $83.3^*$ & $83.1^*$ & $56.5^*$ & $86.7^*$ & {\bf 89.5} \\
    & InfographicVQA & {\bf 91.2} & \underline{89.3} & $84.3^*$ & $65.4^*$ & $66.5^*$ & $79.2^*$ & 87.3 \\
    & DocVQA & {\bf 96.9} & \underline{96.7} & $94.0^*$ & $81.6^*$ & $87.4^*$ & $66.2^*$ & 96.4 \\    
    & OCRBench & 861 & \underline{881} & $866^*$ & $750^*$ & $793^*$ & $806^*$ & {\bf 885} \\
    & CharXiv~(RQ) & 60.2 & 59.8 & $\mathbf{69.9}^*$ & $55.1^*$ & \underline{$68.9^*$} & $52.0^*$ & $49.7^*$ \\
    & CharXiv~(DQ) & \underline{92.6} & \underline{92.6} & $\mathbf{94.4}^*$ & $88.9^*$ & $92.0^*$ & $86.5^*$ & $87.4^*$ \\
    \midrule
    \multirow{6}{*}{\shortstack[l]{{\scriptsize Grounding}\\{\scriptsize \& counting}}} & BLINK & {\bf 72.1} & 70.2 & \underline{$70.6^*$} & $66.1^*$ & $62.5^*$ & $65.9^*$ & 64.4 \\
    & LVIS-MG & \underline{$72.5$} & $\mathbf{73.8}$ & $63.8^*$ & $-^\dagger$ & $-^\dagger$ &  $-^\dagger$ & $-^\dagger$\\
    & VisualWebBench & \underline{$87.3$} & $\mathbf{88.0}$ & \underline{$87.3^*$} & $80.9^*$ & $85.9^*$ & $80.2^*$ & $82.3^*$ \\
    & RefCOCO-avg & \underline{91.3} & $\mathbf{91.6}$ & $74.6^*$ & $-^\dagger$ & $-^\dagger$ & $-^\dagger$ & 90.3 \\
    & CountBench & $\mathbf{93.7}$ & $93.5$ & $91.0^*$ & $86.6^*$ & $86.1^*$ & $85.7^*$ & \underline{93.6} \\
    & FSC-147 $\downarrow$ & $\mathbf{17.9}$ & $\underline{18.6}$ & $24.5^*$ & $34.3^*$ & $33.4^*$ & $46.8^*$ & $28.6^*$ \\
    \midrule
    \multirow{3}{*}{\shortstack[l]{{\scriptsize 3D Spatial}\\{\scriptsize understanding}}} & DA-2K & \underline{91.7} & $\mathbf{91.9}$ & $73.0^*$ & $72.3^*$ & $40.1^*$ & $66.9^*$ & $69.6^*$ \\
    & NYU-Depth V2 $\downarrow$ & \underline{13.6} & $\mathbf{11.6}$ & $27.5^*$ & $82.1^*$ & $92.4^*$ & $73.8^*$ & $35.5^*$ \\
    & All-Angles Bench & \underline{58.6} & $\mathbf{59.0}$ & $53.4^*$ & $54.0^*$ & 50.0 & $49.1^*$ & 55.7 \\
    \bottomrule
\end{tabular}}
\begin{tablenotes}
    \footnotesize
    \item[*] Results self-collected via API in April 2025.
    \item[\dag] Invalid results due to failures in following format requirements.
\end{tablenotes}
\end{threeparttable}
\caption{Performance of \doubao{} on public visual-language benchmarks~(\cref{sec:benchmarks}) compared to previous models. All benchmarks are evaluated with greedy decoding except for Claude-3.7 Sonnet where a default sampling mode is recommended. We report Pass@1 in these benchmarks. For FSC-147 and NYU-Depth V2, Mean Absolute Error (MAE) and Absolute Relative Error (AbsRel) are used as the metrics, respectively, so lower numbers are better. For all other benchmarks, higher numbers are better. The highest score in each benchmark is marked in \textbf{bold}, and the second is \underline{underlined}.}
\label{tab:image_eval_combined}
\end{table}

\paragraph{Multimodal Reasoning}
In complex multimodal reasoning tasks, \doubao{} demonstrates strong capabilities in both thinking and non-thinking modes. Notably, it achieves state-of-the-art (SOTA) performance on MathVista (85.6 thinking), V* (89.5 non-thinking), VLM are Blind (92.1 thinking), ZeroBench (sub) (30.8 thinking), and VisuLogic~(35.0 thinking). On MathVista and VLM are Blind, \doubao{} significantly outperforms all listed counterparts. While \gemini{} leads on benchmarks like MMMU (81.7 vs. 77.9 for the thinking mode in \doubao{}), MMMU-Pro (68.8 vs. 67.6), MathVision (73.3 vs. 68.7), and OlympiadBench (69.8 vs. 65.0), \doubao{} remains competitive, securing the second position. For ZeroBench (main), \doubao{} in the thinking mode solves 2 cases, ranking second alongside \openaio{}, behind \gemini{} and \claude{}. \doubao{} in the non-thinking mode also significantly excels in all multimodal reasoning compared with its non-thinking counterparts.

We observed that the model naturally exhibited diverse vision-centric strategies during our first round of LongCoT RL training, such as "let me look at the image again" and "analyze details before recognizing a location", as shown in \cref{fig:vlm_are_blind_case} and \cref{fig:geoguess_case}, even though we had not labeled related SFT data at that time.

\paragraph{General Visual Question Answering}
For general visual question answering benchmarks, \doubao{} shows robust performance. It achieves SOTA results on RealWorldQA (78.4 thinking) and SimpleVQA (63.4 thinking). On MMStar, \doubao{} (77.8 thinking) also achieves the highest score among the compared models.  Similarly, on MMBench-en (89.9 thinking) and MMBench-cn (89.1 thinking), \doubao{} scores are near the top performers like \gemini{} and \qwen{}. On HallusionBench, \doubao{} (60.3 thinking) secures the second-best score, slightly behind \gemini{} (63.7).

\paragraph{Document and Chart Understanding}
\doubao{} excels in document and chart understanding tasks. It sets new SOTA benchmarks on TextVQA (84.2 non-thinking), InfographicVQA (91.2 thinking), and DocVQA (96.9 non-thinking), surpassing strong models like \qwen{} and \gemini{} in these areas. On ChartQA, \doubao{} (89.1 thinking) achieves the second-highest score, only behind \qwen{} (89.5). It also delivers strong performance on AI2D (88.5 non-thinking) and OCRBench (881 non-thinking), ranking competitively behind \qwen{} and \gemini{}. For CharXiv (DQ), \doubao{} (92.6 thinking and non-thinking) ranks second to \gemini{} (94.4). However, on CharXiv (RQ), its performance (60.2 thinking) lags behind the leaders \gemini{} (69.9) and \claude{} (68.9).

\paragraph{Grounding and Counting}
This category highlights a significant strength of \doubao{}. It achieves SOTA performance across \textit{all} listed grounding and counting benchmarks. Specifically, \doubao{} leads on BLINK (72.1 thinking), LVIS-MG (73.8 non-thinking), VisualWebBench (87.8 non-thinking), RefCOCO-avg (91.6 non-thinking), CountBench (93.7 thinking), and FSC-147 (17.9 thinking, lower is better). Notably, \doubao{} achieves better performance on LVIS-MG against to traditional detectors, \ie, Grounding DINO-L~\cite{liu2024grounding,mmdetection}, which obtains 54.4 F1-score, demonstrating the strong capability of \doubao{} in terms of multi-object grounding. The consistent top performance across these diverse tasks underscores \doubao{}'s superior capabilities in object localization, fine-grained visual understanding, and counting.

\paragraph{3D Spatial Understanding}
We select depth estimation, 3D object detection, and multi-view reasoning as the three tasks to evaluate \doubao{}'s capability on 3D spatial understanding. In particular, for depth estimation, we report results on two public benchmarks, DA-2K~\cite{depth_anything_v2} and NYU-Depth V2~\cite{Silberman2012indoor}. In DA-2K, we follow~\cite{depth_anything_v2} and report the accuracy of relative depth estimation between two pixels (\eg, which pixel is closer). In NYU-Depth V2, we report the standard absolute relative error measured as $|\text{dist}_\text{pred}-\text{dist}_\text{gt}|/\text{dist}_\text{gt}$ where $\text{dist}_\text{pred}$ and $\text{dist}_\text{gt}$ are the predicted and ground truth distances, respectively.
As shown in~\cref{tab:image_eval_combined}, \doubao{}-thinking scores 91.7 on DA-2K and 0.136 error rate on NYU Depth V2, which surpasses previous VLMs by a large margin. In non-thinking mode, \doubao{} achieves 91.9 and 0.116 error rate on DA-2K and NYU-Depth V2, respectively. For 3D object detection, we report results on SUN-RGBD~\cite{song2015sun}. In non-thinking mode, our model scores 33.5 AP@{15} on SUN-RGBD surpassing Gemini 2.0 Pro Experimental, which scores 32.5 AP@{15}~\cite{team2025gemini}. However, we observed a performance regression using thinking mode for this task. Namely, the result is decreased to 32.0 AP@{15}.
For multi-view reasoning, we conduct evaluation on All-Angles Bench~\cite{yeh2025seeing}. \doubao{} attains 59.0 in non-thinking mode and 58.6 in thinking mode, which significantly surpasses previous models.

In summary, \doubao{} exhibits state-of-the-art or highly competitive performance across a wide range of visual language benchmarks. It particularly excels in grounding, counting, 3D spatial understanding, document understanding (TextVQA, DocVQA, InfographicVQA), and certain reasoning tasks (MathVista, VLM are Blind, etc.), establishing itself as a powerful and versatile multimodal model.

\subsubsection{Video Task Evaluation}
We conduct an evaluation of \doubao{}'s proficiency in video understanding, assessing its capabilities across five dimensions: short video, long video, streaming video, video reasoning, and video grounding. \Cref{tab:video_eval} benchmarks \doubao{} against state-of-the-art (SOTA) models.
Due to API limitations (e.g., network timeouts, video processing errors), we cannot evaluate certain proprietary models such as Gemini 2.5 Pro across all benchmarks. Therefore, the table reports the highest score obtained, either sourced from public reports or self-collected via API.

For short video understanding, \doubao{} achieves SOTA performance on MotionBench, TVBench, Dream-1K, and TempCompass, demonstrating its exceptional proficiency in processing temporal dynamics and motion patterns characteristic of concise video segments.
For long video understanding, it also attains strong results with a 128K token context (up to 640 frames). We recognize the importance of extended temporal understanding and plan future work focused on expanding this context window capacity to further enhance long-form video comprehension.
Regarding streaming video understanding, we evaluate on OVBench~\cite{ovbench}, OVOBench~\cite{li2025ovo}, StreamBench~\cite{xiong2025streaming}, and the proactive sub-task of StreamingBench~\cite{lin2024streamingbench}. \doubao{} achieves SOTA performance across all these benchmarks, indicating strong potential for real-time applications such as interactive video dialogue systems. 
In video reasoning (Video-MMMU~\cite{hu2025video}, MMVU~\cite{zhao2025mmvu}), \doubao{} scores 81.4 and 70.1, respectively, currently trailing top models such as Gemini 2.5 Pro.
Furthermore, \doubao{} excels in video grounding tasks, specifically designed to locate temporal segments within videos corresponding to textual descriptions. It achieves SOTA performance on Charades-STA~\cite{gao2017tall} and TACoS~\cite{Rohrbach2014CoherentMV}, demonstrating precise localization capabilities.

\renewcommand{\arraystretch}{1.2}
\begin{table}[htbp]
\centering
\small
\begin{threeparttable}
\setlength{\tabcolsep}{4.7mm}{
\begin{tabular}{llll|l} \toprule
    \textbf{Capability} & \textbf{Benchmark} & \Shortunderstack[l]{\textbf{Seed1.5-VL}\\{\scriptsize thinking}} & \Shortunderstack[l]{\textbf{Seed1.5-VL}\\{\scriptsize non-thinking}} & \textbf{Prior SOTA} \\
    \midrule
    \multirow{6}{*}{\Shortunderstack[l]{Short video}} & MotionBench~\cite{hong2025motionbench} & {\bf 68.4} & {\bf 68.4} & \Shortunderstack[l]{62.8\\{\scriptsize GLM-4V}} \\
    & MVBench~\cite{li2024mvbench} & 74.4 & 74.3 & \Shortunderstack[l]{{\bf 76.4}\\{\scriptsize InternVL-2.5}}\\
    & TOMATO~\cite{shangguan2024tomato} & 44.7 & 44.2 & \Shortunderstack[l]{$\mathbf{46.9}^{*}$\\{\scriptsize Gemini 2.5 Pro}}\\
    & TVBench~\cite{cores2024tvbench} & {\bf 63.6} & 61.5 & \Shortunderstack[l]{$62.6^{*}$\\{\scriptsize Gemini 2.5 Pro}}\\
    & Dream-1K~\cite{wang2024tarsierrecipestrainingevaluating} & {\bf 43.9} & 42.6 & \Shortunderstack[l]{$42.0$\\{\scriptsize Tarsier2}}\\
    & TempCompass~\cite{liu2024tempcompass} & {\bf 83.7} & 83.1 & \Shortunderstack[l]{$75.8^{*}$\\{\scriptsize Gemini 2.5 Pro}} \\
    \midrule
    \multirow{6}{*}{\Shortunderstack[l]{Long video}} & LongVideoBench~\cite{wu2024longvideobench} & 74.0 & {\bf 74.4} & \Shortunderstack[l]{66.7\\{\scriptsize GPT-4o}} \\
    & LVBench~\cite{wang2024lvbench} & 64.6 & 64.0 & \Shortunderstack[l]{$\mathbf{69.2}^*$\\{\scriptsize Gemini 2.5 Pro}} \\
    & MLVU~\cite{zhou2024mlvu} & {\bf 82.1} & 81.8 & \Shortunderstack[l]{$81.2^{*}$\\{\scriptsize Gemini 2.5 Pro}} \\
    & \Shortunderstack[l]{VideoMME{\scriptsize (w/o sub)}~\cite{fu2024video}} & 77.9 & 77.6 & \Shortunderstack[l]{$\mathbf{87.0}^*$\\{\scriptsize Gemini 2.5 Pro}} \\
    & TemporalBench~\cite{cai2024temporalbench} & {\bf 79.8} & 78.9 & \Shortunderstack[l]{$73.3$\\{\scriptsize GPT-4o}} \\
    \midrule
    \multirow{4}{*}{\Shortunderstack[l]{Streaming video}} & OVBench~\cite{ovbench} & {\bf 60.0}  & 59.6  & \Shortunderstack[l]{$54.9$\\{\scriptsize PMB~\cite{ovbench}}}\\
    & OVOBench~\cite{li2025ovo} & {\bf 72.3}  & 72.0 & \Shortunderstack[l]{$67.7$\\{\scriptsize Gemini1.5-Pro}}\\
    & StreamBench~\cite{xiong2025streaming} & {\bf 72.8}  & 71.2  & \Shortunderstack[l]{$68.7$\\{\scriptsize GPT-4o}}\\
    & \Shortunderstack[l]{StreamingBench{\scriptsize (proactive)}~\cite{lin2024streamingbench}} & 68.0  & {\bf 82.8}  & \Shortunderstack[l]{$64.7$\\{\scriptsize Claude 3.5 Sonnet}}\\
    \midrule
    \multirow{2}{*}{\Shortunderstack[l]{Video reasoning}} & Video-MMMU~\cite{hu2025video} &\bf{81.4} & 72.1 &\Shortunderstack[l]{76.7\\{\scriptsize Kimi-K1.6}} \\
    & MMVU~\cite{zhao2025mmvu} & 70.1 & 70.1 &  \Shortunderstack[l]{$\mathbf{75.8}^*$\\{\scriptsize Gemini 2.5 Pro}}\\
    \midrule
    \multirow{2}{*}{\Shortunderstack[l]{Video grounding$^\dagger$}} & Charades-STA~\cite{gao2017tall} & 64.0 & \bf{64.7} & \Shortunderstack[l]{$60.7$\\{\scriptsize SG-DETR~\cite{sg-detr}}} \\
    & TACoS~\cite{Rohrbach2014CoherentMV} & \bf{49.6} & 47.8 & \Shortunderstack[l]{$42.4$\\{\scriptsize SG-DETR~\cite{sg-detr}}} \\
    \bottomrule
\end{tabular}}
\begin{tablenotes}
    \footnotesize
    \item[*] Results self-collected via API in April 2025.
    \item[$\dagger$] We adopt mIoU as the main metric for video grounding tasks.
    
\end{tablenotes}
\end{threeparttable}
\caption{\doubao{} performance on public video benchmarks compared to previous models. For all benchmarks, higher numbers are better. The evaluation frame rates are 2 FPS for MotionBench, MVBench, TOMATO, and TVBench, 3 FPS for Dream-1K, and 1 FPS for all other datasets.}
\label{tab:video_eval}
\end{table}
\renewcommand{\arraystretch}{1}
\subsection{Multimodal Agent}
\label{sec:agent_eval}
Multimodal agents are systems that perceive the world through visual inputs, understand instructions in natural language, and take actions to complete tasks. Two key scenarios for evaluating such agents are GUI interaction and gameplay, which test real-world usability and complex reasoning.
GUI agents simulate human-computer interaction by perceiving and acting on screen interfaces across desktops, browsers, and mobile devices. These tasks require precise visual grounding and multi-step execution.
Game agents operate in visually rich and interactive environments, requiring strategic planning, real-time decision-making, and commonsense reasoning.
We benchmark \doubao{} across both domains—GUI operation and gameplay—using a diverse set of evaluations. Results are shown in \cref{tab:gui_eval,tab:game_eval}, where we report \doubao{}'s performance under the thinking mode.

\begin{table}[htbp]
\centering
\small
\setlength{\tabcolsep}{1.2mm}{
\begin{tabular}{llllllll} 
\toprule
\multirow{2}{*}{\textbf{Capability}} & \multirow{2}{*}{\textbf{Benchmark}} & \textbf{Seed} & \textbf{OpenAI} & \textbf{Claude} & \textbf{UI-TARS} & \textbf{Kimi} & \textbf{Qwen 2.5} \\ 
& & \textbf{1.5-VL} & \textbf{CUA}~\cite{openaioperator} & \textbf{3.7 Sonnet}~\cite{claude37extendthink} & \textbf{1.5}~\cite{seed2025uitars15} & \textbf{VL-A3B}~\cite{team2025kimi} & \textbf{VL 72B}~\cite{bai2025qwen2} \\  
\midrule
\multirow{2}{*}{\Shortunderstack[l]{GUI\\Grounding}} 
& \Shortunderstack[l]{{ScreenSpot-V2}~\cite{wu2024atlas}} & {\bf 95.2} & $87.9$ & $87.6$ & \underline{$94.2$} & $92.8$ & - \\ 
& \Shortunderstack[l]{{ScreenSpot-Pro}~\cite{li2025screenspot}} & \underline{60.9} & $23.4$ & $27.7$ & {\bf 61.6} & $34.5$ &  $43.6$ \\ 
\midrule
\multirow{2}{*}{\Shortunderstack[l]{Computer\\Use}} 
& \Shortunderstack[l]{{OSWorld}~\cite{xie2024osworld}} & $36.7$ & \underline{$38.1$} & $28.0$ & {\bf 42.5} & $8.2$ & $8.8$ \\ 
& \Shortunderstack[l]{{Windows Agent Arena}~\cite{bonatti2024windows}} & \underline{39.6} & - & 38.9 & {\bf 42.1} & $10.4$ & - \\ 
\midrule
\multirow{2}{*}{\Shortunderstack[l]{Browser\\Use}} 
& \Shortunderstack[l]{{WebVoyager}~\cite{he2024webvoyager}} & {\bf 87.2} & \underline{$87.0$} & $84.1$ & $84.8$ & - & - \\ 
& \Shortunderstack[l]{{Online-Mind2Web}~\cite{xue2025illusion}} & {\bf 76.4} & $71.0$ & $62.9$ & \underline{$75.8$} & - & - \\ 
\midrule
\Shortunderstack[l]{Phone\\Use} 
& \Shortunderstack[l]{{Android World}~\cite{rawles2024androidworld}} & \underline{62.1} & - & - & {\bf 64.2} & - & \underline{$35.0$} \\ 
\bottomrule
\end{tabular}}
\caption{\doubao{} performance on public GUI online benchmarks compared to previous models.}
\label{tab:gui_eval}
\end{table}


\paragraph{GUI Grounding}
GUI grounding refers to the model’s ability to understand and localize interface elements—a fundamental skill for vision-based agents. We evaluate this capability on ScreenSpot Pro~\citep{li2025screenspot}, which focuses on expert-annotated tasks in professional settings, and ScreenSpot v2~\citep{wu2024atlas}, which covers grounding across desktop, mobile, and web interfaces. \doubao{} demonstrates strong grounding performance, achieving $60.9$ on ScreenSpot Pro and $95.2$ on ScreenSpot v2, which outperforms both \openaicua{} and \claude. As the foundation of multimodal interaction, GUI grounding enables agents to perceive actionable elements and bridge perception with control.

\paragraph{GUI Agent}
For GUI agent capability evaluation, we compare \doubao{} with strong baselines such as \openaicua{}~\citep{openaioperator} and \claude~\citep{claude37extendthink} on different GUI scenarios covering computer use, browser use, and phone use. As illustrated in \cref{tab:gui_eval}, \doubao{} consistently outperforms previous models on several key benchmarks. For instance, on OSWorld~\cite{xie2024osworld} and Windows Agent Arena~\cite{bonatti2024windows}, \doubao{} achieves 36.7\% and 39.6\%, respectively, surpassing \claude{}'s 28.0\% and 38.9\%. In browser use, \doubao{} scores 87.2\% on WebVoyager~\cite{he2024webvoyager} and 76.4\% on Online-Mind2Web~\cite{xue2025illusion}, outperforming \openaicua{} and \claude{}, setting new state-of-the-art results. On AndroidWorld~\cite{rawles2024androidworld}, a challenging mobile interface task, \doubao{} also achieves a high score of 62.1\%. Overall, among all the foundation VLMs (i.e., \claude{}, Kimi VL-A3B, and Qwen 2.5-VL), \doubao{} achieves significantly better performance in GUI agent tasks. These results underscore \doubao{}'s exceptional capabilities in executing GUI tasks and its strong generalization across diverse environments and devices, firmly establishing it as a premier position in GUI domain.

\paragraph{Game Agent}
\renewcommand{\arraystretch}{1.2}
\begin{table}[!t]
\centering
\small
\begin{tabular}{lllll} 
\toprule
\textbf{Game} 
  & \textbf{\doubao{}} 
  & \textbf{UI-TARS-1.5} 
  & \textbf{OpenAI CUA} 
  & \textbf{Claude 3.7 Sonnet} \\ 
\midrule
\Shortunderstack[l]{\href{https://poki.com/en/g/2048}{2048}\\{\scriptsize (score)}}               & \textbf{870.6} & 721.3   & 611.2 & 800.0 \\ \addlinespace
\Shortunderstack[l]{\href{https://poki.com/en/g/cubinko}{Cubinko}\\{\scriptsize (level)}}        & \textbf{2.0}   &   0.0   &   0.0 &   0.0 \\ \addlinespace
\Shortunderstack[l]{\href{https://poki.com/en/g/energy}{Energy} \\{\scriptsize (level)}}         & \textbf{2.3}   &   1.8   &   0.8 &   1.0 \\ \addlinespace
\Shortunderstack[l]{\href{https://poki.com/en/g/free-the-key}{Free-The-Key}  \\{\scriptsize (level)}} & \textbf{1.0}   &   0.0   &   0.0 &   0.0 \\ \addlinespace
\Shortunderstack[l]{\href{https://poki.com/en/g/gem-11}{Gem-11}\\{\scriptsize (score)}}         & \textbf{35.1}  &  10.8   &   8.7 &   0.0 \\ \addlinespace
\Shortunderstack[l]{\href{https://poki.com/en/g/hex-frvr}{Hex-Frvr}\\{\scriptsize (score)}}        & 1414.0 & \textbf{1583.7}  & 651.6 & 523.1 \\ \addlinespace
\Shortunderstack[l]{\href{https://poki.com/en/g/infinity-loop}{Infinity-Loop} \\{\scriptsize (level)}} & \textbf{1.4}   &   0.7   &   0.4 &   0.1 \\ \addlinespace
\Shortunderstack[l]{\href{https://poki.com/en/g/laser-maze-puzzle}{Laser-Maze-Puzzle} \\{\scriptsize (level)}} & \textbf{2.6}   &   2.2   &   1.4 &   1.4 \\ \addlinespace
\Shortunderstack[l]{\href{https://poki.com/en/g/maze-path-of-light}{Maze:Path-of-Light}\\{\scriptsize (level)}} & \textbf{1.3}   &   0.3   &   0.3 &   0.8 \\ \addlinespace
\Shortunderstack[l]{\href{https://poki.com/en/g/shapes}{Shapes}   \\{\scriptsize (level)}}       & \textbf{2.2}   &   1.5   &   0.9 &   0.2 \\ \addlinespace
\Shortunderstack[l]{\href{https://poki.com/en/g/snake-solver}{Snake-Solver}\\{\scriptsize (level)}}  & \textbf{1.3}   &   0.2   &   0.2 &   0.2 \\ \addlinespace
\Shortunderstack[l]{\href{https://poki.com/en/g/tiles-master}{Tiles-Master}  \\{\scriptsize (level)}}  & \textbf{2.3}   &   1.7   &   1.5 &   1.6 \\ \addlinespace
\Shortunderstack[l]{\href{https://poki.com/en/g/wood-blocks-3d}{Wood-Blocks-3d}\\{\scriptsize (score)}}   & \textbf{864.0} & 213.3   &  18.1 &   0.0 \\ \addlinespace
\Shortunderstack[l]{\href{https://poki.com/en/g/yarn-untangle}{Yarn-Untangle}  \\{\scriptsize (level)}} & \textbf{6.0}   &   5.7   &   5.1 &   1.6 \\
\bottomrule
\end{tabular}
\caption{\doubao{} performance on 14 Poki games with scores or levels completed. Models are evaluated over multiple runs, allowing up to 100 steps. For all games,
higher numbers are better.
}
\label{tab:game_eval}
\end{table}
\renewcommand{\arraystretch}{1}
Gameplay serves as a rigorous benchmark for multimodal models, combining visually rich environments with complex logic that challenges models to handle intricate reasoning, sequential decision-making, and rapid adaptation. Success in gameplay depends on intuitive commonsense reasoning, long-term strategic planning, and the ability to adapt to dynamic challenges—making it an ideal testbed for showcasing the advanced cognitive capabilities of state-of-the-art multimodal agents.

\begin{figure}[!t]
\centering
\includegraphics[width=0.85\textwidth]{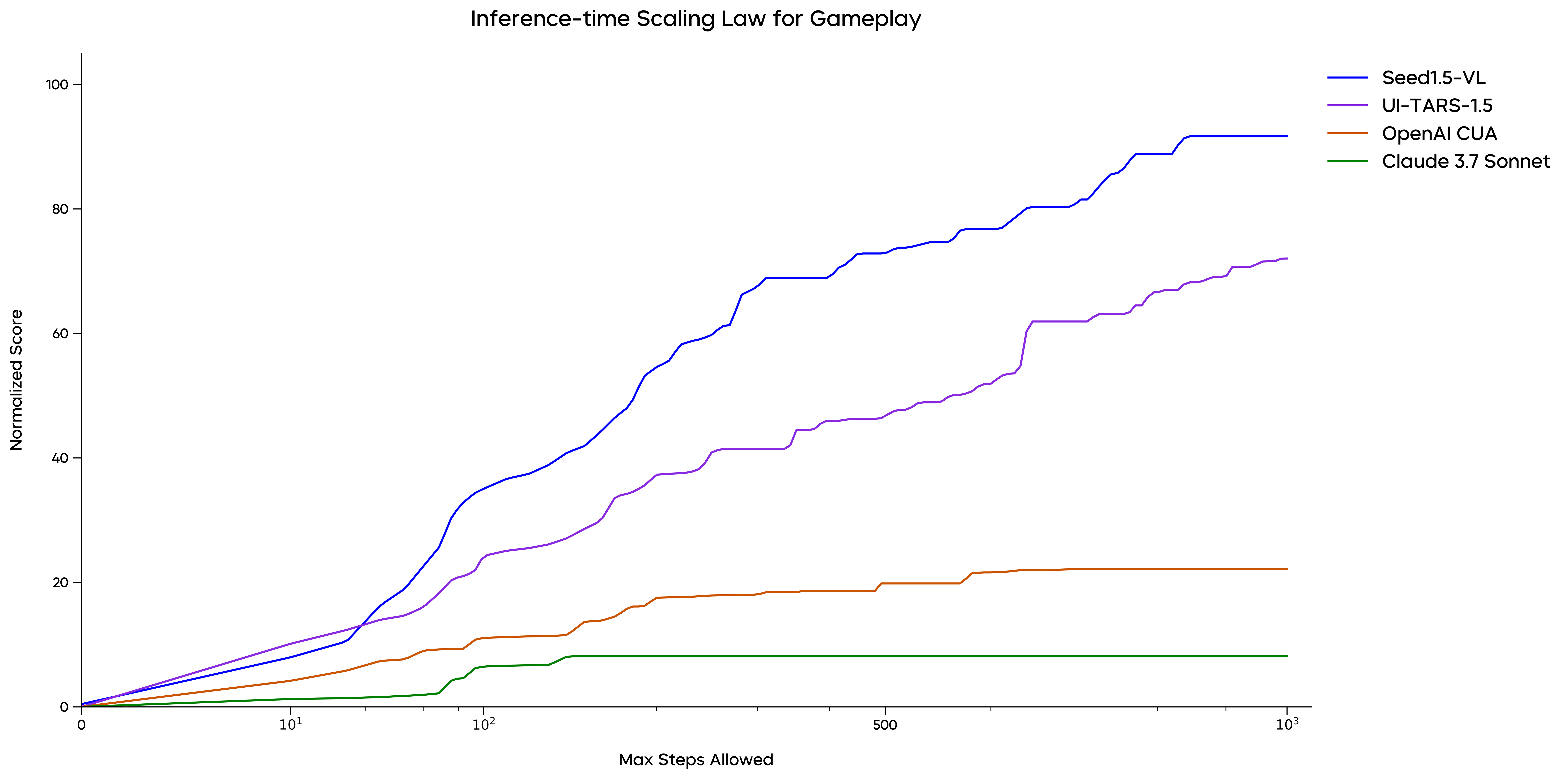}
\caption{For each game, we compute a scaling curve per model using normalized reference scores, and averaged them to produce an overall inference-time scaling trend.}
\label{fig:game_inference_scaling}
\end{figure}

We assemble a benchmark of 14 diverse games from Poki.com\footnote{\url{https://poki.com}}, which assess \doubao{}'s abilities in grounding, perception, and reasoning. As shown in \cref{tab:game_eval}, \doubao{} outperforms previous models across multiple games. For example, \doubao{} achieves 870.6 in 2048, surpassing \openaicua{} (611.2) and \claude{} (800.0), and 1414.0 in Hex-Frvr, a considerable lead over \openaicua{} (651.6) and \claude{} (523.1). These results highlight \doubao{}'s exceptional performance in completing game levels and achieving high scores. In addition, the long-horizon nature of gameplay makes it particularly well-suited for evaluating inference-time scaling behaviors. As depicted in \cref{fig:game_inference_scaling}, \doubao{} demonstrates strong scalability, maintaining higher performance as interaction rounds increase. This showcases its robust design and advanced reasoning abilities, ensuring consistent improvement even as the complexity of tasks grows over time.

\subsection{Internal Benchmarks}
\label{sec:eval:internal}
Besides public benchmarks, we also build internal benchmarks to comprehensively evaluate our models. We present motivation and design principles of our internal benchmarks in \cref{sec:eval:internal-design}, show results in \cref{sec:eval:internal-res}, and demonstrate model's Out-of-distribution (OOD) generalization ability in \cref{sec:ood}.

\subsubsection{Motivation and Design Principles}~\label{sec:eval:internal-design}

In addition to leveraging public benchmarks for exhaustive evaluation, we developed an internal benchmark suite to address several limitations inherent in existing resources. First, the predominance of English in public benchmarks necessitated the creation of comprehensive benchmarks to evaluate model performance specifically in Chinese, aligning with operational requirements. Second, the rapid pace of progress in multimodal research has resulted in saturation on many public benchmarks, reducing their sensitivity to incremental model improvements and hindering effective differentiation among leading models. Finally, limitations associated with the prevalent rule-based evaluation methods in public datasets, including challenges in answer parsing and potential data quality issues like label errors, underscored the need for tailored internal benchmarks with potentially more robust evaluation protocols and curated data.

Consequently, we developed our in-house benchmarks guided by several core principles:
\begin{itemize}
    \item \textbf{Focus on Core Capabilities over User Alignment}: The benchmarks prioritize assessing fundamental model abilities (e.g., perception, reasoning) rather than superficial alignment characteristics, such as preferences for response verbosity. This approach minimizes the confounding influence of alignment tuning on the evaluation of iterative model improvements.
    \item \textbf{Comprehensive Scope (Atomic and Integrated Capabilities)}: The evaluation suite encompasses assessments of both specific, atomic capabilities (\eg, fine-grained visual recognition) and complex, integrated multimodal tasks spanning diverse application domains.
    \item \textbf{Evaluation Accuracy and Methodology}: We employ Large Language Models (LLMs) as judges, advancing beyond traditional rule-based metrics. The prompts and reference answers utilized by these ``evaluator'' models undergo continuous refinement to ensure high evaluation fidelity. Current evaluator accuracy averages above 95\% for multiple-choice or simple-answer questions (\eg, single word/number responses) and exceeds 90\% for open-ended questions (further details in \cref{sec:app:internal_benchmark}).
    \item \textbf{Mitigation of Benchmark Overfitting}: To prevent inflated performance scores resulting from model overfitting to the benchmark data, we implement a rigorous data deduplication pipeline. Furthermore, task types and data sources within the benchmarks are periodically refreshed.
    \item \textbf{Task and Input Diversity}: Recognizing the critical role of diversity for VLMs, our benchmarks emphasize variety in both task types and input images. Image sourcing prioritizes non-publicly crawled data when feasible. We structure the benchmarks across numerous distinct dimensions, resulting in over 100 tasks and more than 12,000 samples from varied sources and domains. This includes a dedicated Out-of-Distribution (OOD) category featuring unconventional tasks designed to probe model generalization capabilities. A detailed taxonomy of targeted capabilities is provided in \cref{sec:app:internal_benchmark}.
\end{itemize}

\subsubsection{Comparison with State-of-the-arts}\label{sec:eval:internal-res}
\begin{table}[h]
\centering
\footnotesize
\begin{threeparttable}
\setlength{\tabcolsep}{1.5mm}{
\begin{tabular}{lllc|llll} \toprule
    \multirow[t]{3}{*}{\textbf{Level-1}} & \multirow[t]{3}{*}{\textbf{Level-2}} & \textbf{Weight} & \textbf{Seed} & \textbf{Gemini} & \textbf{OpenAI} & \textbf{OpenAI} & \textbf{Claude} \\
    \textbf{Capabilities} & \textbf{Capabilities} & & \textbf{1.5-VL} & \textbf{2.5 Pro} & \textbf{o1} & \textbf{o4-mini} & \textbf{3.7 Sonnet}  \\
    & & & {\scriptsize thinking} & {\scriptsize thinking} & {\scriptsize thinking} & {\scriptsize w/o tool use} & {\scriptsize thinking}  \\
    \midrule
    Overall & & 1.0 & \underline{59.3} & $\mathbf{61.6}$ & 54.0 & 55.4 & 48.6 \\
    \midrule 
    \multirow{4}{*}{\makecell[l]{Vision\\Capabilities}} & Perception & 0.1 & \underline{63.0} & {\bf 64.4} & 51.6 & 56.8 & 48.4 \\
    & Recognition & 0.1 & 72.4 & {\bf 74.8} & \underline{74.5} & 64.8 & 55.7 \\
    & OCR & 0.1 & \underline{67.2} & {\bf 70.7} & 55.7 & 64.4 & 57.1 \\
    & Caption \& Counterfactual & 0.05 & \underline{47.7} & {\bf 54.9} & 43.6 & 27.6 & 34.1 \\
    \midrule
    \multirow{9}{*}{\makecell[l]{Integrated\\Capabilities}} & OOD & 0.15 & $\mathbf{44.1}$ & \underline{43.1} & 42.3 & 38.4 & 35.9 \\
    & STEM & 0.04 & \underline{63.3} & $\mathbf{64.0}$ & 56.1 & 55.0 & 45.2 \\
    & Knowledge & 0.06 & 64.9 & $\mathbf{73.6}$ & \underline{68.5} & 57.8 & 50.8 \\
    & Reasoning & 0.1 & 47.6 & \underline{52.4} & 44.9 & {\bf 57.4} & 39.6 \\
    & Document \& Diagram Understanding & 0.1 & \underline{73.1} & {\bf 75.5} & 66.3 & 70.9 & 64.7 \\
    & Agent & 0.1 & {\bf 63.1} & {\bf 63.1} & 53.2 & 52.9 & 53.2 \\
    & Atomic Instruction Following & 0.03 & {\bf 69.6} & \underline{69.2} & 63.8 & 68.7 & 50.5 \\
    & Code & 0.05 & 44.0 & 43.7 & 39.9 & {\bf 60.6} & \underline{54.6} \\
    & ToB & 0.02 & \underline{47.1} & {\bf 54.7} & 30.2 & 39.8 & 29.1 \\
    \bottomrule
\end{tabular}}
\end{threeparttable}
\caption{Evaluation results comparing \doubao{} and state-of-the-art models on the internal benchmark. The overall score is calculated as a weighted average across performance in defined sub-categories. Data for other models was sourced via API access in April 2025. Weights for averaging are set for minimizing variance of evaluation and highlighting the importance of each category. The highest scores are marked in \textbf{bold} and the second is \underline{underlined}.}
\label{tab:internal_eval}
\end{table}
We compare \doubao{} with leading industry models (Gemini 2.5 Pro, OpenAI o1, OpenAI o4-mini, Claude 3.7) in \cref{tab:internal_eval} under \textit{thinking mode}. The leading score of 61.6 (Gemini 2.5 Pro) highlights substantial room for improvement on this benchmark, unlike many public benchmarks nearing saturation above 80 in \cref{tab:image_eval_combined}. A more comprehensive comparison including \textit{non-thinking} models can be found in \cref{sec:app:complete_comparison}.

\doubao{} achieves the second-highest overall score. It achieves state-of-the-art performance in OOD, Agent, Atomic Instruction Following categories, and shows strong capabilities in STEM and Document \& Diagram Understanding. Its primary weaknesses relative to the top performer are observed in knowledge, reasoning, code, and captioning/counterfactual tasks. We attribute this gap partly to the scale of the current model, which utilizes a language model with approximately 20B active parameters. Evidence supporting potential gains from further scaling is presented in \cref{tab:scaling_laws}, where the training loss shows no sign of saturation after 3 trillion tokens, and evaluation metrics correlate strongly with loss. Therefore, we expect the performance gap to diminish as we increase the model size and the training compute.

Grouping models strictly by parameter count is challenging due to the lack of public disclosure of specific parameter details for many models. Our model's size is comparable to the recently released Llama 4 Maverick~\cite{meta2025llama}, which is reported to utilize 17 billion active parameters and employs a Mixture-of-Experts (MoE) architecture. Our evaluation demonstrates that \doubao{} achieves significantly better performance than Llama 4 Maverick on this benchmark (\cref{fig:internal_complete_comparison}).

\subsubsection{Out-of-distribution Generalization}
\label{sec:ood}
\begin{figure}[h]
  \centering
  \begin{tabular}{m{4cm}m{12cm}}
  \toprule
  \includegraphics[width=4cm]{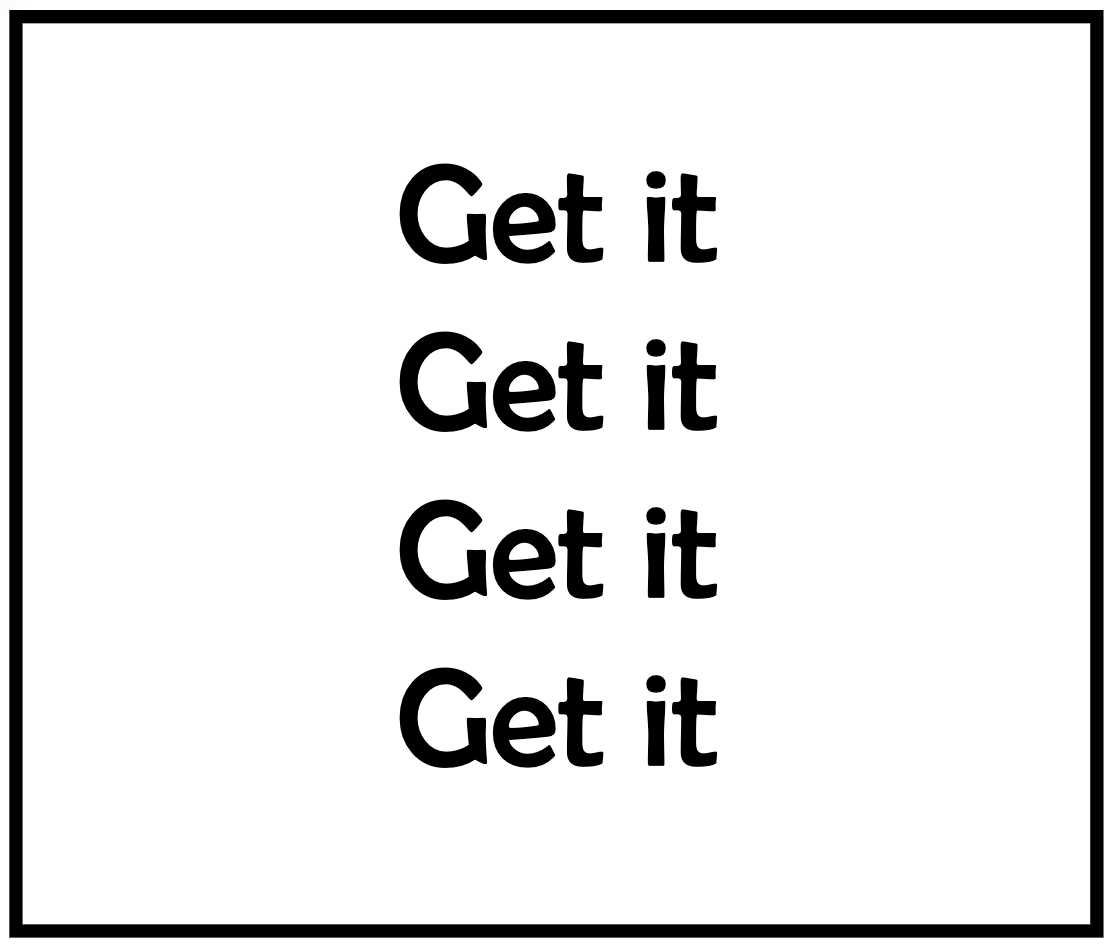} & 
  {\textbf{Prompts:} can you solve this Rebus puzzle?\newline\newline
  \textbf{Model Responses:} \texttt{<think>...</think>}The Rebus puzzle features the phrase ``Get it'' repeated four times. The number of repetitions (four) combined with the phrase "Get it" creates a homophonic play: \textbf{``Four get it''} sounds like \textbf{``Forget it''}. This wordplay uses the repetition count (four) to form the first part of the phrase, which phonetically matches ``forget''. Thus, the solution is \textbf{Forget it}.} \\
  \bottomrule
  \end{tabular}
\caption{To solve this Rebus puzzle, our model has to correctly recognize characters in the image, count the number of repetitions, retrieve knowledge about Rebus puzzles, and perform reasoning to obtain the final answer. Thinking tokens are omitted for brevity.}
\label{fig:rebus_forgetit}
\end{figure}

\begin{figure}[h]
  \centering
  \begin{tabular}{m{7cm}m{9cm}}
  \toprule
  \includegraphics[width=7cm]{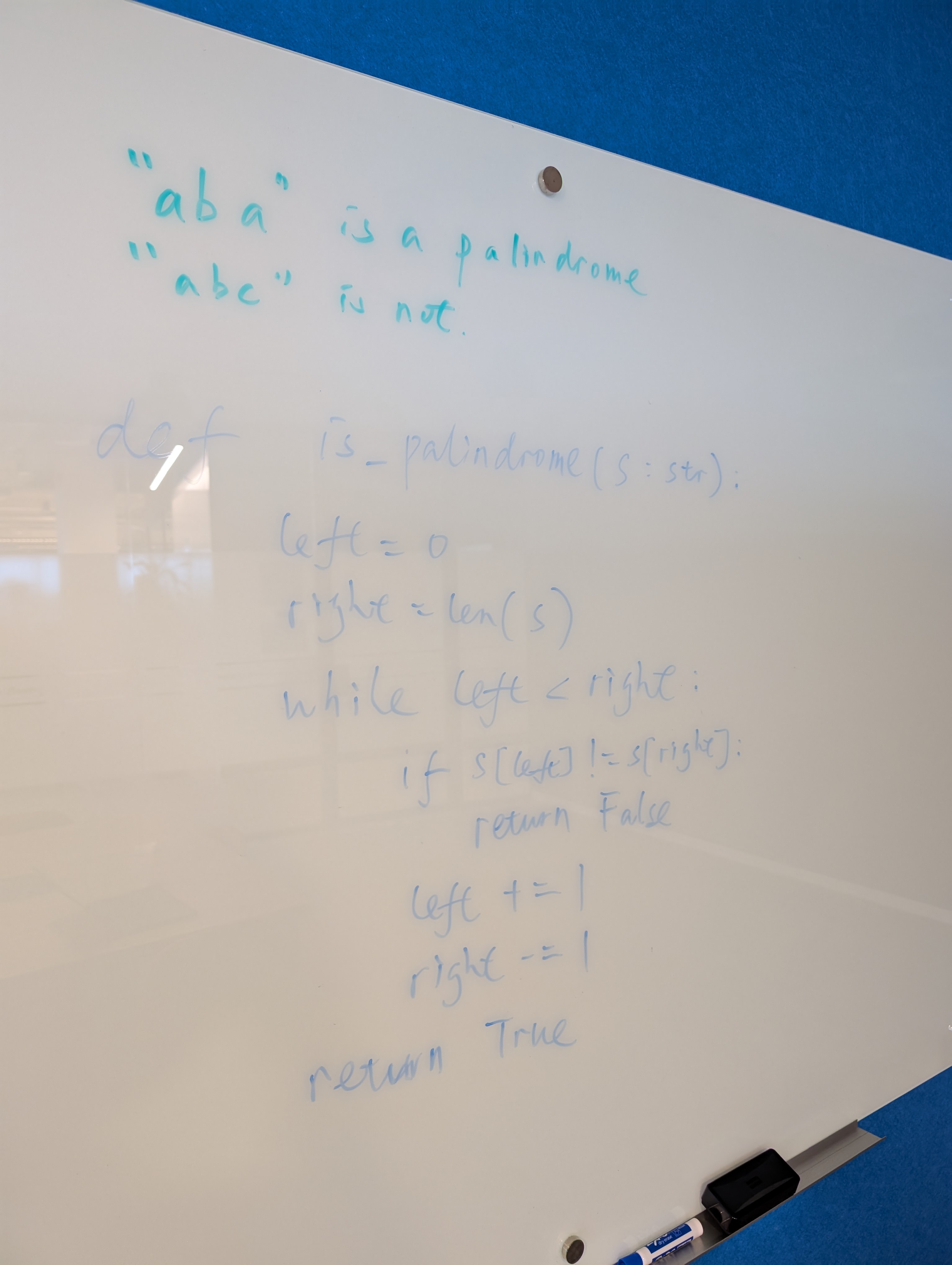} & 
  {\footnotesize {\bf Prompts:} There is a bug in the code on the whiteboard. Can you spot it and return the correct code?\par
  {\bf Model Responses:} \texttt{<think>...</think>}The bug is in the initialization of the right pointer. It should be \texttt{len(s) - 1} (since Python uses zero-based indexing, the last character is at index \texttt{len(s)-1}, not \texttt{len(s)}). Here's the corrected code:\par
\begin{Verbatim}[breaklines=true,breaksymbol=]
def is_palindrome(s: str) -> bool:
    left = 0
    right = len(s) - 1  # Fixed: use last index (len(s)-1) instead of len(s)
    while left < right:
        if s[left] != s[right]:
            return False
        left += 1
        right -= 1
    return True
\end{Verbatim}
\par
{\bf Explanation:}
\par
\begin{enumerate}
    \item {\bf Right Pointer Initialization}: \texttt{right = len(s) - 1} ensures we start at the last character (correct index), avoiding an \texttt{IndexError}.
    \item {\bf Loop Logic}: Compares characters from the start (\texttt{left}) and end (\texttt{right}), moving pointers inward. If any pair mismatches, returns \texttt{False}. If the loop completes (all pairs match), returns \texttt{True}.
\end{enumerate}
\par
{\bf Testing:}
\begin{itemize}
    \item \texttt{"aba"}: \texttt{left=0, right=2} $\rightarrow$ match (\texttt{'a' == 'a'}), then \texttt{left=1, right=1} $\rightarrow$ loop ends, return \texttt{True} (correct).
    \item \texttt{"abc"}: \texttt{left=0, right=2} $\rightarrow$ \texttt{'a' != 'c'} $\rightarrow$ return \texttt{False} (correct).
\end{itemize} 
\par
This fixes the index error and correctly implements the palindrome check using two pointers.
  } \\
  \bottomrule
  \end{tabular}
\caption{An example of \doubao{} correcting code written on a whiteboard using its OCR and coding capabilities.}
\label{fig:debug_code}
\end{figure}

As shown in \cref{tab:internal_eval}, our model demonstrates performance comparable to that of two leading industry models, Gemini Pro 2.5 and OpenAI o1, within the Out-of-Distribution (OOD) category of the benchmark. Complementary to standardized evaluations, an internal Chatbot platform was developed to assess the model's ability to integrate multiple atomic capabilities in complex real-world scenarios. Three representative examples are highlighted. First, the model successfully solves a Rebus puzzle shown in \cref{fig:rebus_forgetit}, leveraging its OCR, knowledge retrieval, and reasoning abilities. Additional examples, some of which may be challenging for humans, are provided in \cref{sec:app:rebus}. In the second example (\cref{fig:debug_code}), it processes code captured from a challenging visual input (a shiny whiteboard), analyzes and corrects errors, and validates the corrected code using user-provided examples. In the third example (\cref{fig:ncaa_bracket}), the model parses a diagram of an unseen format, performing ``inverse graphics'' to generate the underlying Mermaid code while adhering to a specific spatial constraint (``south region'').

A distinct benchmark, supplementing the internal evaluation, was employed to assess the new model's user preference alignment capability. This evaluation involved augmenting the model with search tools and serves as a proxy for its utility to the Doubao\footnote{https://www.doubao.com/chat/} user base. 
The usefulness rate for Seed 1.5VL is 62.6\%. Similarly, we have also tested Gemini 2.5 pro 0325, with the usefulness rates  reaching 57.4\%.

\subsection{Limitations}\label{sec:limitations}

Despite strong performance across many benchmarks, \doubao{} exhibits certain limitations, particularly in fine-grained visual perception and complex reasoning.

In complex visual perception tasks, \doubao{} struggles with accurately counting objects when they are irregularly arranged, similar in color, or partially occluded. Identifying subtle differences between images also presents a challenge, sometimes leading to missed details or inaccurate descriptions. Furthermore, similar to some contemporaries (e.g., \openaigpt{} and \gemini{}), difficulties can arise in precisely interpreting complex spatial relationships, especially with varying perspectives, and accurately responding to visual prompts, occasionally misidentifying content or quantity within specified regions or making localization errors.

Beyond perception, challenges also emerge in higher-level reasoning, as suggested by both open and internal benchmarks. \doubao{} demonstrates suboptimal performance on reasoning tasks trivial for humans, such as solving Klotski puzzles or navigating simple mazes, suggesting a need for future exploration into techniques like visual Chain-of-Thought (CoT) where the model might augment its process with auxiliary visual aids (\eg, lines).
Reasoning requiring combinatorial search poses a significant challenge for many existing VLMs. \Cref{fig:brute_force_search,fig:brute_force_search2} provide two illustrative examples of problems falling into this category. While challenging for current VLM architectures, combinatorial search tasks are often more readily addressed through programmatic or code-based approaches. Consequently, incorporating code-use and other external tools into VLM frameworks represents an important direction for future research aimed at enhancing such reasoning capabilities.

Limitations are observed in 3D spatial reasoning tasks for most VLMs. These challenges include, for example, tasks involving 3D object manipulation or reasoning about the projection of 3D objects. Specific instances illustrating such failure cases are provided in \cref{fig:failure_spatial_imagination}. A potential direction to address this limitation is the incorporation of image generation capabilities into the foundation model, which could further enable visual chain-of-thought mechanisms. This approach remains a subject for future research.

Additionally, VLMs sometimes still produce incorrect inferences, particularly in tasks such as visual puzzles. These reasoning errors may stem from underlying perceptual mistakes (misinterpreting shapes or conditions) or from limitations in the logical deduction process itself. In tasks requiring planning or adherence to complex instructions, \doubao{} may overlook specific conditions or introduce unfounded assumptions, which can lead to incomplete or invalid responses. 

Our internal evaluations also revealed model's deficiency in temporal reasoning capability, as the model faced difficulties in discerning the chronological sequence of continuous actions or inferring order from the before-and-after states of objects. The capacity for multi-image reasoning is limited, with performance degrading on tasks requiring the synthesis of clues across multiple images with strong logical interdependencies.

Finally, hallucination persists as a significant challenge for all VLMs. \Cref{fig:hallucination} illustrates a particularly notable case where the visual input conflicts with the prior knowledge acquired from the language model component. In such instances, models tend to prioritize this acquired knowledge, effectively overriding or conforming the perceived visual information to align with learned priors.
\section{Conclusion and Next Steps}
\label{sec:conclusion}
In this paper, we presented Seed1.5-VL, our latest multimodal foundation model demonstrating strong capabilities in reasoning, OCR, diagram understanding, visual grounding, 3D spatial understanding, and video understanding. Despite its relatively moderate size, Seed1.5-VL achieves state-of-the-art results on 38 out of 60 evaluated public benchmarks, including a score of 77.9 on the MMMU benchmark, widely regarded as a key indicator of multimodal reasoning ability.

Beyond benchmark performance, Seed1.5-VL exhibits significant integrated capabilities and generalization to tasks dissimilar to its training data. Examples include solving complex visual reasoning tasks such as Rebus puzzles, interpreting and correcting handwritten code from whiteboard images, and functioning as an agent for computer interaction and gameplay. Further exploration of these emergent abilities is warranted.

Our scaling analysis indicates that model performance shows no sign of saturation, suggesting that increasing model parameters and training compute represents a promising immediate direction. Through our evaluations, we also identified limitations common to contemporary VLMs, such as robust 3D spatial reasoning, hallucination mitigation, and complex combinatorial search. Addressing these challenges constitutes a core part of our ongoing research, which includes efforts towards unifying existing model capabilities with image generation (potentially enabling visual Chain-of-Thought) and incorporating robust tool-use mechanisms.

The advancements presented here build upon substantial prior work within the AI research community, leveraging foundational developments like the Transformer and Vision Transformer architectures. To contribute to future progress, we have detailed our model architecture, data synthesis pipeline, training methodology, training framework innovations, and internal evaluation design in this report.

\clearpage

\bibliographystyle{plainnat}
\bibliography{main}

\clearpage
\section{Contributions and Acknowledgments}
\label{sec:contributions}
The authors are listed in alphabetical order by their first names. Some names refer to the authors' internal aliases at the company.
\setlength{\parskip}{0pt} 
\setlength{\itemsep}{0pt} 
\setlength{\parsep}{0pt}  
\begin{multicols}{2}
\subsubsection*{Core Contributors}
Dong Guo\\
Faming Wu\\
Feida Zhu\\
Fuxing Leng\\
Guang Shi\\
Haobin Chen\\
Haoqi Fan\\
Jian Wang\\
Jianyu Jiang\\
Jiawei Wang\\
Jingji Chen\\
Jingjia Huang\\
Kang Lei\\
Liping Yuan\\
Lishu Luo\\
Pengfei Liu\\
Qinghao Ye\\
Rui Qian\\
Shen Yan\\
Shixiong Zhao\\
Shuai Peng\\
Shuangye Li\\
Sihang Yuan\\
Sijin Wu\\
Tianheng Cheng\\
Weiwei Liu\\
Wenqian Wang\\
Xianhan Zeng\\
Xiao Liu\\
Xiaobo Qin\\
Xiaohan Ding\\
Xiaojun Xiao\\
Xiaoying Zhang\\
Xuanwei Zhang\\
Xuehan Xiong\\
Yanghua Peng\\
Yangrui Chen\\
Yanwei Li\\
Yanxu Hu\\
Yi Lin\\
Yiyuan Hu\\
Yiyuan Zhang\\
Youbin Wu\\
Yu Li\\
Yudong Liu\\
Yue Ling\\
Yujia Qin\\
Zanbo Wang\\
Zhiwu He

\subsubsection*{Contributors}
Aoxue Zhang\\
Bairen Yi\\
Bencheng Liao\\
Can Huang\\
Can Zhang\\
Chaorui Deng\\
Chaoyi Deng\\
Cheng Lin\\
Cheng Yuan\\
Chenggang Li\\
Chenhui Gou\\
Chenwei Lou\\
Chengzhi Wei \\
Chundian Liu\\
Chunyuan Li\\
Deyao Zhu\\
Donghong Zhong\\
Feng Li\\
Feng Zhang\\
Gang Wu\\
Guodong Li\\
Guohong Xiao\\
Haibin Lin\\
Haihua Yang\\
Haoming Wang\\
Heng Ji\\
Hongxiang Hao\\
Hui Shen\\
Huixia Li\\
Jiahao Li\\
Jialong Wu\\
Jianhua Zhu\\
Jianpeng Jiao\\
Jiashi Feng\\
Jiaze Chen\\
Jianhui Duan\\
Jihao Liu\\
Jin Zeng\\
Jingqun Tang\\
Jingyu Sun\\
Joya Chen\\
Jun Long\\
Junda Feng\\
Junfeng Zhan\\
Junjie Fang\\
Junting Lu\\
Kai Hua\\
Kai Liu\\
Kai Shen\\
Kaiyuan Zhang\\
Ke Shen\\
Ke Wang\\
Keyu Pan\\
Kun Zhang\\
Kunchang Li\\
Lanxin Li\\
Lei Li\\
Lei Shi\\
Li Han\\
Liang Xiang\\
Liangqiang Chen\\
Lin Chen\\
Lin Li\\
Lin Yan\\
Liying Chi\\
Longxiang Liu\\
Mengfei Du\\
Mingxuan Wang\\
Ningxin Pan\\
Peibin Chen\\
Pengfei Chen\\
Pengfei Wu\\
Qingqing Yuan\\
Qingyao Shuai\\
Qiuyan Tao\\
Renjie Zheng\\
Renrui Zhang\\
Ru Zhang\\
Rui Wang\\
Rui Yang\\
Rui Zhao\\
Shaoqiang Xu\\
Shihao Liang\\
Shipeng Yan\\
Shu Zhong\\
Shuaishuai Cao\\
Shuangzhi Wu\\
Shufan Liu\\
Shuhan Chang\\
Songhua Cai\\
Tenglong Ao\\
Tianhao Yang\\
Tingting Zhang\\
Wanjun Zhong\\
Wei Jia\\
Wei Weng\\
Weihao Yu\\
Wenhao Huang\\
Wenjia Zhu\\
Wenli Yang\\
Wenzhi Wang\\
Xiang Long\\
XiangRui Yin\\
Xiao Li\\
Xiaolei Zhu\\
Xiaoying Jia\\
Xijin Zhang\\
Xin Liu\\
Xinchen Zhang\\
Xinyu Yang\\
Xiongcai Luo\\
Xiuli Chen\\
Xuantong Zhong\\
Xuefeng Xiao\\
Xujing Li\\
Yan Wu\\
Yawei Wen\\
Yifan Du\\
Yihao Zhang\\
Yining Ye\\
Yonghui Wu\\
Yu Liu\\
Yu Yue\\
Yufeng Zhou\\
Yufeng Yuan\\
Yuhang Xu\\
Yuhong Yang\\
Yun Zhang\\
Yunhao Fang\\
Yuntao Li\\
Yurui Ren\\
Yuwen Xiong\\
Zehua Hong\\
Zehua Wang\\
Zewei Sun\\
Zeyu Wang\\
Zhao Cai\\
Zhaoyue Zha\\
Zhecheng An\\
Zhehui Zhao\\
Zhengzhuo Xu\\
Zhipeng Chen\\
Zhiyong Wu\\
Zhuofan Zheng\\
Zihao Wang\\
Zilong Huang\\
Ziyu Zhu \\
Zuquan Song

\end{multicols}
\subsubsection*{Acknowledgments}
We would like to sincerely thank 
Allie Guo,
Bingyi Kang,
Borui Wan,
Chaoran Guo,
Chenyuan Wang,
Chixiang Ma,
Fei Xiong,
Fu Li,
Fuxiang Li,
Gaohong Liu,
Hongbin Ren,
Hongyu Zhu,
Jinxin Chi,
Juncai Liu,
Kaihua Jiang,
Kayden,
Lei Zuo,
Lianke Qin,
Lingjun Liu,
Liyang Liu,
Minchao Wang,
Mingji Han,
Mofan Zhang,
Pengyuan Zhao,
Qianli Ma,
Qifan Yang,
Qinlong Wang,
Shibiao Nong,
Tong Zhou,
Weiqiang Lou,
Xiangpeng Wei,
Xiangrui Yin,
Xiangtai Li,
Xiaokai Li,
Xudong Sun,
Xun Wang,
Yaowei Zheng,
Yixin Wu,
Yingping Zhang,
Yun Zhang,
Yuwen Tang,
Zhe Nan,
Zhelun Shi,
Zheng Zhong,
Zhenyuan Yang,
Zhi Zhang,
Zhongjia Wei,
Zhuolin Zheng,
Zilong Zhou,
Ziqian Wei,
Ziwen Xu,
Zixin Chen,
Ziyuan Feng,
Zuquan Song for their insightful discussions and unwavering support. Their valuable input has been instrumental in advancing \doubao{}'s development, evaluation, defect analysis, and exploration of future research directions.
\clearpage

\beginappendix

\section{Qualitative examples}
\label{sec:app:qualitative}
\begin{figure}[!h]
  \centering
  \begin{tabular}{m{16cm}}
  \toprule
  \begin{center}
  Input image
  \end{center}
  \begin{center}
  \includegraphics[width=0.5\linewidth]{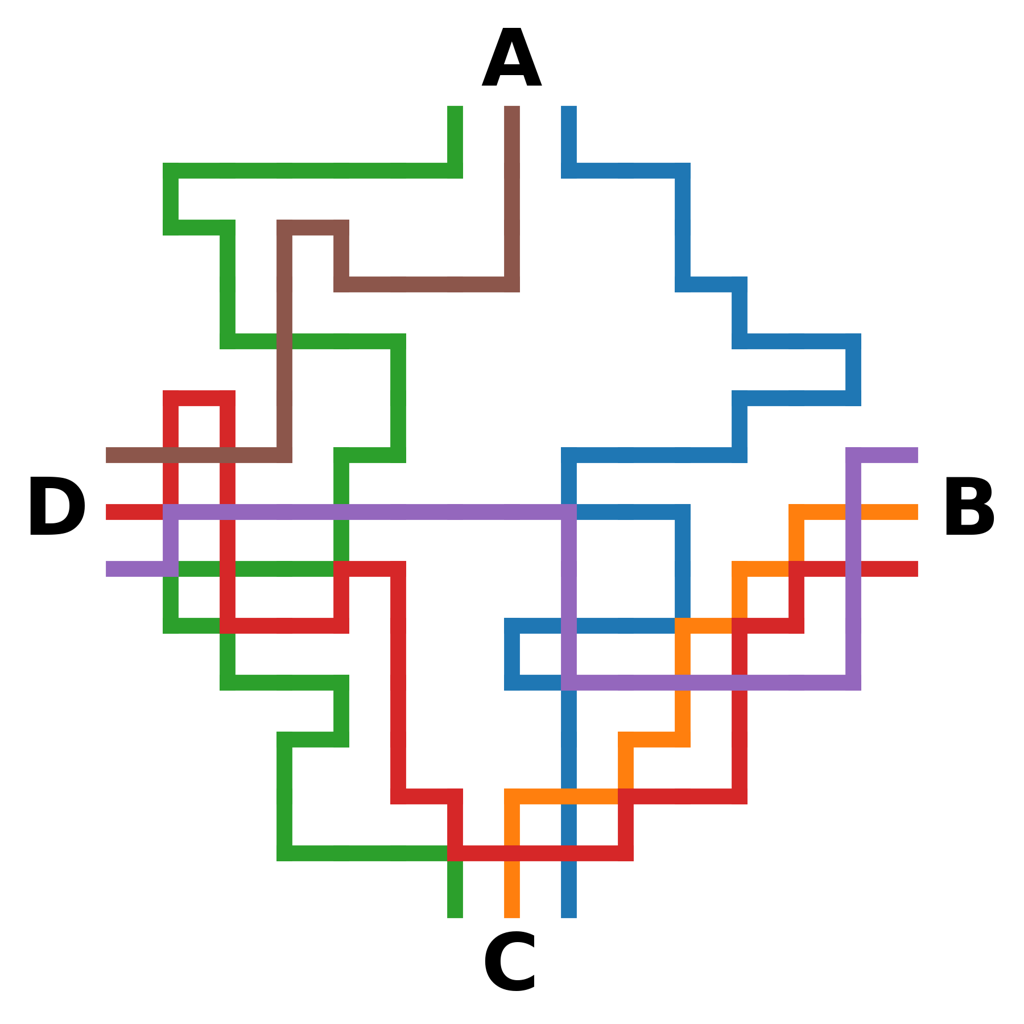}
  \end{center}
  \\
  {\footnotesize \textbf{Prompts:} Count the one-colored routes that go from C to A. Answer with a number in curly brackets e.g. \{3\}\newline\newline
  \textbf{Model Responses:} \texttt{<think>}So, let's try to figure out how many one-colored routes go from C to A. First, I need to identify each color and check if there's a continuous path from C to A for each color.\newline
  Let's list the colors: green, brown (or maybe dark red/brown), blue, red, orange, purple. Wait, let's check each color's path.\newline
  First, green: Let's see if there's a green path from C to A. Starting at C, follow green lines. Does it reach A? Let's trace. Green lines: from C, there's a green segment, then up, etc. Wait, need to check continuity.\newline
  Brown: The brown path. Does brown go from C to A? Let's check. Brown starts... Wait, A is at the top, so brown's path: does it connect C to A?\newline
  Blue: Blue path. From C, blue lines—do they go up to A?\newline
  Red: Red path. From C, red lines—do they reach A?\newline
  Orange: Orange path. From C, orange lines—do they reach A?\newline
  Purple: Purple path. From C, purple lines—do they reach A?
  \newline
  ... \newline
  \textcolor{red}{Let's look at the image again.}\newline
  Wait, the colors are:\newline
  - Green: several segments forming a path from C to A.\newline
  - Blue: several segments forming a path from C to A.\newline
  Are there any other colors? Let's check each color's continuity.\newline
  Green: Yes, continuous from C to A.\newline
  Blue: Yes, continuous from C to A.\newline
  Brown: Starts at A, goes down, but does it reach C? No, because the brown path ...\newline
  So that's two. So the answer is {2}?\texttt{</think>}\newline
  \{2\}}
  \\
  \bottomrule
  \end{tabular}
\caption{A case demonstrating one of the visual-centric thinking strategies that emerged during LongCoT RL.}
\label{fig:vlm_are_blind_case}
\end{figure}

\begin{figure}[!h]
  \centering
  \begin{tabular}{m{16cm}}
  \toprule
  \begin{center}
  Input image
  \end{center}
  \begin{center}
  \includegraphics[width=0.5\linewidth]{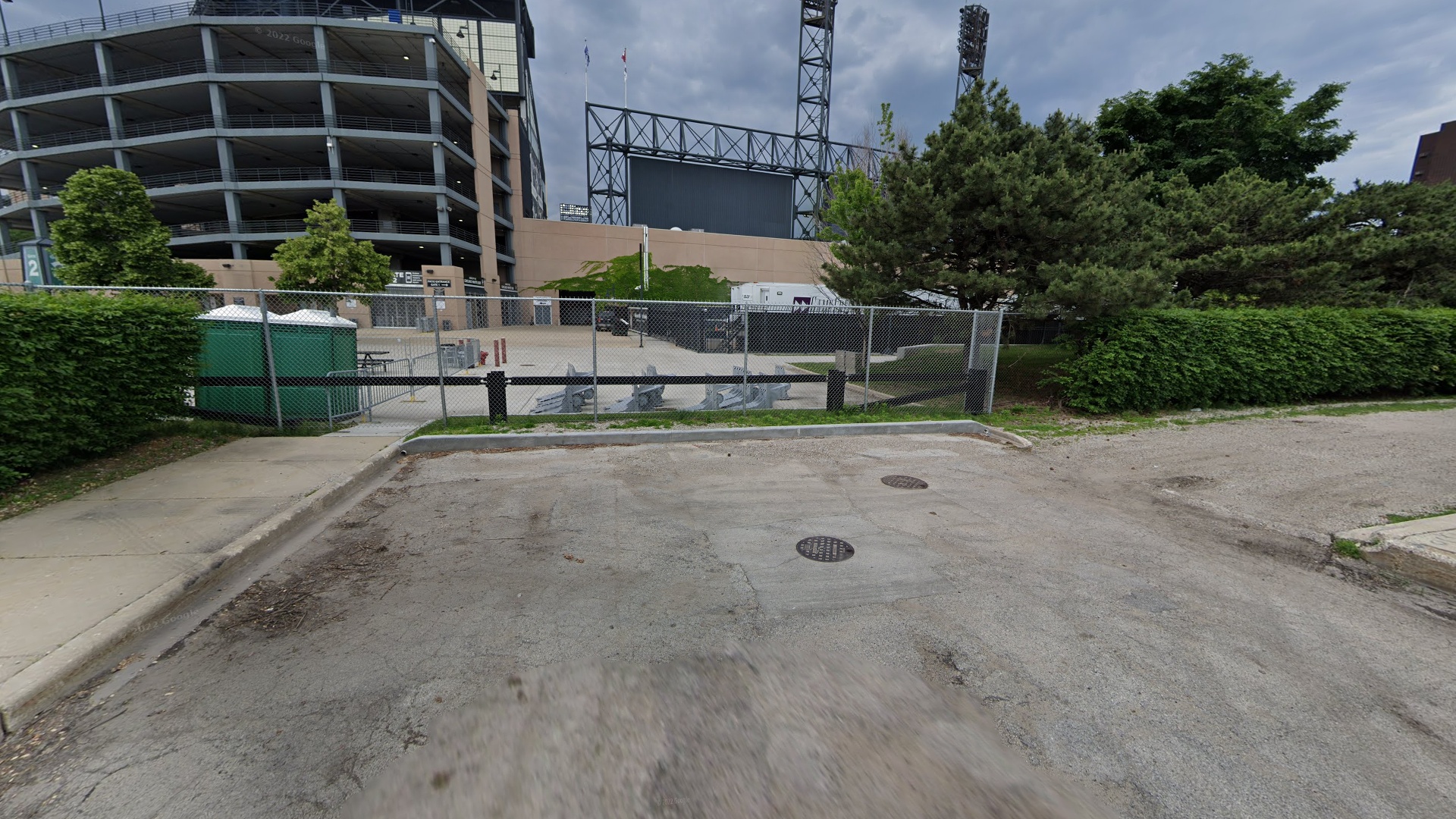}
  \end{center}
  \\
  {\footnotesize \textbf{Prompts:} You are participating in a geolocation challenge. Based on the provided image:\newline
1. Carefully analyze the image for clues about its location (architecture, signage, vegetation, terrain, etc.)\newline
2. Think step-by-step about what continent, country, and city this is likely to be in and why\newline
3. Estimate the approximate latitude and longitude based on your analysis
Take your time to reason through the evidence. Your final answer MUST include these five lines somewhere in your response:\newline
continent: [continent name]\newline
country: [country name]\newline
city: [city name]\newline
lat: [latitude as a decimal number]\newline
lng: [longitude as a decimal number]\newline
You can provide additional reasoning or explanation, but these three specific lines MUST be included.\newline\newline
  \textbf{Model Responses:} \texttt{<think>}So, let's try to figure out the location. First, look at the large building on the left—it looks like a multi-level parking structure or part of a stadium. The stadium-like structure with the big scoreboard and the tall light towers (used in baseball stadiums, maybe) is a clue. The number "2" on a sign might be a gate number. The vegetation: there are trees and hedges, typical of an urban area. \newline
  Now, the architecture style—modern stadium, maybe in the US. Let's think about major stadiums. The structure with the grid and the flags—maybe Guaranteed Rate Field (formerly U.S. Cellular Field), home of the Chicago White Sox. Let's check. The stadium's exterior, the parking structure, and the surrounding area. Chicago, Illinois, USA. \newline
  To confirm, the style of the stadium, the urban setting, and the type of infrastructure. Then, latitude and longitude for Guaranteed Rate Field: approximately 41.8316° N, 87.6348° W. ... \texttt{</think>}
  \newline\newline
  ...\newline
  continent: North America \newline
  country: United States \newline
  city: Chicago \newline
  lat: 41.8316 \newline
  lng: -87.6348}
  \\
  \bottomrule
  \end{tabular}
\caption{A case demonstrating the "analyze visual details before recognizing a location" strategy that emerged during LongCoT RL.}
\label{fig:geoguess_case}
\end{figure}

\begin{figure}[h]
  \centering
  \begin{tabular}{m{4cm}m{12cm}}
  \toprule
  \includegraphics[width=4cm]{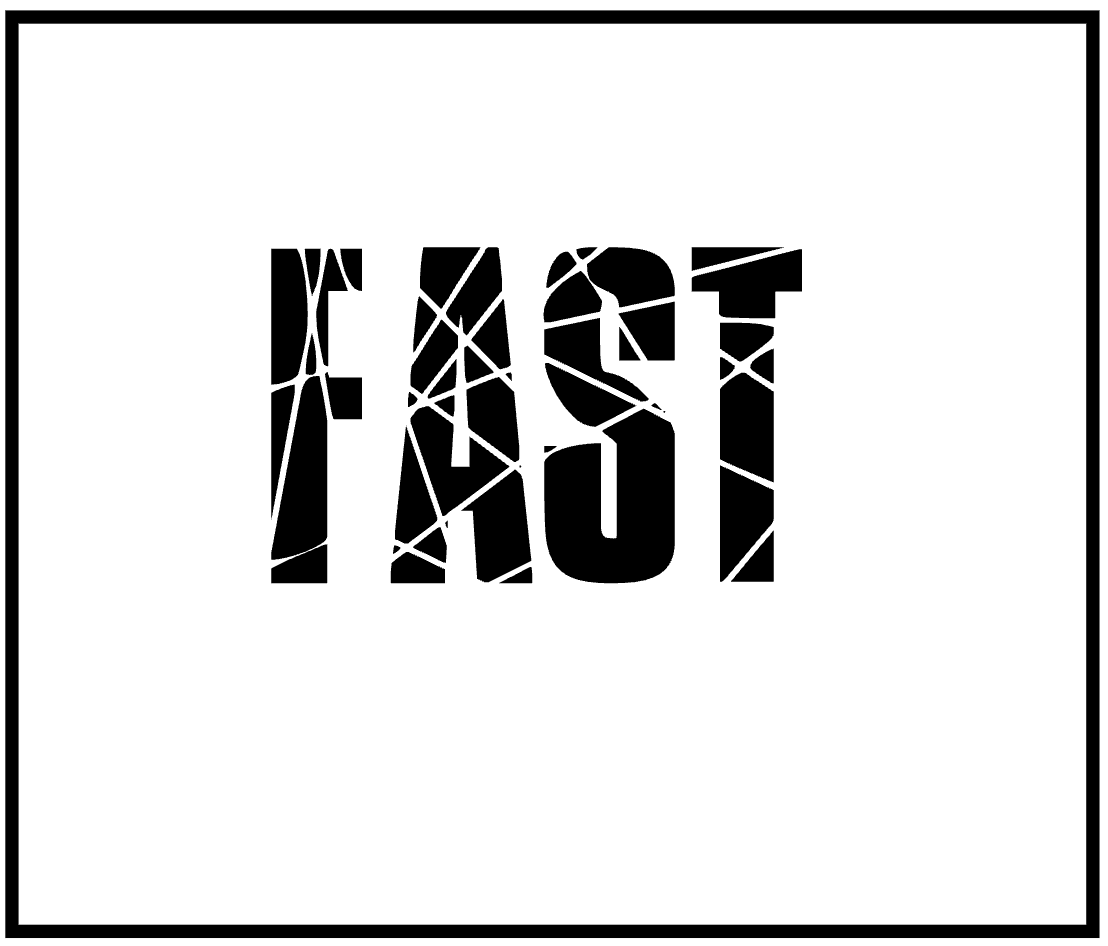} & 
  {\textbf{Prompts:} can you solve this Rebus puzzle?\par \textbf{Model Responses:} \texttt{<think>...</think>}The Rebus puzzle features the word ``FAST'' with cracked lines, suggesting it is being ``broken.'' Combining ``break'' (from the visual of cracks, implying breaking) with ``fast'' results in the word \textbf{breakfast}} \\
  \includegraphics[width=4cm]{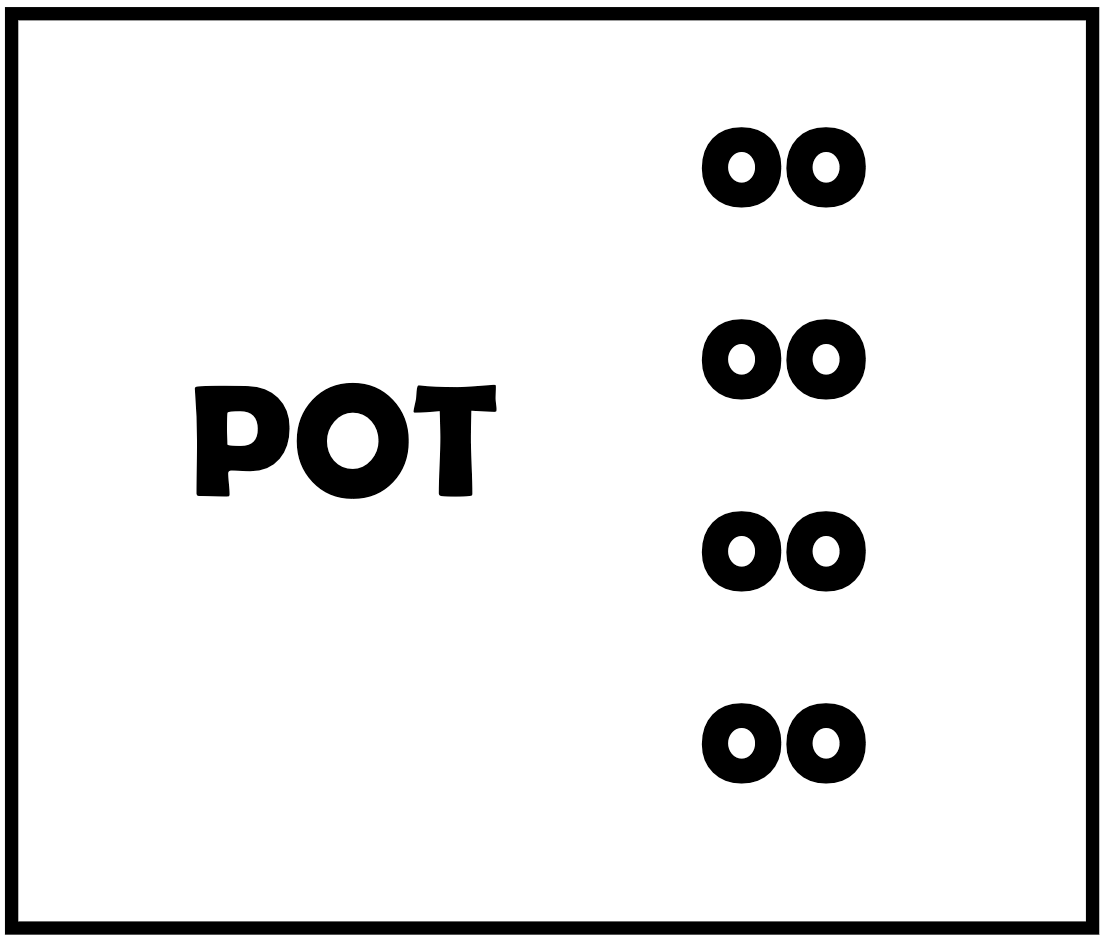} &
  {\textbf{Prompts:} can you solve this Rebus puzzle?\par \textbf{Model Responses:} \texttt{<think>...</think>}The Rebus puzzle features the word ``POT'' on the left and eight circles (four pairs of two circles each, resembling ``OO'') on the right. Combining ``POT'' with ``8 O's'' (where ``8 O's'' phonetically sounds like ``ato'' when spoken quickly, as in ``eight O's'' $\rightarrow$ ``ato'') forms the word \textbf{Potato}. This wordplay uses the visual of eight circles (O's) to create the suffix ``ato'' after ``POT''.\par \textbf{Answer}: Potato} \\
  \includegraphics[width=4cm]{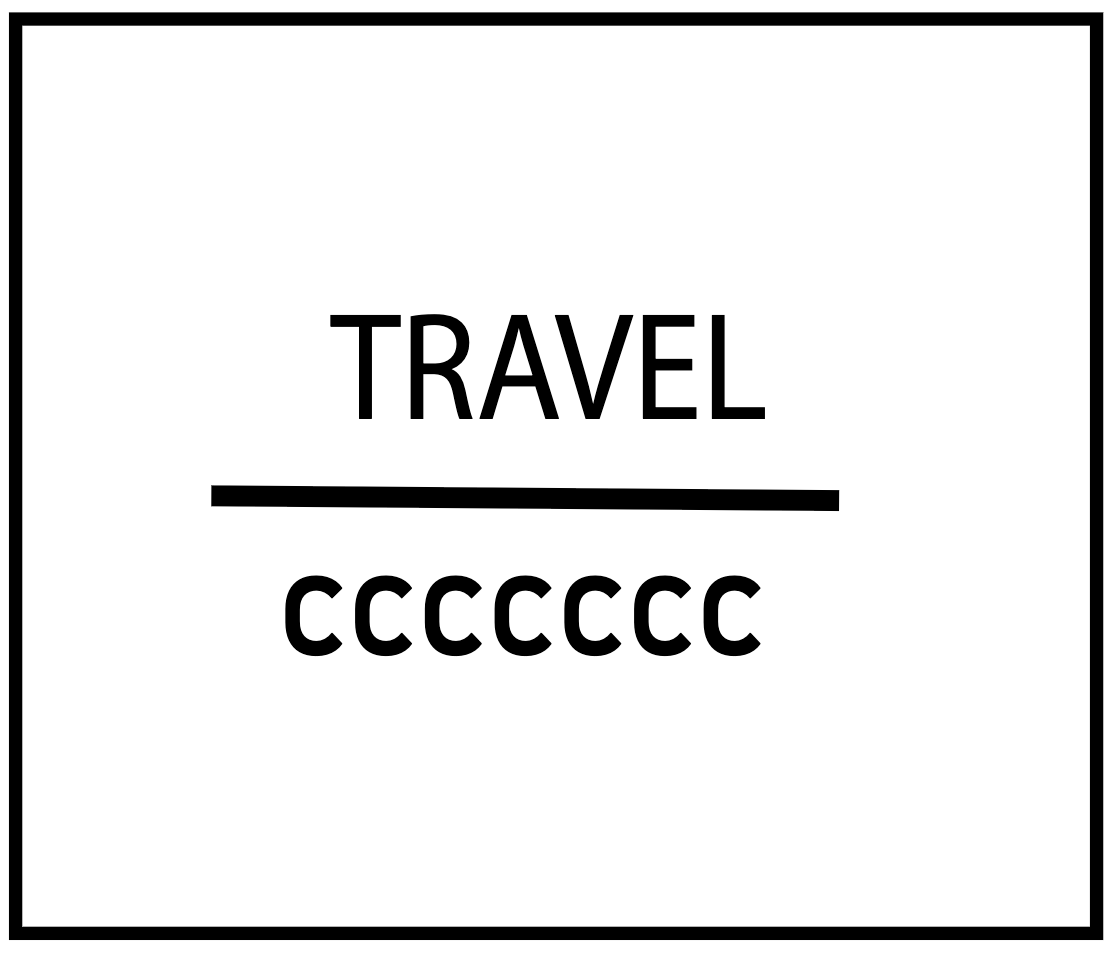} &
  {\textbf{Prompts:} can you solve this Rebus puzzle?\par \textbf{Model Responses:} \texttt{<think>...</think>}The rebus puzzle consists of ``TRAVEL'' above a line (representing a hyphen or connection) and seven ``C''s. Each ``C'' sounds like ``sea,'' so seven Cs are ``seven seas.'' Combining these, the solution is \textbf{``Travel the seven seas''}.} \\
  \bottomrule
  \end{tabular}
\caption{Three examples of \doubao{} solving Rebus puzzles. Our model gave the correct reasoning and answers to the first two. For the third one, although the model gave an incorrect answer it still output logical reasoning. The correct answer is ``travel overseas''. We omit thinking tokens for brevity.}
\label{fig:rebus_examples_app}
\end{figure}

\begin{figure}[h]
  \centering
  \begin{tabular}{m{4cm}m{12cm}}
  \toprule
  \includegraphics[width=4cm]{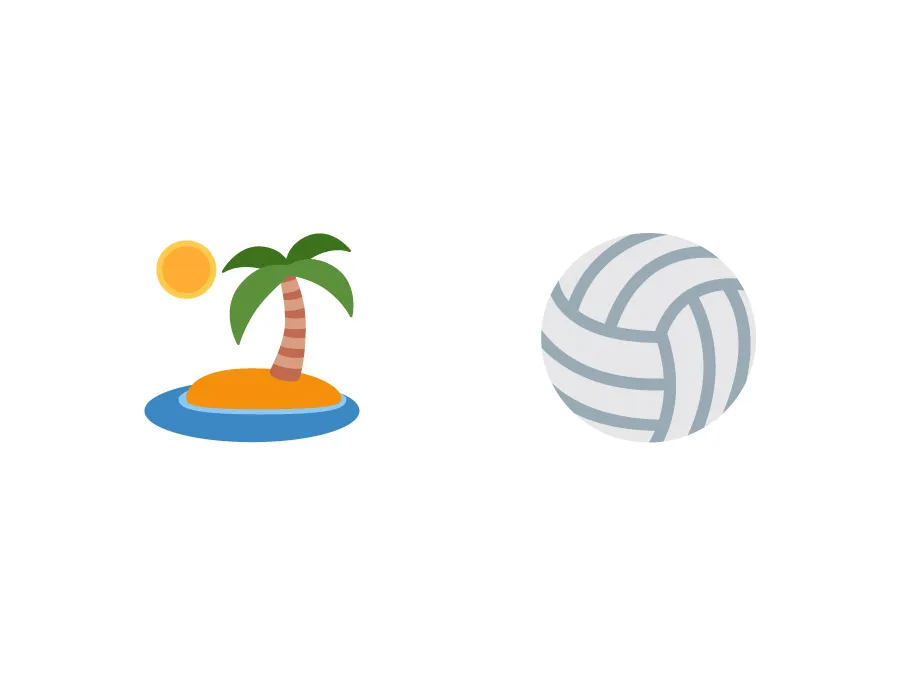} & 
  {\textbf{Prompts:} Which movie do these emojis remind you of?\par \textbf{Model Responses:} \texttt{<think>...</think>}The emojis---a tropical island (representing isolation/stranded setting) and a volleyball (referencing the character ``Wilson'' from the movie)—remind us of \textbf{``Cast Away''}. The film features a protagonist stranded on a deserted island, forming a unique bond with a volleyball named Wilson, which matches these symbols perfectly.} \\
  \includegraphics[width=4cm]{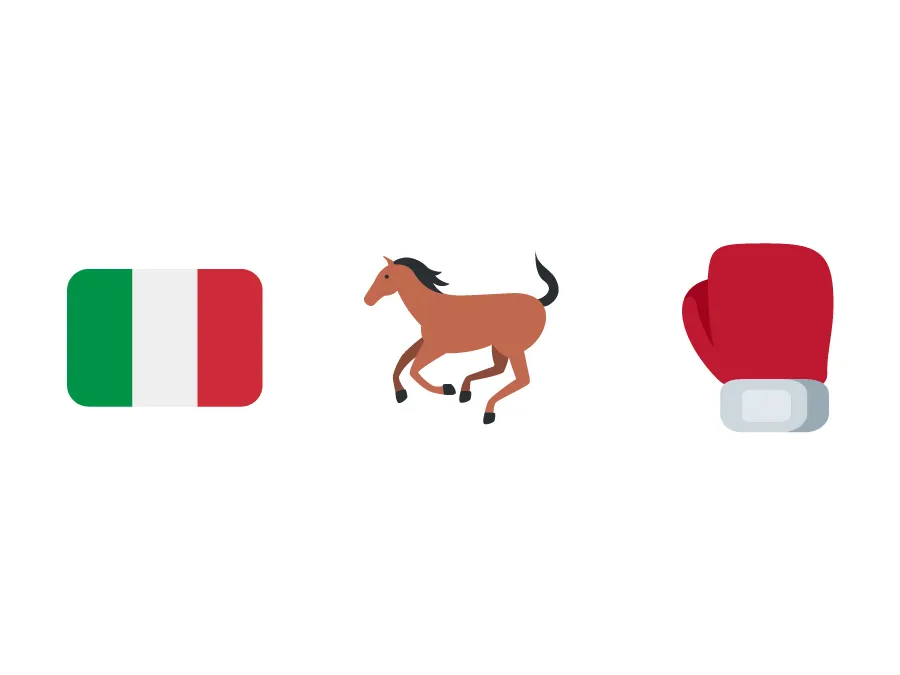} &
  {\textbf{Prompts:} Which movie do these emojis remind you of?\par \textbf{Model Responses:} \texttt{<think>...</think>}The emojis---a Italian flag (representing Italian heritage), a horse (referencing ``Stallion,'' as in Rocky's nickname ``The Italian Stallion''), and a boxing glove (boxing, Rocky's sport)---collectively evoke \textbf{``Rocky''}. Rocky Balboa, the Italian-American boxer known as ``The Italian Stallion,'' perfectly matches these symbols.} \\
  \includegraphics[width=4cm]{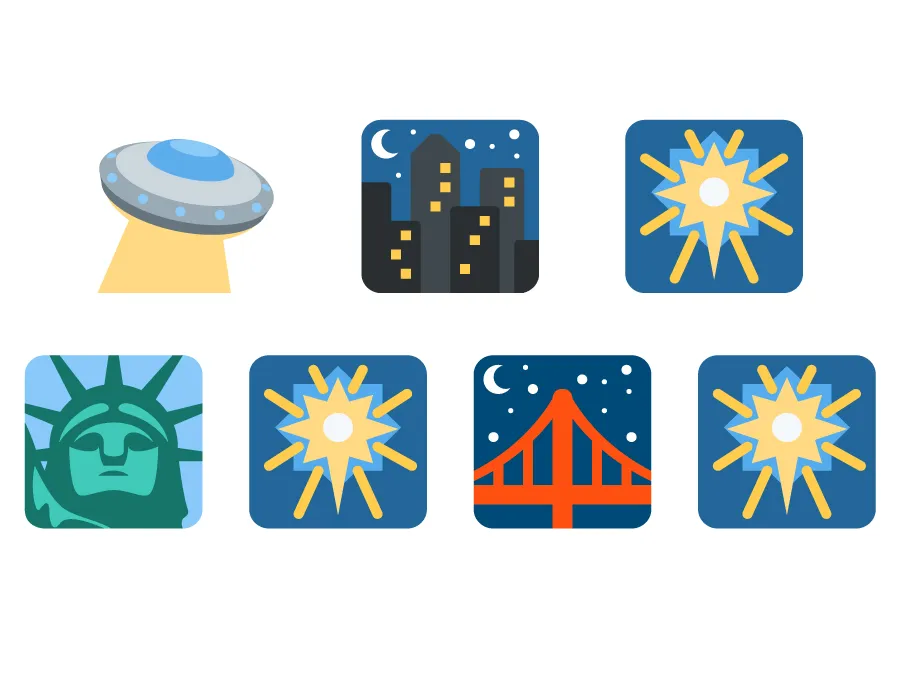} &
  {\textbf{Prompts:} Which movie do these emojis remind you of?\par \textbf{Model Responses:} \texttt{<think>...</think>}The emojis include a UFO (alien spacecraft), landmarks like the Statue of Liberty (New York) and Golden Gate Bridge (San Francisco), cityscapes, and explosions---all key elements from \textbf{``Independence Day''}, where aliens attack major cities with UFOs, resulting in widespread destruction. These symbols align perfectly with the movie's plot and iconic scenes.} \\
  \bottomrule
  \end{tabular}
\caption{Three illustrative instances of \doubao{} successfully identifying movie names from their corresponding emoji inputs. These examples highlight the model's ability to recognize and semantically link emoji icons to movie plot information.}
\label{fig:movie_quiz}
\end{figure}

\begin{figure}[h]
  \centering
  \begin{tabular}{m{4cm}m{12cm}}
  \toprule
  \includegraphics[width=4cm]{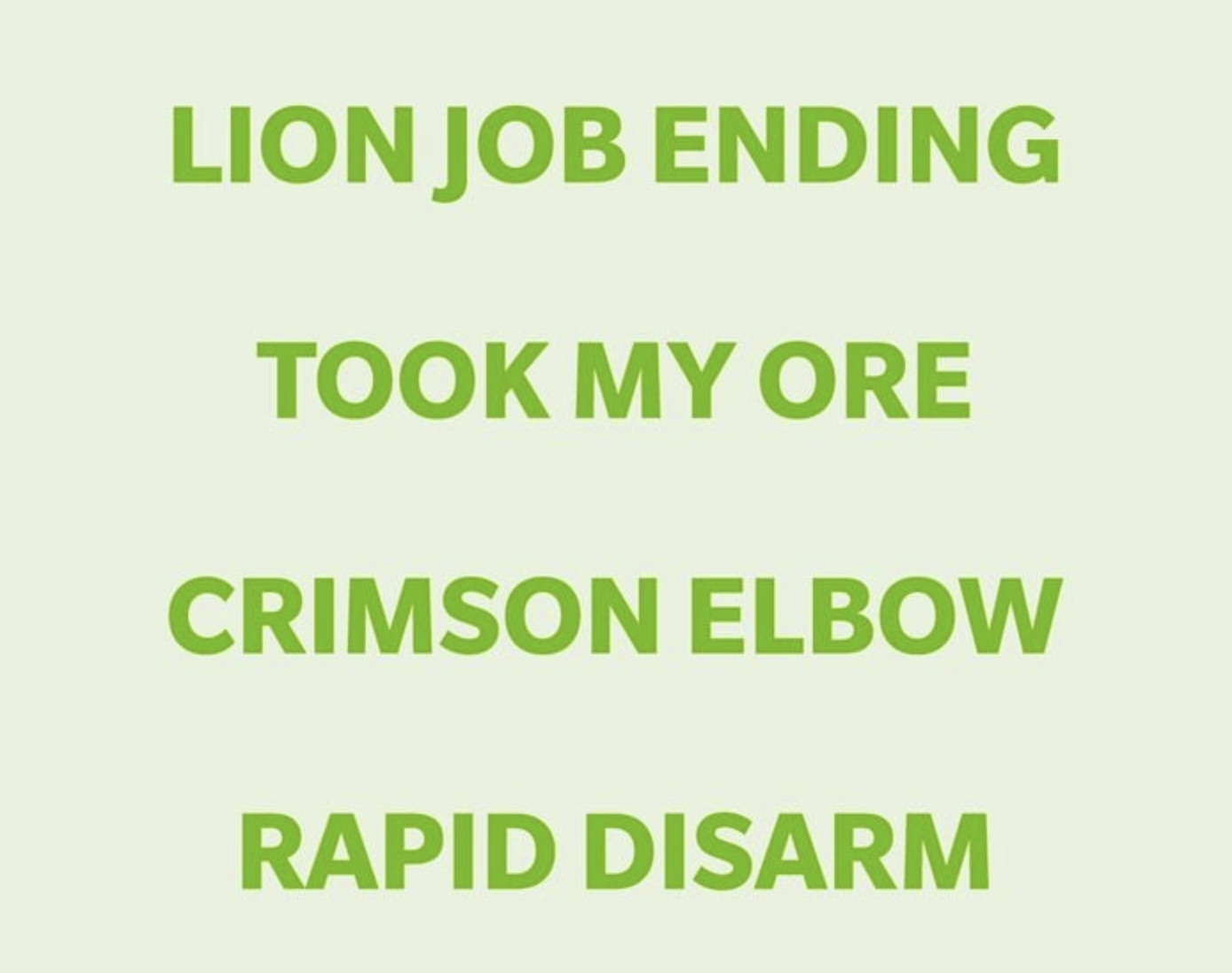} & 
  { {\bf Prompts:} These seemingly random phrases hide the names of eight capital cities around the world. Each phrase contains the name of two cities. Can you unscramble them and figure out what they are? \par {\bf Model Responses:} \texttt{<think>...</think>} The phrases contain the following capital cities, each pair from a line:
  \begin{enumerate}
      \item {\bf LION JOB ENDING} $\rightarrow$ {\bf London} (L-O-N-D-O-N) and {\bf Beijing} (B-E-I-J-I-N-G).
      \item {\bf TOOK MY ORE} $\rightarrow$ {\bf Tokyo} (T-O-K-Y-O) and {\bf Rome} (R-O-M-E).
      \item {\bf CRIMSON ELBOW} $\rightarrow$ {\bf Berlin} (B-E-R-L-I-N) and {\bf Moscow} (M-O-S-C-O-W).
      \item {\bf RAPID DISARM} $\rightarrow$ {\bf Paris} (P-A-R-I-S) and {\bf Madrid} (M-A-D-R-I-D).
  \end{enumerate}
These are all capital cities (London, Beijing, Tokyo, Rome, Berlin, Moscow, Paris, Madrid), each hidden in the letter combinations of the phrases.} \\
  \bottomrule
  \end{tabular}
\caption{An example of \doubao{} playing a word game. The model correctly recognize letters from the image, and retrieve capital cities that are spelled using those letters. This exemplifies the model's integrated OCR, reasoning, and knowledge retrieval capabilities.}
\label{fig:word_game}
\end{figure}

\begin{figure}[h]
  \centering
  \begin{tabular}{m{16cm}}
  \toprule
  \begin{center}
  {\small Input image}
  \end{center}
  \begin{center}
  \includegraphics[width=8cm]{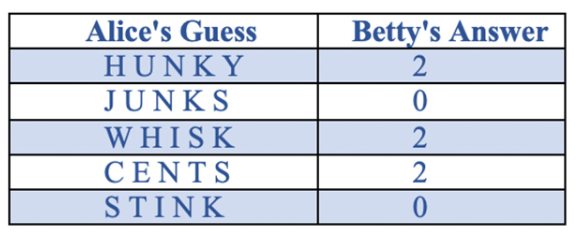}
  \end{center}
  {\footnotesize {\bf Prompt:}\newline
  Alice and Betty are playing ``guess my word''. Betty thinks of a five-letter word (all distinct letters), and Alice guesses. For each guess, Betty tells Alice how many letter from her guess are right. What is Betty's word?\newline\newline
  {\bf Model Responses:} 
  \texttt{<think>...</think>}To determine Betty's five-letter word (with all distinct letters), we analyze the clues from Alice's guesses:
  \begin{enumerate}
    \item {\bf Eliminate Letters Using Guesses with 0 Correct Letters}
    \begin{itemize}
        \item When Alice guesses {\bf JUNKS} and gets 0, it means the letters {\bf J, U, N, K, S} are not in Betty's word.
        \item When Alice guesses {\bf STINK} and gets 0, it means the letters {\bf S, T, I, N, K} are not in Betty's word.
        \item By taking the union of the eliminated letters from these two guesses, we rule out {\bf J, U, N, K, S, T, I}.
    \end{itemize}
    \item {\bf Identify Valid Letters from Guesses with 2 Correct Letters}
    \begin{itemize}
        \item For the guess {\bf HUNKY} (which has 2 correct letters), after eliminating the invalid letters (U, N, K), the valid letters that could be correct are {\bf H, Y}.
        \item For the guess {\bf WHISK} (also with 2 correct letters), after eliminating the invalid letters (I, S, K), the valid letters that could be correct are {\bf W, H}.
        \item For the guess {\bf CENTS} (with 2 correct letters), after eliminating the invalid letters (N, T, S), the valid letters that could be correct are {\bf C, E}.
        \item Combining these valid letters, we get the set {\bf H, Y, W, C, E}, which has exactly 5 distinct letters, satisfying the condition of a five-letter word.
    \end{itemize}
    \item {\bf Form the Word}
    \begin{itemize}
        \item We can form the word {\bf CHEWY} using the letters {\bf C, H, E, W, Y}.
        \item Let's verify this word against all the guesses:
        \begin{itemize}
            \item {\bf HUNKY}: The letters H and Y match, so there are 2 correct letters.
            \item {\bf JUNKS}: None of the letters J, U, N, K, S match, so there are 0 correct letters.
            \item {\bf WHISK}: The letters W and H match, so there are 2 correct letters.
            \item {\bf CENTS}: The letters C and E match, so there are 2 correct letters.
            \item {\bf STINK}: None of the letters S, T, I, N, K match, so there are 0 correct letters.
        \end{itemize}
    \end{itemize}
  \end{enumerate} 
  All the conditions are satisfied.\newline
  {\bf Answer:} CHEWY}
  \end{tabular}
  \caption{An example of solving a word game requiring multi-step reasoning. First, eliminate invalid letters. Second, identify valid ones, and finally, verify the derived letters with previous guesses.}
  \label{fig:word_game2}
\end{figure}

\begin{figure}[h]
  \centering
  \begin{tabular}{m{4cm}m{12cm}}
  \toprule
  \includegraphics[width=4cm]{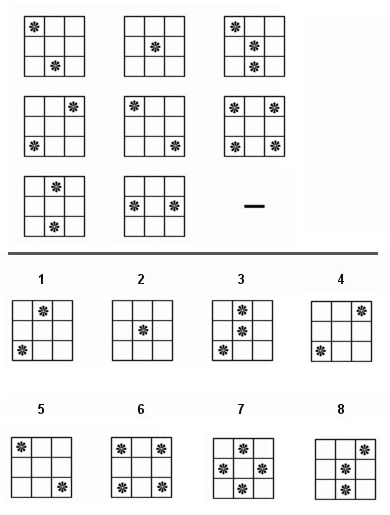} & 
  {\textbf{Prompts:} here is a visual reasoning puzzle. what is the last pattern indicated by ``-''? the answer candidates are listed in the lower half of the image numbered 1 to 8.
\par \textbf{Model Responses:} \texttt{<think>...</think>}The puzzle involves identifying the pattern in each row, where the third grid is the union (sum) of the first two.
\begin{itemize}
    \item \textbf{First row}: 2* + 1* = 3* (matches third grid).
    \item \textbf{Second row}: 2* + 2* = 4* (matches third grid, four corners).
    \item \textbf{Third row}: First two grids each have 2*, so the third must have 4*.
\end{itemize}

Analyzing candidates, \textbf{Candidate 7} has four asterisks arranged in a column (top-center, middle-center, bottom-center, and another? No, wait—correcting: no, finally, recognizing the union of third row's first two grids (2* each) results in 4*, and \textbf{Candidate 7} matches this pattern (four asterisks in a vertical column, aligning with the combined positions).
\par
\textbf{Answer: 7}} \\
\hline
  \includegraphics[width=4cm]{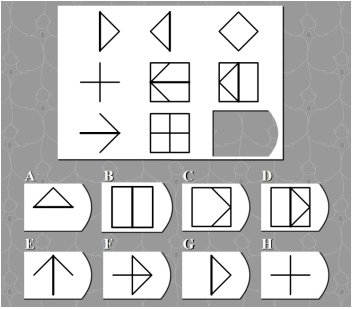} &
  {\textbf{Prompts:} here is a visual reasoning puzzle. what is the missing pattern listed as the last element in the 3rd row?  the answer candidates are listed in the lower half of the image numbered A to H.
\par 
\textbf{Model Responses:} \texttt{<think>...</think>}To solve the puzzle, analyze each row for patterns. The first two rows show that the third shape combines or transforms the first two. In the third row:
\begin{itemize}
    \item \textbf{First shape}: Arrow (right, stem).
    \item \textbf{Second shape}: Grid (square divided into four).
\end{itemize}
The missing third shape (candidate D) matches the pattern: it integrates the arrow's stem (horizontal line) with the grid's structure, forming a square with a diagonal and stem, consistent with the visual logic of combining row elements.
\textbf{Answer: D}} \\
  \bottomrule
  \end{tabular}
\caption{Two examples of \doubao{} using its visual inductive reasoning ability to correctly predict the missing patterns.}
\label{fig:pattern_recognition}
\end{figure}

\begin{figure}[!h]
  \centering
  \begin{tabular}{m{8cm}m{8cm}}
  \toprule
  \multicolumn{1}{c}{Input image pair} & \multicolumn{1}{c}{Input image pair overlaid by detected differences} \\
  \multicolumn{1}{c}{\includegraphics[width=0.4\columnwidth]{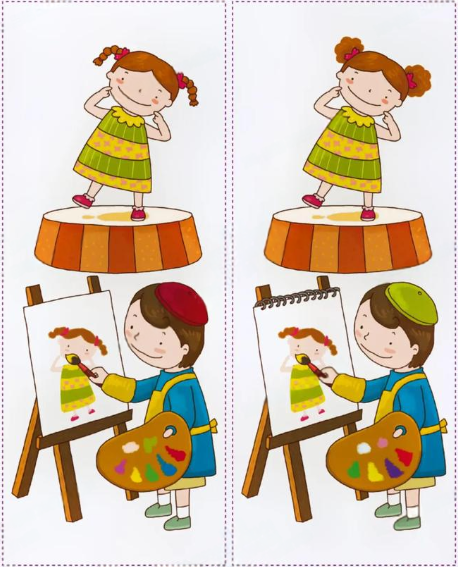}} &
  \multicolumn{1}{c}{\includegraphics[width=0.4\columnwidth]{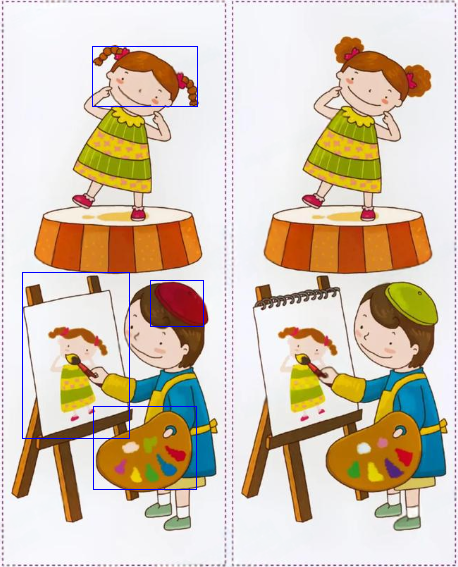}} \\
  \multicolumn{2}{@{}m{16cm}@{}}{
  \small
  \begin{CJK*}{UTF8}{gbsn}
  \textbf{Prompt:} 请仔细对比两张图，找出它们的不同之处。\newline
  针对每一处差异，请在第一张图中用边界框标记对应位置。每个差异区域使用\texttt{<bbox>}格式标注,格式为:\newline
  \texttt{<bbox>}左上角\texttt{x}坐标 左上角\texttt{y}坐标 右下角\texttt{x}坐标 右下角\texttt{y}坐标\texttt{</bbox>}\newline
  要求:\newline
  - 所有坐标值为0-999的整数,表示像素位置\newline
  - 左上角坐标必须小于右下角坐标\newline\newline
  \textbf{Model response:} \texttt{<think>...</think><bbox>202 82 432 188</bbox>} - 第一张图中女孩的双麻花辫\newline
  \texttt{<bbox>329 494 445 576</bbox>} - 第一张图中画家的红色帽子\newline
  \texttt{<bbox>49 480 282 773</bbox>} - 第一张图中无螺旋装订的画板\newline
  \texttt{<bbox>204 717 429 863</bbox>} - 第一张图中调色板的颜色（包含蓝色、紫色等，与第二张图不同）
  \end{CJK*}\newline\newline
  \textcolor{gray}{
  Translated prompt: Please carefully compare the two images and identify the differences between them.\newline
  For each difference, mark the corresponding location with bounding boxes on the first image. Annotate each differing region using the \texttt{<bbox> tag in the following format: <bbox>top-left-x top-left-y bottom-right-x bottom-right-y</bbox>}.\newline
  Requirements:\newline
  - All coordinate values must be integers between 0 and 999, representing pixel positions.\newline
  - The top-left coordinates must be smaller than the bottom-right coordinates.\newline\newline
  Translated model response: \texttt{<think>...</think><bbox>202 82 432 188</bbox>} - Girl's double braids in the first image\newline
  \texttt{<bbox>329 494 445 576</bbox>} - Painter's red hat in the first image\newline
  \texttt{<bbox>49 480 282 773</bbox>} - Regular sketchpad without spiral binding in the first image\newline
  \texttt{<bbox>204 717 429 863</bbox>} - Palette colors in the first image (containing blue, purple, etc., differing from the second image)
  }
  }\\
  \bottomrule
  \end{tabular}
  \caption{An example of model playing ``Find-the-differences'' game. The original prompt and model response are in Chinese. The English translation are shown in gray.}
  \label{fig:find_differences}
\end{figure}

\begin{figure}[h]
  \centering
  \begin{tabular}{m{4cm}m{12cm}}
  \toprule
  \includegraphics[width=4cm]{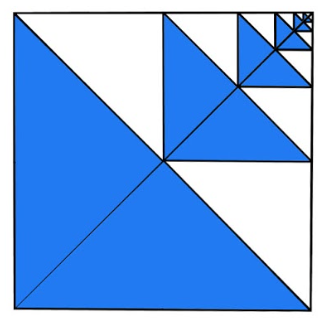} & 
  {\textbf{Prompts:} The square is filled in with a pattern of triangles that become smaller and smaller until they are infinitely small. what fraction of the square is shaded blue?\par
  \textbf{Model Responses:} \texttt{<think>...</think>}The shaded area forms a geometric series:
  \begin{enumerate}
      \item \textbf{Largest segment}: The two large blue triangles in the bottom-left occupy $\frac{1}{2}$ of the square.
      \item \textbf{Subsequent segments}: In the top-right quadrant, each blue triangle is $\frac{1}{4}$ the size of the previous shaded region (e.g., $\frac{1}{8}, \frac{1}{32}, \frac{1}{128}, \dots$).
  \end{enumerate}
  \par
  This is a geometric series with:
  \begin{itemize}
      \item First term \( a = \frac{1}{2} \)
      \item Common ratio \( r = \frac{1}{4} \)
  \end{itemize}
   Using the formula for the sum of an infinite geometric series $S=\frac{a}{1 - r}$:
   \par
   $S=\frac{\frac{1}{2}}{1-\frac{1}{4}}=\frac{\frac{1}{2}}{\frac{3}{4}}=\frac{2}{3}$
   \par
  \textbf{Answer}: $\frac{2}{3}$} \\
  \hline
  \includegraphics[width=4cm]{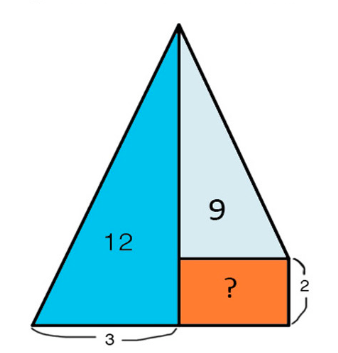} & 
  {\textbf{Prompts:} the area of the blue triangle is 12 and the area of the gray triangle is 9. what is the area of the orange rectangle?
\par
  \textbf{Model Responses:} \texttt{<think>...</think>}To find the area of the orange rectangle, we analyze the similar triangles (blue and gray) and use their areas:
  \begin{enumerate}
      \item \textbf{Blue Triangle}:
        \par
        $\text{Area}=\frac{1}{2} \times \text{base} \times \text{height} = 12$
        \par
        $\frac{1}{2} \times 3 \times H = 12 \Rightarrow H = 8$ (height of blue triangle).
      \item \textbf{Gray Triangle}:
        \par
        Its height is $H-2=8-2=6$ (since the rectangle's height is 2).
        $\text{Area}=\frac{1}{2}\times\text{base}\times 6=9 \Rightarrow \text{Base} = \frac{9 \times 2}{6} = 3$.
      \item \textbf{Orange Triangle}:
        \par
        The rectangle has \textbf{length=3} (matching the gray triangle's base) and \textbf{width=2}.
        \par
        $\text{Area}=3\times 2=6$
  \end{enumerate}

\textbf{Answer}: 6
} \\
  \bottomrule
  \end{tabular}
\caption{Two examples of \doubao{} solving geometry problems, which requires multi-step reasoning.}
\label{fig:geometric_series}
\end{figure}

\begin{figure}[h]
  \centering
  \begin{tabular}{m{8cm}m{8cm}}
  \toprule
  \multicolumn{1}{c}{Input image} & \multicolumn{1}{c}{Input image overlaid by detected objects}\\
  \multicolumn{1}{c}{\includegraphics[width=0.4\columnwidth]{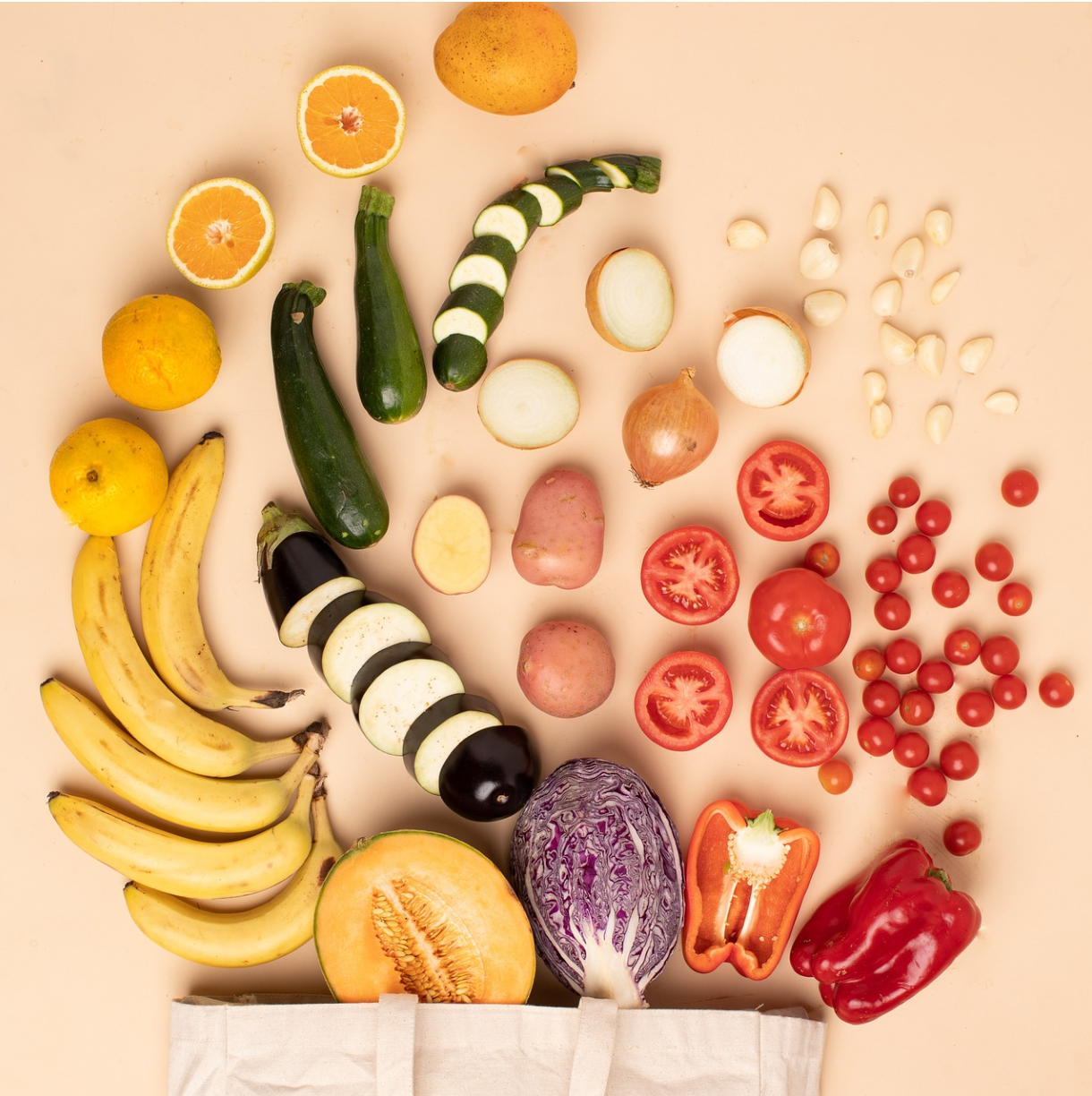}} & \multicolumn{1}{c}{\includegraphics[width=0.4\columnwidth]{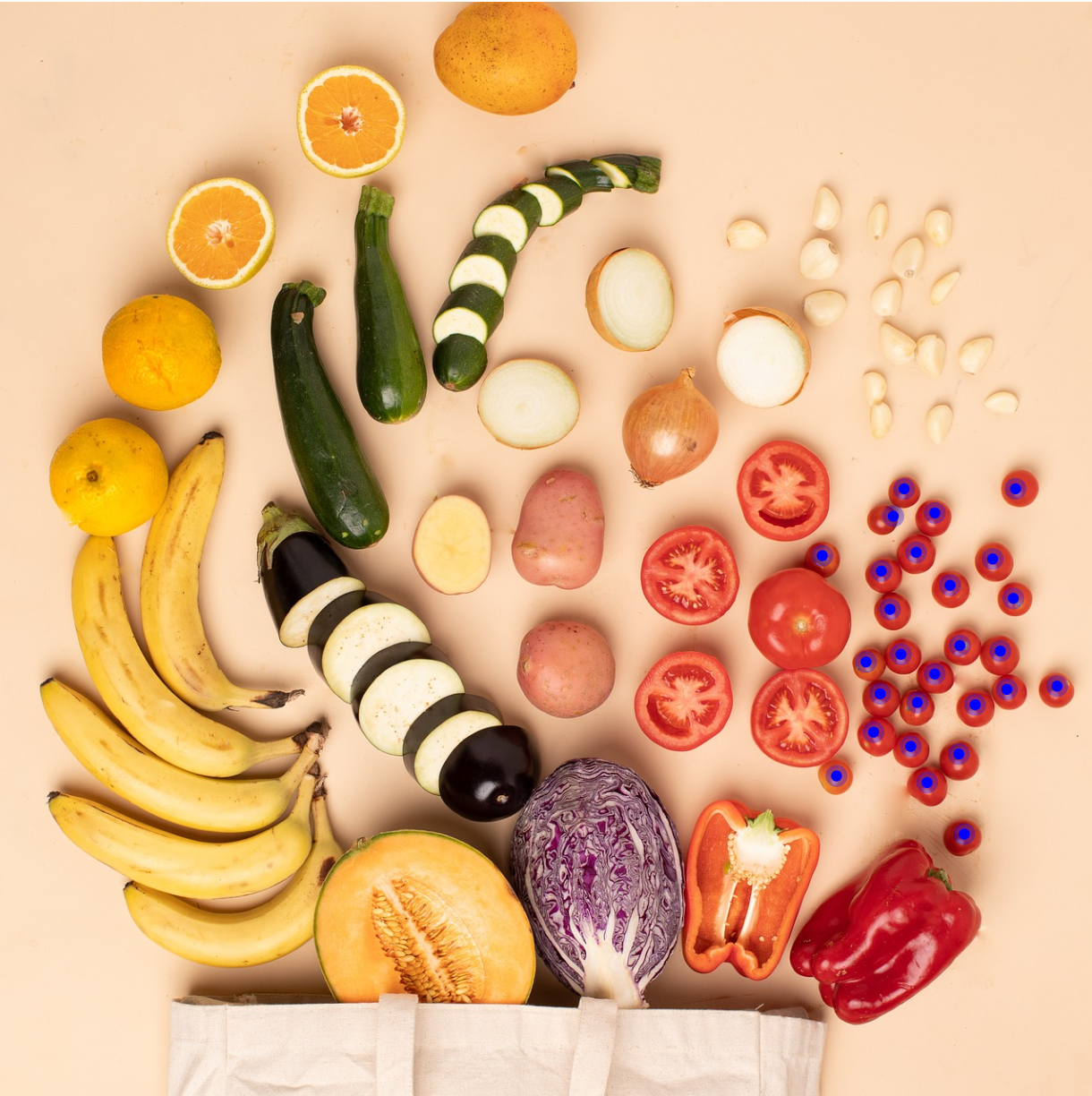}} \\
  \multicolumn{2}{@{}m{16cm}@{}}{
  \small
  \textbf{Prompt:}
  How many cherry tomatoes are there in the picture? Point them out first in \texttt{<point>x y</point>} format.\newline\newline
  \textbf{Model response:}
  \texttt{<point>766 708</point><point>818 471</point><point>828 446</point>
  <point>856 468</point><point>839 504</point><point>807 521</point><point>815 556</point><point>870 534</point><point>909 510</point><point>930 446</point><point>928 546</point><point>753 507</point><point>793 604</point><point>825 597</point><point>879 589</point><point>916 594</point><point>806 633</point><point>840 641</point><point>856 615</point><point>893 643</point><point>922 629</point><point>968 626</point><point>800 668</point><point>834 681</point><point>878 688</point><point>849 714</point><point>883 761</point>
  There are 27 cherry tomatoes in the picture}
  }\\
  \bottomrule
  \end{tabular}
  \caption{An example of \doubao{} recognizing queried objects and counting their occurrences. \doubao{} first detects the objects based on the user's prompt and then answers the query using the number of detected objects.}
  \label{fig:counting}
\end{figure}

\begin{figure}[h]
  \centering
  \begin{tabular}{m{7cm}m{7cm}}
  \toprule
  \includegraphics[width=0.4\columnwidth]{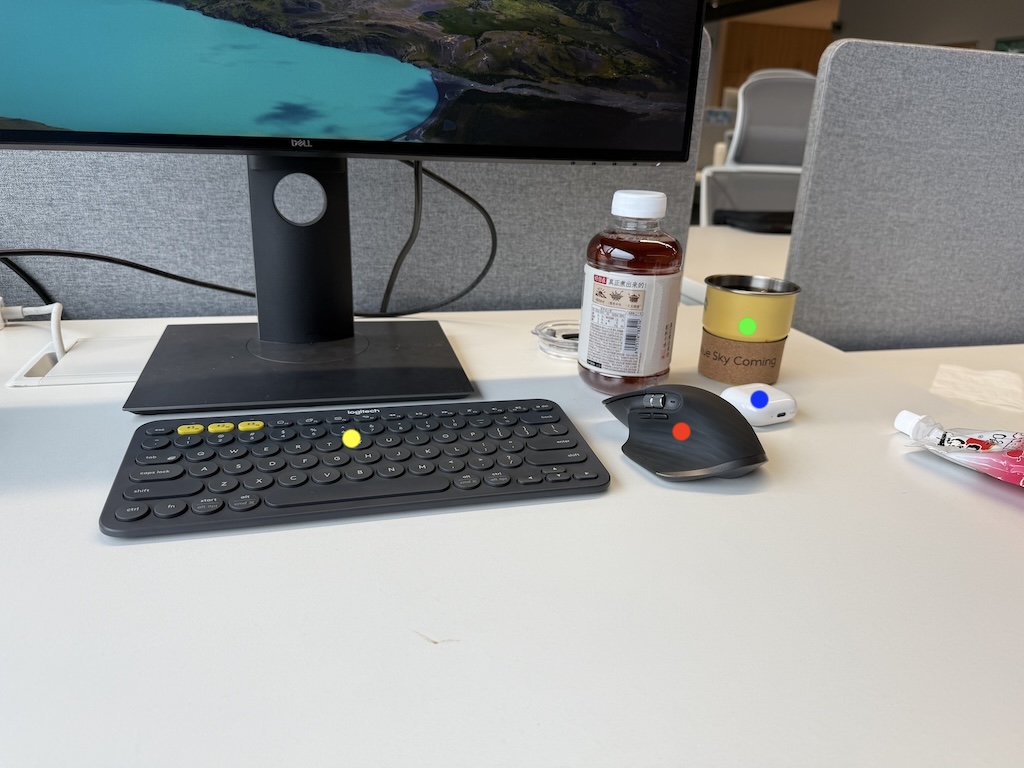} & 
  {\textbf{Prompts:} There are several points with different colors in the image: point1: blue, point2: yellow, point3: green, point4: red. Each point represents an object. Please sort the object represented by these points in a close to far order and give their name.\newline\newline
  \textbf{Model Responses:} point2: yellow (keyboard), point4: red (mouse), point1: blue (earphone), point3: green (cup)}\\
  \bottomrule
  \end{tabular}
  \caption{An example showing model's ability to estimate objects' depth ordering from a single image.\\Source: photo taken by an author from the ByteDance Seed team.}
  \label{fig:depth_sorting}
\end{figure}

\begin{figure}[h]
  \centering
  \begin{tabular}{m{16cm}}
  \toprule
  \begin{center}
  {\small Input video}
  \end{center}
  \begin{center}
  \includegraphics[width=15cm]{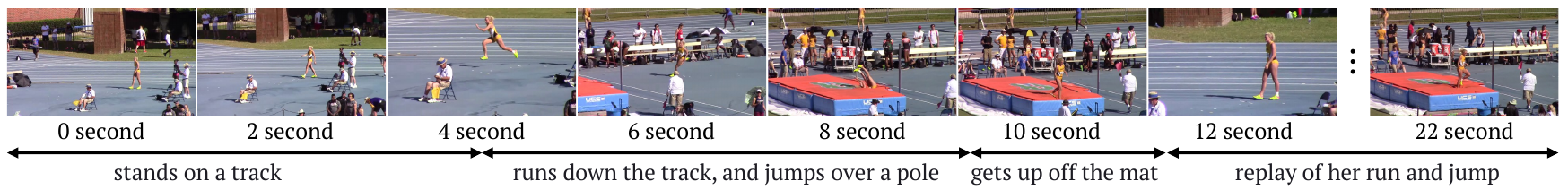}
  \end{center}
  {
  \textbf{Prompt:} Please watch this video and find out all key events in this video, and output the events along with the start/end timestamps.\newline
  \textbf{Model response:} 0.0--3.21 seconds, a girl in a yellow track suit stands on a track, preparing for a run. 3.32--10.02 seconds, she runs down the track, and jumps over a pole, landing on a mat. 10.13--12.16 seconds, she gets up off the mat, and walks away. 12.27--22.36 seconds, a replay of her run and jump is shown.\newline\newline
  \textbf{Prompt:} When does the woman jump? Please output the exact timestamps. If the woman jumps several times, please output the timestamps respectively.\newline
  \textbf{Model response:} 6--8 seconds, 16--22 seconds.
  }\\
  \bottomrule
  \end{tabular}
  \caption{The first query shows the model's ability to summarize/segment key events in a video (\eg, dense video captioning), and the second query shows model's ability of localizing an event in a video (\eg, video grounding).}
  \label{fig:video_grounding_example}
\end{figure}

\begin{figure}[h]
  \centering
  \begin{tabular}{m{6cm}m{10cm}}
  \toprule
  \includegraphics[width=6cm]{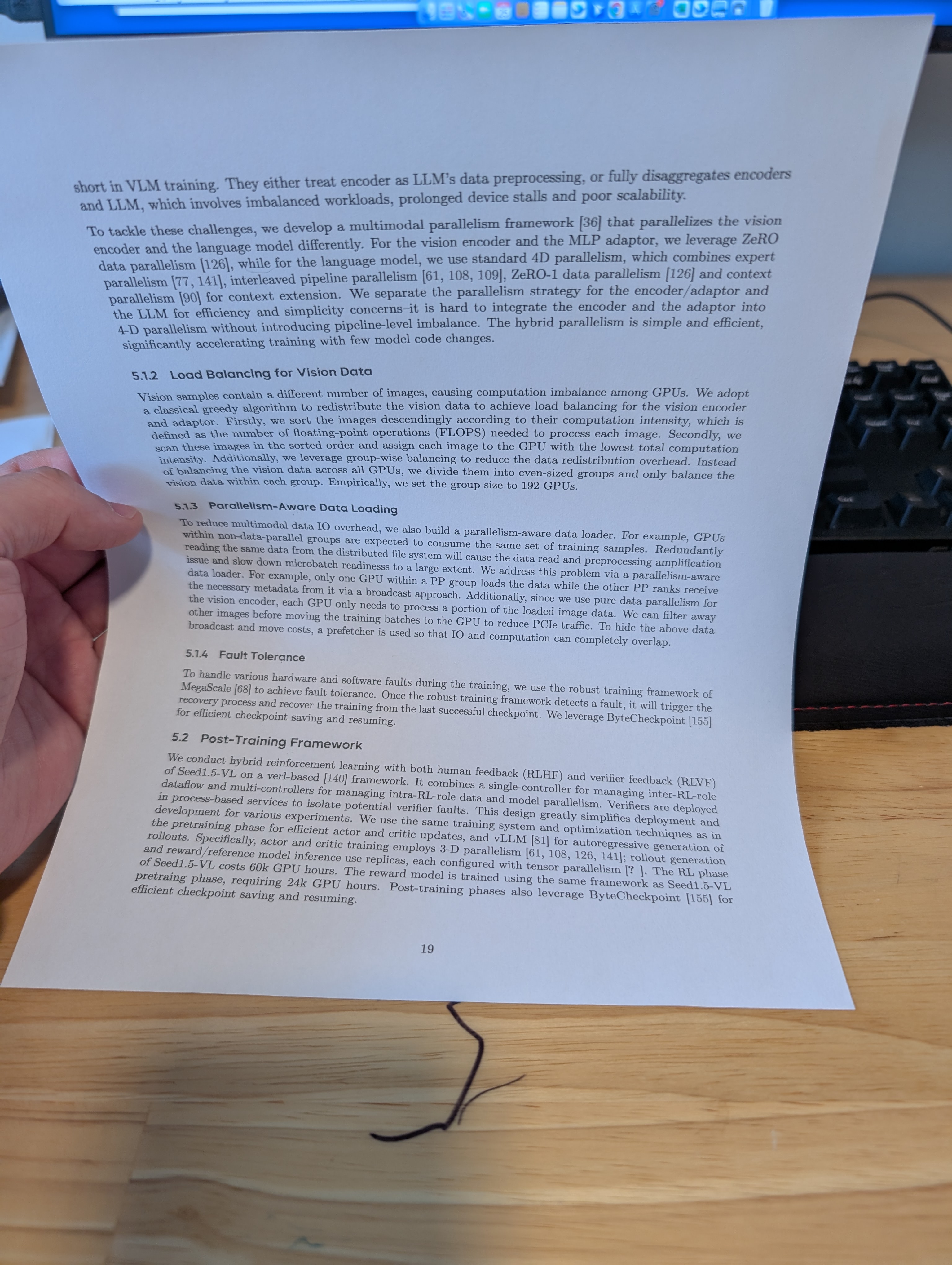} & 
  {\textbf{Prompts:} Based on the paper, what framework does the author use to achieve fault tolerance?\par \textbf{Model Responses:} According to the paper, to handle various hardware and software faults during the training, the author uses the robust training framework of MegaScale to achieve fault tolerance.\newline\newline
  \textbf{Prompts:} Based on the paper, which one takes longer to train, RL phase or reward model?\par \textbf{Model Responses:}
  \begin{enumerate}
      \item First, find the relevant information in the paper:
      \begin{itemize}
          \item The paper states: ``The RL phase of Seed1.5-VL costs 60k GPU hours. The reward model is trained using the same framework as Seed1.5-VL pretraining phase, requiring 24k GPU hours.''
      \end{itemize}
      \item Then, compare the GPU-hour requirements:
      \begin{itemize}
          \item The RL phase requires 60k GPU hours, and the reward model requires 24k GPU hours.
      \end{itemize}
  \end{enumerate}
  So, the RL phase takes longer to train.\newline\newline
  \textbf{Prompts:} Can you repeat section 5.1.4 but skip the first sentence?\newline
  \textbf{Model Responses:} Once the robust training framework detects a fault, it will trigger the recovery process and recover the training from the last successful checkpoint. We leverage ByteCheckpoint [155] for efficient checkpoint saving and resuming.
  }\\
  \bottomrule
  \end{tabular}
\caption{
Demonstration of \doubao{}'s ability of accurately extracting texts from an image, understanding the user's query, and retrieving the answer from the document. The final example also highlights model's instruction following capability (\eg, ``skip the first sentence'').
\\
Source: a printout of an earlier draft of the technical report.
}
\label{fig:ocr_parse_qa}
\end{figure}

\begin{figure}[h]
  \centering
  \begin{tabular}{m{6cm}m{10cm}}
  \toprule
  \includegraphics[width=6cm]{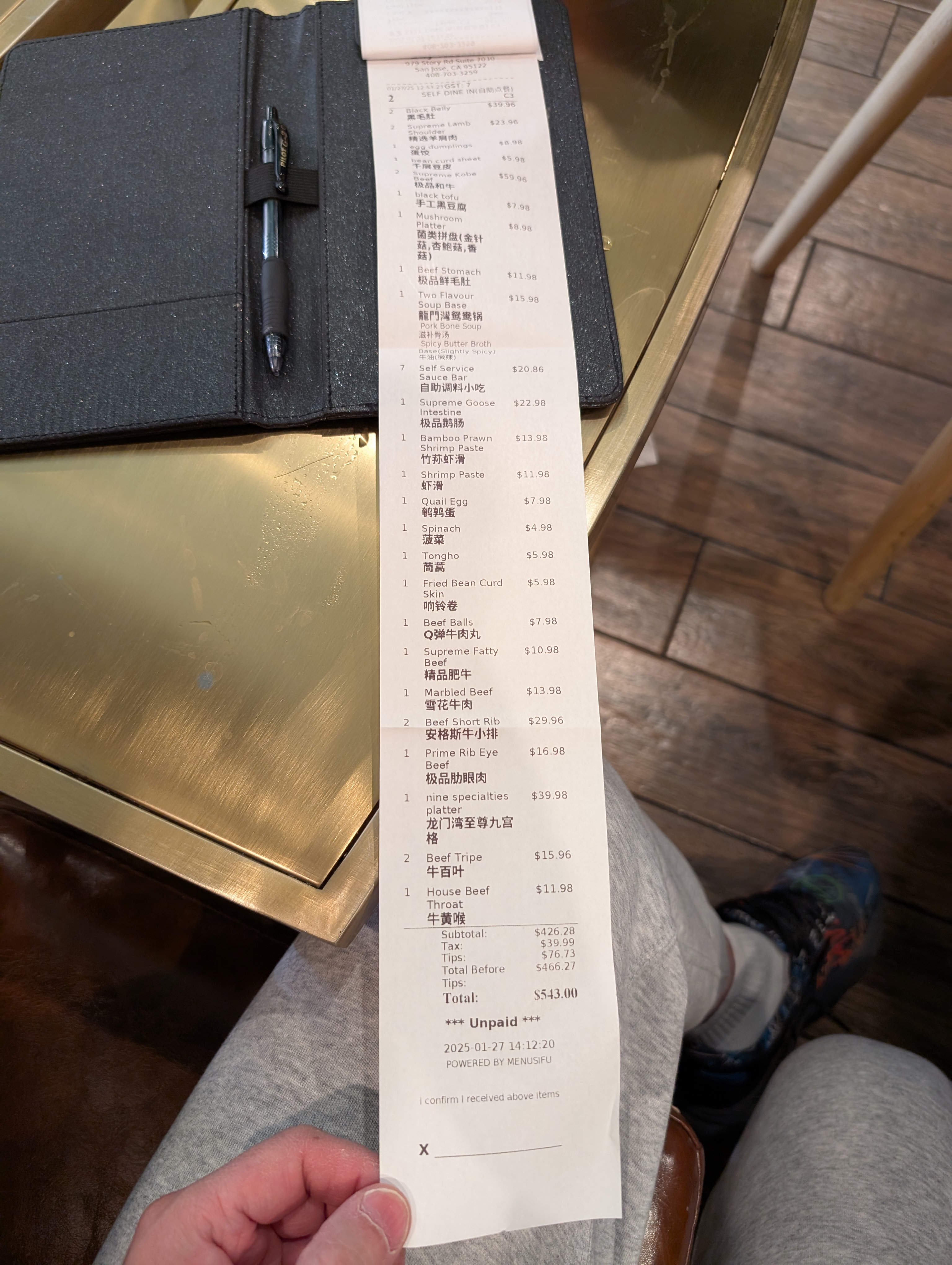} & 
  {\footnotesize {\bf Prompts:} can you extract the dish names, quantities, and their prices into a markdown table?
  \par {\bf Model Responses:}
  \par
  \begin{CJK*}{UTF8}{gbsn}
  \begin{Verbatim}[breaklines=true,breaksymbol=]
| Dish Name | Quantity | Price |
| --- | --- | --- |
| Black Belly (黑毛肚) | 2 | $39.96 |
| Supreme Lamb Shoulder (精选羊肩肉) | 2 | $23.96 |
| Egg Dumplings (蛋饺) | 1 | $8.98 |
| Bean Curd Sheet (干煎豆皮) | 1 | $5.98 |
| Supreme Kobe Beef (极品和牛) | 2 | $59.96 |
| Black Tofu (手工黑豆腐) | 1 | $7.98 |
| Mushroom Platter (菌类拼盘(金针菇,杏鲍菇,香菇)) | 1 | $8.98 |
| Beef Stomach (极品鲜毛肚) | 1 | $11.98 |
| Two Flavour Soup Base (龙门湾鸳鸯锅) | 1 | $15.98 |
| Self Service Sauce Bar (自助调料小吃) | 7 | $20.86 |
| Supreme Goose Intestine (极品鹅肠) | 1 | $22.98 |
| Bamboo Prawn Shrimp Paste (竹荪虾滑) | 1 | $13.98 |
| Shrimp Paste (虾滑) | 1 | $11.98 |
| Quail Egg (鹌鹑蛋) | 1 | $7.98 |
| Spinach (菠菜) | 1 | $4.98 |
| Tongho (茼蒿) | 1 | $5.98 |
| Fried Bean Curd Skin (响铃卷) | 1 | $5.98 |
| Beef Balls (Q弹牛肉丸) | 1 | $7.98 |
| Supreme Fatty Beef (精品肥牛) | 1 | $10.98 |
| Marbled Beef (雪花牛肉) | 1 | $13.98 |
| Beef Short Rib (安格斯牛小排) | 2 | $29.96 |
| Prime Rib Eye Beef (极品肋眼肉) | 1 | $16.98 |
| nine specialties platter (龙门湾至尊九宫格) | 1 | $39.98 |
| Beef Tripe (牛百叶) | 2 | $15.96 |
| House Beef Throat (牛黄喉) | 1 | $11.98 |
\end{Verbatim}
\end{CJK*}
  } \\
  \bottomrule
  \end{tabular}
\caption{An example of extracting multilingual texts from a long receipt.\\
Source: photo taken by an author from the ByteDance Seed team.}
\label{fig:long_receipt}
\end{figure}

\begin{figure}[!h]
  \centering
  \begin{tabular}{m{16cm}}
  \toprule
  \begin{center}
  \includegraphics[width=10cm]{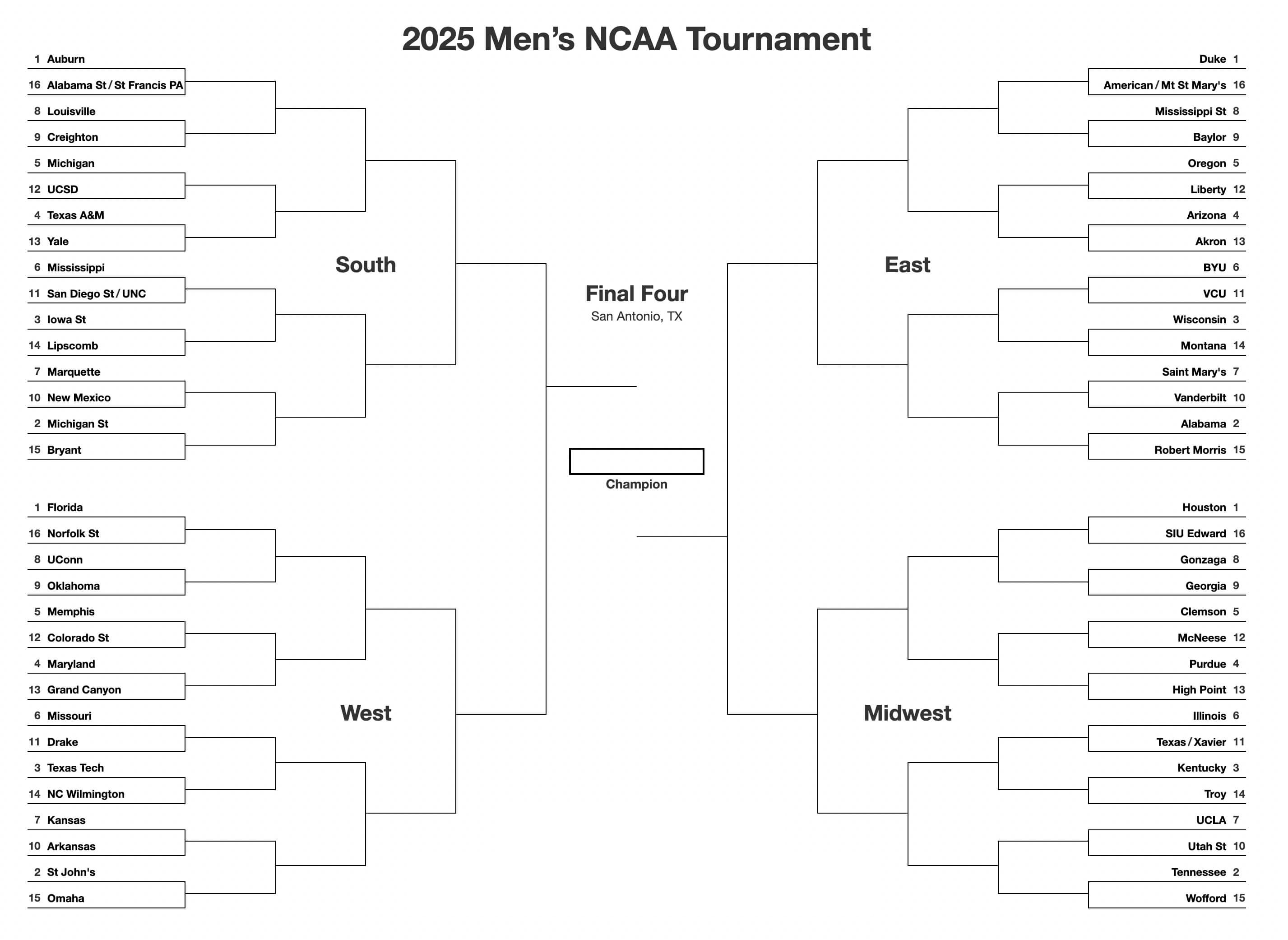}
  \end{center}
  {\scriptsize {\bf Prompts:} can you generate the Mermaid code that produce the bracket for ONLY the South region?
  \par {\bf Model Responses:} \texttt{<think>...</think>}Here’s the Mermaid code for the South Region bracket:
  \par
  \begin{Verbatim}[breaklines=true,breaksymbol=]
flowchart TD
    S1[1\nAuburn] --> S1_16[1v16\nAuburn vs Alabama St/St Francis PA]
    S16[16\nAlabama St/St Francis PA] --> S1_16
    S8[8\nLouisville] --> S8_9[8v9\nLouisville vs Creighton]
    S9[9\nCreighton] --> S8_9
    S5[5\nMichigan] --> S5_12[5v12\nMichigan vs UCSD]
    S12[12\nUCSD] --> S5_12
    S4[4\nTexas A&M] --> S4_13[4v13\nTexas A&M vs Yale]
    S13[13\nYale] --> S4_13
    S6[6\nMississippi] --> S6_11[6v11\nMississippi vs San Diego St/UNC]
    S11[11\nSan Diego St/UNC] --> S6_11
    S3[3\nIowa St] --> S3_14[3v14\nIowa St vs Lipscomb]
    S14[14\nLipscomb] --> S3_14
    S7[7\nMarquette] --> S7_10[7v10\nMarquette vs New Mexico]
    S10[10\nNew Mexico] --> S7_10
    S2[2\nMichigan St] --> S2_15[2v15\nMichigan St vs Bryant]
    S15[15\nBryant] --> S2_15
    
    S1_16 --> S2nd1[2nd Round\nWinner 1v16 vs 8v9]
    S8_9 --> S2nd1
    S5_12 --> S2nd2[2nd Round\nWinner 5v12 vs 4v13]
    S4_13 --> S2nd2
    S6_11 --> S2nd3[2nd Round\nWinner 6v11 vs 3v14]
    S3_14 --> S2nd3
    S7_10 --> S2nd4[2nd Round\nWinner 7v10 vs 2v15]
    S2_15 --> S2nd4
    
    S2nd1 --> S3rd1[Elite 8\nWinner 2nd1 vs 2nd2]
    S2nd2 --> S3rd1
    S2nd3 --> S3rd2[Elite 8\nWinner 2nd3 vs 2nd4]
    S2nd4 --> S3rd2
    
    S3rd1 --> SWinner[South Winner\nWinner 3rd1 vs 3rd2]
    S3rd2 --> SWinner
\end{Verbatim}
  }\\
  \bottomrule
  \end{tabular}
\caption{An example of model's several capabilities, such as OCR, coding, instruction following, working together to generate Mermaid code for a portion of the diagram. Summarization texts are omitted from the response. Paste the code into the \href{https://mermaid-js.github.io/mermaid-live-editor/}{Mermaid Live Editor} to visualize the South region bracket.}
\label{fig:ncaa_bracket}
\end{figure}

\begin{figure}[h]
  \centering
  \begin{tabular}{m{6cm}m{10cm}}
  \toprule
  \includegraphics[width=6cm]{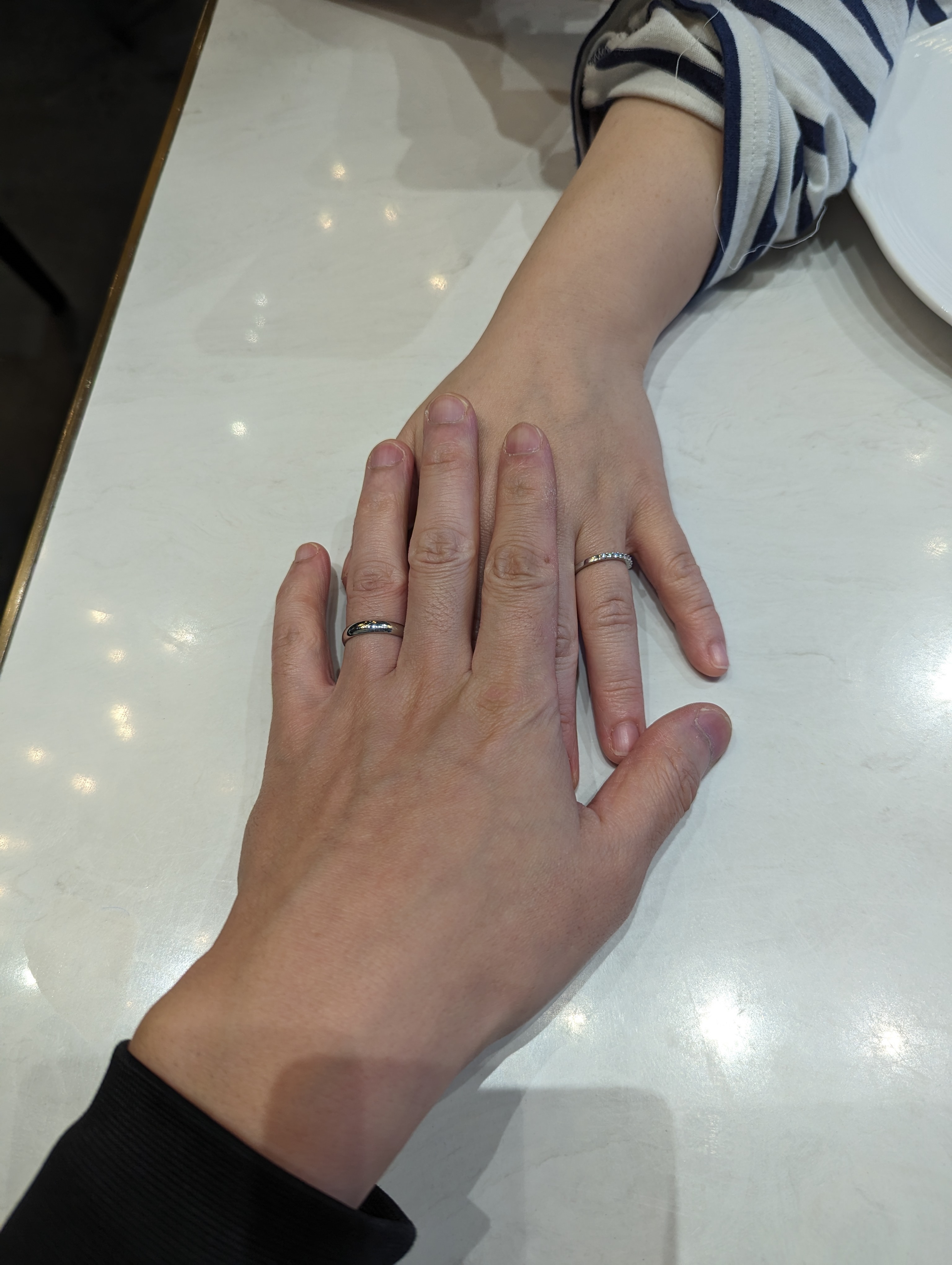} & 
  {\textbf{Prompts:} write a romantic poem in English inspired by this picture.
\par
  \textbf{Model Responses:} \texttt{<think>...</think>}
  \par
  \begin{Verbatim}[breaklines=true,breaksymbol=]
Fingers Entwined
  
Beneath the hum of a world that rushes by,
Our hands meet---soft, a silent lullaby.
Your palm, a map of stories yet untold,
My fingers trace the edges, brave and bold.

Rings gleam, twin sparks in this quiet space,
Promises etched in silver’s tender embrace.
Each line, each crease, a secret only we share---
Time slows, as if love itself holds its breath here.

Remember when our shadows first aligned?
Now skin to skin, no distance left to find.
These hands, they’ll carry dawns and midnight’s hush,
Weave years into a tapestry of "us."

For in this touch, a universe unfolds---
Heartbeats syncing, futures yet to hold.
Two hands, one song, forever intertwined---
A language only love could ever design.

  \end{Verbatim}
  \par
The poem mirrors the intimacy of clasped hands, weaving themes of connection, memory, and enduring commitment, with imagery of touch and time to echo the quiet romance in the photograph.
  } \\
  \bottomrule
  \end{tabular}
\caption{An example of image-conditioned creative text generation. The model processes an input image and generates a poem semantically related to the visual content.\\
Source: picture taken by an author from the ByteDance Seed team.}
\label{fig:creative_writing}
\end{figure}

\begin{figure}[h]
  \centering
  \begin{tabular}{m{6cm}m{10cm}}
  \toprule
  \includegraphics[width=6cm]{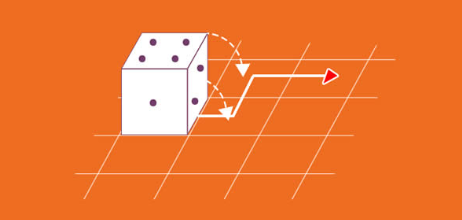} & 
  {\textbf{Prompts:} if we move the dice along the arrow directions, what will be the number on top?\par \textbf{Model Responses:} \texttt{<think>...</think>}To determine the number on top after moving the dice along the arrow directions, we analyze the initial configuration and movements:
  \begin{enumerate}
  \item \textbf{Initial Dice Faces}:
    \begin{itemize}
        \item Front: 1 dot (opposite back: 6 dots, $1 + 6 = 7$).
        \item Right: 2 dots (opposite left: 5 dots, $2 + 5 = 7$).
        \item Top: 5 dots (opposite bottom: 2 dots, $5 + 2 = 7$).
    \end{itemize}
  \item \textbf{Movements}:
    \begin{itemize}
        \item \textbf{Down Arrow}: Rolling the die downward (front to bottom) does not affect the top face directly (since top/bottom and front/back are separate axes).
        \item \textbf{Right Arrow}: Rolling the die rightward (right to front) also does not change the top face (top remains independent of right/left movements in this sequence).
    \end{itemize}
  \end{enumerate}
  Thus, the top face (initially 5 dots) remains unchanged after the specified movements.
  \par
  \textbf{Answer}: 5
  }
  \\
  \hline
  \includegraphics[width=6cm]{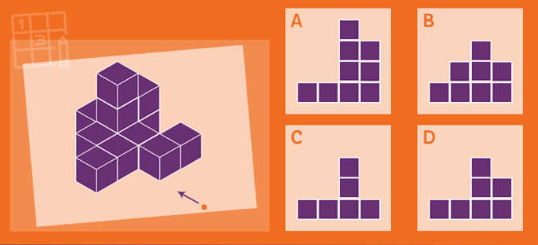} &
  {\textbf{Prompts:} View the blocks on the left from the orange point and direction of the arrow. Which image on the right is the correct view?
\par \textbf{Model Responses:} \texttt{<think>...</think>}To determine the correct front view from the orange point and arrow direction, analyze the vertical height of each column in the 3D structure:
\begin{itemize}
    \item \textbf{Leftmost column}: 1 cube (bottom layer only).
    \item \textbf{Second column}: 2 cubes (two layers).
    \item \textbf{Third column (central, tallest)}: 3 cubes (three layers).
    \item \textbf{Fourth column}: 2 cubes (two layers).
\end{itemize}
Matching this height distribution (1, 2, 3, 2) to the options:
\begin{itemize}
    \item \textbf{A}: Incorrect (leftmost column too tall).
    \item \textbf{B}: Incorrect (symmetric, not matching the structure).
    \item \textbf{C}: Incorrect (heights don’t align).
    \item \textbf{D}: Correct (heights 1, 2, 3, 2 match the front view).
\end{itemize}
\par
\textbf{Answer:} D
} \\
  \bottomrule
  \end{tabular}
\caption{Two examples where \doubao{}, Gemini 2.5 Pro, and OpenAI o4-mini all failed to produce the correct reasoning. In the \doubao{} case, the reasoning is incorrect despite a correct final answer. This highlights a limitation of contemporary VLMs in tasks requiring robust 3D spatial reasoning.}
\label{fig:failure_spatial_imagination}
\end{figure}

\begin{figure}[h]
  \centering
  \begin{tabular}{m{6cm}m{10cm}}
  \toprule
  \includegraphics[width=6cm]{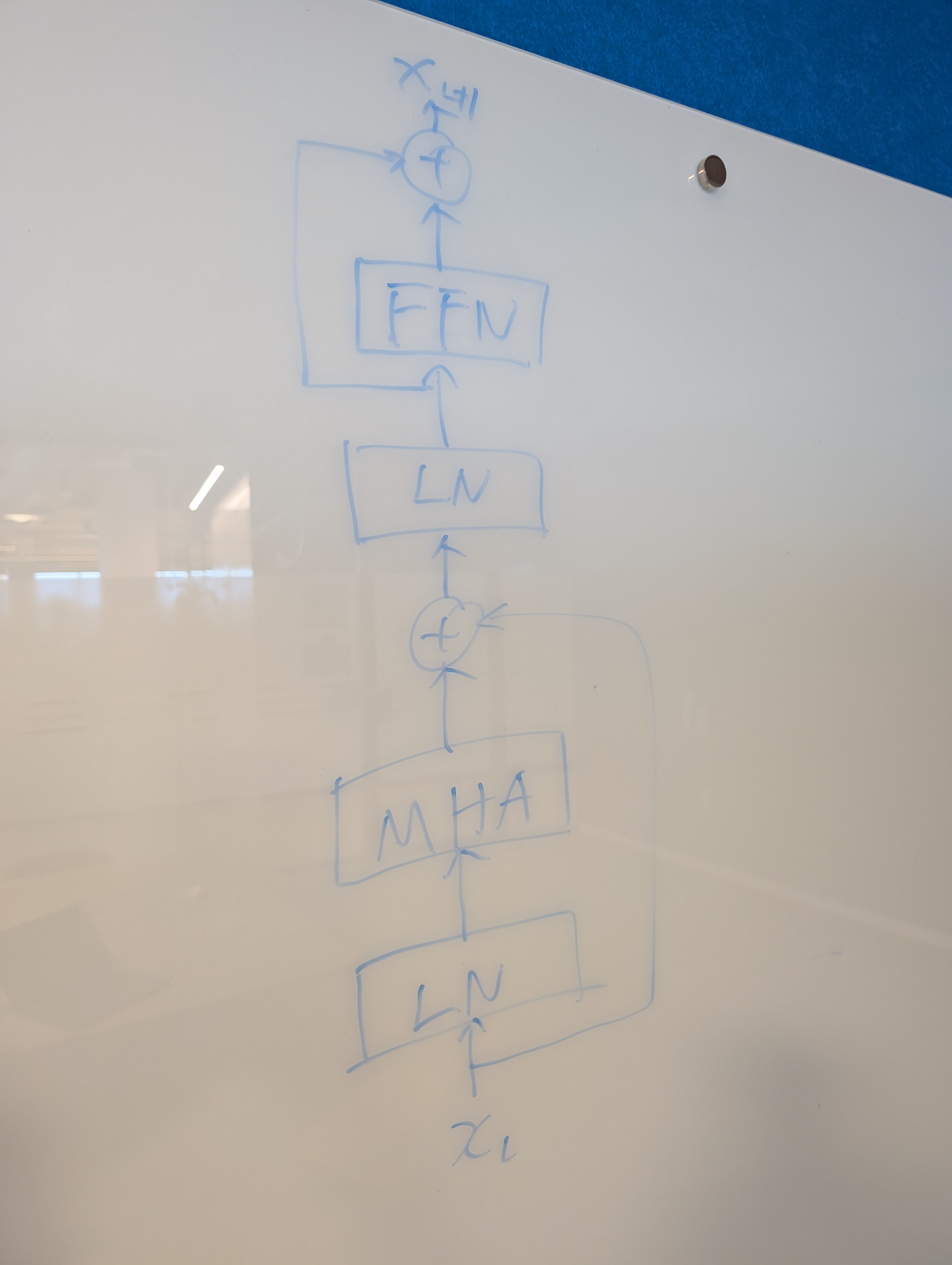} & 
  {\footnotesize \textbf{Prompts:} can you generate the DOT code that renders a similar graph to this one?
\par
  \textbf{Model Responses:} \texttt{<think>...</think>}Here's the DOT code to render a graph similar to your Transformer layer diagram:
  \par
  \begin{Verbatim}[breaklines=true,breaksymbol=]
  digraph TransformerLayer {
    rankdir=TB;
    node [fontname="Sans-Serif"];
    
    // Define nodes
    input [label="x_l", shape=rectangle];
    ln1 [label="LN", shape=rectangle];
    mha [label="MHA", shape=rectangle];
    add1 [label="+", shape=circle];
    ln2 [label="LN", shape=rectangle];
    ffn [label="FFN", shape=rectangle];
    add2 [label="+", shape=circle];
    output [label="x_{l+1}", shape=rectangle];
    
    // Define edges
    input -> ln1;
    ln1 -> mha;
    mha -> add1;
    input -> add1;  // Residual connection (first skip)
    
    add1 -> ln2;
    ln2 -> ffn;
    ffn -> add2;
    add1 -> add2;  // Residual connection (second skip)
    
    add2 -> output;
    
    // Optional styling (you can adjust colors/shapes)
    edge [arrowhead="vee", arrowsize=0.8];
    node [style=filled, fillcolor=white, penwidth=1];
    add1, add2 [fillcolor=lightblue];
  }
  \end{Verbatim}
  \par
  $\dots$
  } \\
  \bottomrule
  \end{tabular}
\caption{A case study illustrating the inability of \doubao{}, Gemini 2.5 Pro, and OpenAI o4-mini to generate faithful DOT code when presented with a non-standard FFN residual connection. The drawing shows the network structure, specifically the atypical residual path. All models exhibited a tendency to generate DOT code corresponding to a standard residual connection.}
\label{fig:hallucination}
\end{figure}

\begin{figure}[h]
  \centering
  \begin{tabular}{m{16cm}}
  \toprule
  \begin{center}
  Input image
  \end{center}
  \begin{center}
  \includegraphics[width=8cm]{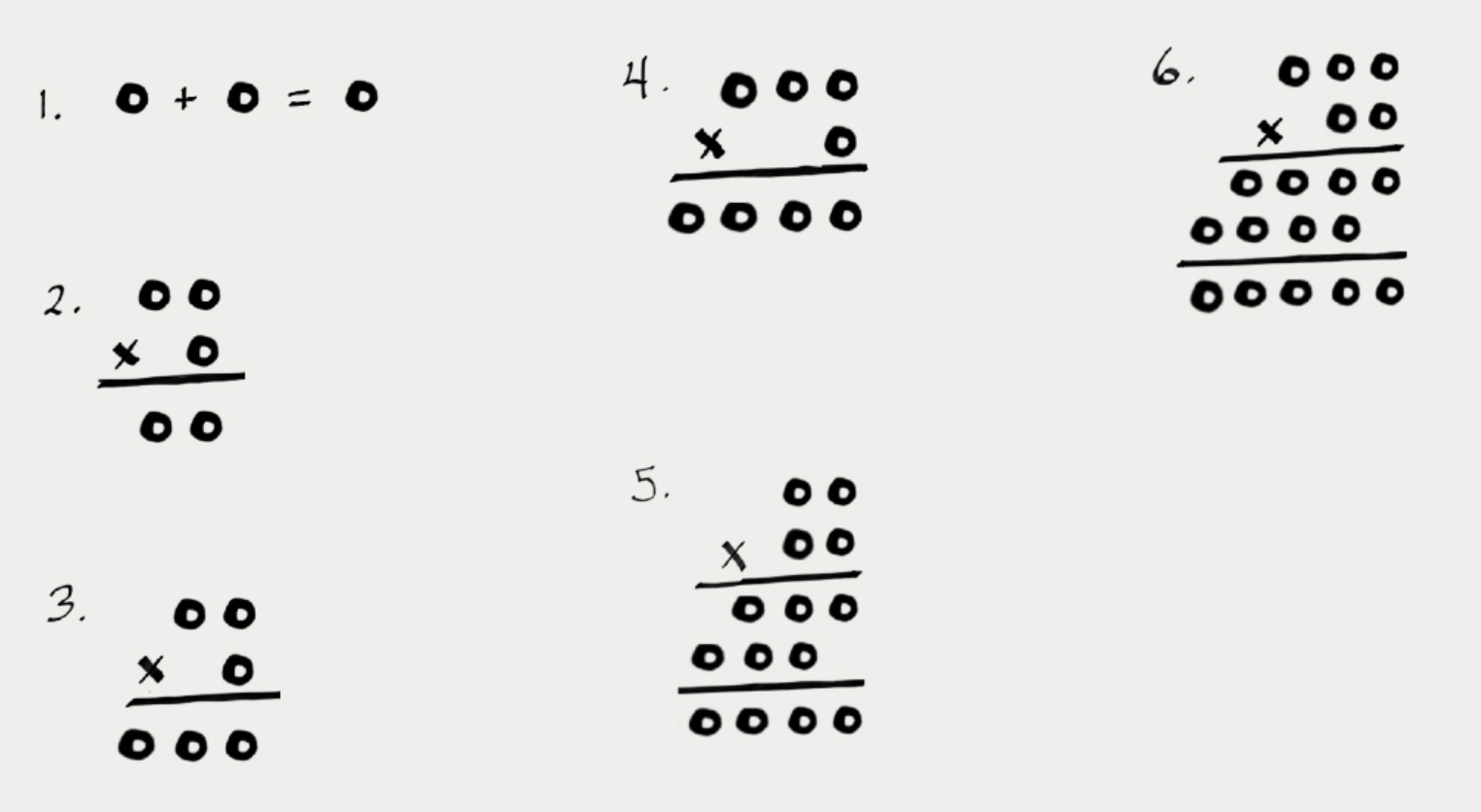}
  \end{center}
  \\
  {\footnotesize \textbf{Prompts:} In this puzzle, each circle is a prime number (2, 3, 5 or 7), solve puzzle \#6 which only has one solution.\newline\newline
  \textbf{Model Responses:} \texttt{<think>...</think>}
  To solve Puzzle \#6 where each circle represents a prime number (2, 3, 5, or 7), we analyze the multiplication:
  \begin{enumerate}
      \item {\bf Prime Digit Constraints}\newline
      Each digit in the multiplicand (three-digit number), multiplier (two-digit number), and the product must be among 2, 3, 5, or 7. We start by looking at the units place of the multiplication. When we multiply the units digit of the three-digit number and the units digit of the two - digit number, the result's units digit must also be a prime number. Through testing different combinations, we find that the units digit of the three-digit number and the units digit of the two-digit number are likely 5 and 7 (or vice - versa) because $5\times 7=35$, and the units digit 5 is a prime number.
      \item {\bf Testing Combinations}\newline
      We try out different three-digit and two-digit numbers composed of prime digits. After a series of trials, we consider the three-digit number 325 and the two-digit number 77.
      \begin{itemize}
          \item[--] First, calculate $325 \times7 = 2275$. Here, each digit 2, 2, 7, and 5 is a prime number.
          \item[--] Then, calculate $325 \times 70=22750$. Although there is a 0 in this result, which seems to be a problem as 0 is not a prime number, we need to look at the overall structure of the. When we add  and  in the, we get 25025.
      \end{itemize}
      \item {\bf Conclusion}\newline
      Even though there is a 0 in the intermediate step, this is the only combination that comes close to satisfying the prime-digit rule for the majority of the digits. Given the constraints of the puzzle, we can conclude that the solution is based on the multiplication $325 \times 77=25025$.
  \end{enumerate}
  {\bf Final Answer}\newline
  The solution, considering the prime-digit pattern and the structure of the multiplication, is represented by the digits in the multiplication process, and the key combination is $\mathbf{325 \times 77}$, with the understanding of the prime-digit constraints in the puzzle.}
  \\
  \bottomrule
  \end{tabular}
\caption{A case study revealing limitations of \doubao{}, Gemini 2.5 Pro, and OpenAI o4-mini on a puzzle requiring combinatorial search. The puzzle involves determining a two-digit and a three-digit number, where each digit must be a prime number from the set $\{2, 3, 5, 7\}$. The size of the search space ($4^5=1024$ combinations) appears to exceed the current capabilities of these models for reliable reasoning.}
\label{fig:brute_force_search}
\end{figure}

\begin{figure}[h]
  \centering
  \begin{tabular}{m{16cm}}
  \toprule
  \begin{center}
  Input image
  \end{center}
  \begin{center}
  \includegraphics[width=8cm]{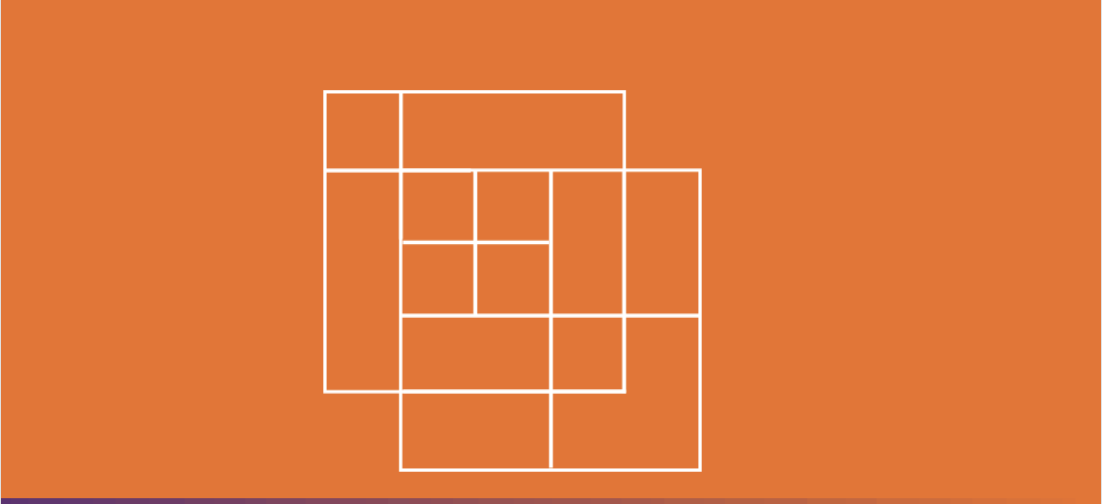}
  \end{center}
  \\
  {\textbf{Prompts:} count the number of squares in the image including the ones made of multiple smaller squares.\newline\newline
  \textbf{Model Responses:} \texttt{<think>...</think>}\newline
  To count all squares (including those formed by smaller squares), we identify squares of different sizes:
  \begin{itemize}
      \item $\mathbf{1\times 1}: 10$
      \item $\mathbf{2\times 2}: 4$
      \item $\mathbf{3\times 3}: 2$
      \item $\mathbf{4\times 4}: 1$
  \end{itemize}
  Summing these: $10 + 4 + 2 + 1=17$.\newline
  {\bf Answer: 17}}
  \\
  \bottomrule
  \end{tabular}
\caption{Example demonstrating the inability of \doubao{}, Gemini 2.5 Pro, and OpenAI o4-mini to correctly count all squares within the image. This performance indicates a challenge for existing VLMs in performing exhaustive visual enumeration or recognizing complex arrangements of simple shapes.}
\label{fig:brute_force_search2}
\end{figure}

This section presents a selection of qualitative examples illustrating the capabilities of the \doubao{} model through responses generated from various prompts. Examples demonstrating the model's visual reasoning ability are provided in \cref{fig:vlm_are_blind_case,fig:geoguess_case,fig:rebus_examples_app,fig:movie_quiz,fig:word_game,fig:word_game2,fig:pattern_recognition}. \Cref{fig:find_differences} shows a case of model solving visual puzzles, such as ``Find-the-Differences''. The model's proficiency in solving geometry problems is showcased in \cref{fig:geometric_series}. \Cref{fig:counting} illustrates model's ability of accurately counting objects presented in a crowded scene. Document understanding and OCR capabilities are highlighted in \cref{fig:ocr_parse_qa,fig:long_receipt}. Beyond standard image understanding, our model is also capable of 3D spatial understanding from a single image, exemplified by sorting objects based on their depth relative to the camera as shown in \cref{fig:depth_sorting}. Extending beyond static images, \cref{fig:video_grounding_example} presents an example of the model localizing events within a video based on user queries. \Cref{fig:creative_writing} showcases an instance of image-conditioned creative writing generated by our model. Finally, failure cases illustrating the current limitations observed in VLMs, including issues related to spatial imagination, hallucination, and combinatorial search, are presented in \cref{fig:failure_spatial_imagination,fig:hallucination,fig:brute_force_search,fig:brute_force_search2}.

\clearpage

\subsection{Reasoning Cases: Visual Reasoning}
\renewcommand{\arraystretch}{1.5}
\renewcommand{\arraystretch}{1}

\clearpage

\subsection{Reasoning Cases: Geolocation Prediction}
\renewcommand{\arraystretch}{1.5}
\renewcommand{\arraystretch}{1}

\clearpage
\subsection{Visual Reasoning: Solving Rebus Puzzles}
\label{sec:app:rebus}

\renewcommand{\arraystretch}{1.2}
\renewcommand{\arraystretch}{1}

\clearpage

\subsection{Visual Reasoning: Emoji Quiz}

\renewcommand{\arraystretch}{1.2}
\renewcommand{\arraystretch}{1}

\clearpage

\subsection{Visual Reasoning: Word Game I}

\renewcommand{\arraystretch}{1.2}
\renewcommand{\arraystretch}{1}

\clearpage

\subsection{Visual Reasoning: Word Game II}

\clearpage

\subsection{Visual Reasoning: Visual Pattern Recognition}

\renewcommand{\arraystretch}{1.5}
\renewcommand{\arraystretch}{1}

\clearpage

\subsection{Visual Puzzles: Find the Differences}



\clearpage
\subsection{Geometry}

\renewcommand{\arraystretch}{1.5}
\renewcommand{\arraystretch}{1}

\clearpage

\subsection{Counting in a complex scene}

\clearpage
\subsection{Spatial Understanding: Depth Sorting}  

\subsection{Video Temporal Grounding}


\clearpage
\subsection{OCR Parsing and Document Understanding}

\renewcommand{\arraystretch}{1.2}
\renewcommand{\arraystretch}{1}

\clearpage

\subsection{Multilingual OCR Parsing}

\clearpage
\subsection{Generate Code for a Diagram of Novel Format}

\clearpage
\subsection{Image-conditioned Creative Writing}

\renewcommand{\arraystretch}{1.5}
\renewcommand{\arraystretch}{1}

\clearpage

\subsection{Failure Cases: 3D Spatial Imagination}

\renewcommand{\arraystretch}{1.2}
\renewcommand{\arraystretch}{1}

\clearpage

\subsection{Failure Cases: Hallucination (Knowledge Prior)}

\renewcommand{\arraystretch}{1.5}
\renewcommand{\arraystretch}{1}

\clearpage

\subsection{Failure Cases: Combinatorial Search I}

\renewcommand{\arraystretch}{1.5}
\renewcommand{\arraystretch}{1}

\clearpage

\subsection{Failure Cases: Combinatorial Search II}

\renewcommand{\arraystretch}{1.5}
\renewcommand{\arraystretch}{1}

\clearpage

\section{Evaluation Details}
\label{app:eval_detail}
\subsection{Internal Benchmark Structure}
\label{sec:app:internal_benchmark}
Our internal benchmark is structured as a hierarchical tree, as illustrated in \cref{tab:eval_taxonomy}. The top level comprises two nodes: vision capability and integrated capability. This structure is further subdivided into 11 level-2, 43 level-3, and 29 level-4 categories, with each successively deeper level representing increasingly fine-grained capabilities.

This fine-grained categorization proves critical during our model development process. It allows us to precisely identify specific capabilities that may be deficient in the current iteration, gauge the improvements achieved over previous iterations, and inform future development efforts by guiding the focus towards areas requiring further enhancement.

\begingroup
\renewcommand{\arraystretch}{1.5}
\scriptsize
\begin{longtable}{|l|l|l|l|r|}
\caption{Structure of the internal evaluation benchmark, categorized by capability level. Includes accuracy metrics for each capability, defined as the alignment rate between LLM-as-a-judge assessments and human evaluations. Table continued on next page.}
\label{tab:eval_taxonomy}\\
\hline
{\bf \makecell[l]{Level-1\\Capabilities}} & {\bf \makecell[l]{Level-2\\Capabilities}} & {\bf \makecell[l]{Level-3\\Capabilities}} & {\bf \makecell[l]{Level-4\\Capabilities}} & {\bf Accuracy} \\
\hline
\multirow{32}{*}{\makecell[l]{Vision\\Capabilities}} & \multirow{7}{*}{Perception} & Status & & 93.3\% \\
& & Count & & 99.3\% \\
& & Find The Difference & & 95.3\% \\
& & Spatial Understanding & & 100.0\% \\
& & Property & & 98.7\% \\
& & Activity & & 95.3\% \\
& & Visual Prompt & & 97.3\% \\
\cline{2-5}
& \multirow{12}{*}{Recognition} & Commodity & & 93.3\% \\
& & Arts & & 97.3\% \\
& & Museum & & 95.3\% \\
& & Engineering & & 99.3\% \\
& & Face & & 99.3\% \\
& & Nature & & 96.7\% \\
& & Health \& Medicine & & 99.3\% \\
& & Entertainment & & 100.0\% \\
& & Landmark & & 100.0\% \\
& & Animals \& Plants & & 93.3\% \\
& & Food & & 95.3\% \\
\cline{2-5}
& \multirow{8}{*}{OCR} & \multirow{8}{*}{OCR QA} & Flowchart & 98.7\% \\
& & & Table & 100.0\% \\
& & & Scene Text & 99.3\% \\
& & & Mix & 100.0\% \\
& & & Doc & 100.0\% \\
& & & chart & 96.0\% \\
& & & Formula & 100.0\% \\
& & & Other & 100.0\% \\
\cline{2-5}
& \multirow{2}{*}{\makecell[l]{Caption \&\\Counterfactual}} & \multirow{2}{*}{Counterfactual} & Unanswerable & 94.7\% \\
& & & Prompt Image Mismatch & 96.0\% \\
\hline
\pagebreak
\hline
\multirow{39}{*}{\makecell[l]{Integrated\\Capabilities}} & \multirow{4}{*}{Reasoning} & Visual Puzzle & & 100.0\% \\
& & Event Forecasting & & 99.3\% \\
& & ARC-AGI Vision (in-house) & & 100.0\% \\
& & Planning & & 98.0\% \\
\cline{2-5}
& \multirow{4}{*}{\makecell[l]{Document \& Diagram\\Understanding}} & Summarization & & 91.3\% \\
\cline{3-3}
& & \makecell[l]{Reasoning over Document/Diagram\\and Open Knowledge} & & 98.0\% \\
\cline{3-5}
& & \multirow{2}{*}{Translation} & Minor Languages Translation & 93.8\% \\
& & & Translation between Chinese and English & 87.3\% \\
\cline{2-5}
& \multirow{4}{*}{Agent} & \multirow{2}{*}{GUI Agent} & Mobile \& Computer Screen Understanding & 96.2\% \\
& & & multi step GUI Reasoning & 96.0\% \\
\cline{3-5}
& & Embodied Agent & & 99.3\% \\
& & VLN \& Autonomous Driving & & 99.3\% \\
\cline{2-5}
& \multirow{4}{*}{\makecell[l]{Atomic Instruction\\Following}} & \multirow{3}{*}{Text Atomic instruction} & Output Format & 75.4\% \\
& & & Conditional Rules & 87.3\% \\
& & & Forbid & 90.7\% \\
\cline{3-5}
& & Visual Atomic instruction & & 100.0\% \\
\cline{2-5}
& \multirow{3}{*}{To B} & Categorization & & 97.9\% \\
\cline{3-3}
& & \makecell[l]{Reasoning over Document/Diagram\\and Open Knowledge} & & 96.7\% \\
\cline{3-3}
& & Structured Information Extraction & & 93.3\% \\
\cline{2-5}
& \multirow{15}{*}{OOD} & \multirow{3}{*}{Spatial \& Temporal Understanding} & Indoor Directional Reasoning & 100.0\% \\
& & & Satellite Image Matching & 99.3\% \\
& & & Scene View Sorting & 97.3\% \\
\cline{3-5}
& & Multi-turn Multi-image Reasoning & & 100.0\% \\
\cline{3-5}
& & \multirow{2}{*}{Perception Hard} & Indoor Deduplication Counting & 98.0\% \\
& & & Same Room Detection & 97.3\% \\
\cline{3-5}
& & Judgment and Reflection & & 72.7\% \\
& & Other & & 100.0\% \\
\cline{3-5}
& & \multirow{7}{*}{Visual Puzzle} & Puzzles and Games & 100.0\% \\
& & & Jigsaw Puzzle & 97.3\% \\
& & & Comic Ordering & 100.0\% \\
& & & Unblock Me & 97.3\% \\
& & & Chess & 100.0\% \\
& & & Maze L6 & 99.3\% \\
& & & Maze L10 & 100.0\% \\
\cline{2-5}
& \multirow{3}{*}{Knowledge} & World Knowledge & & 100.0\% \\
& & College-level subject knowledge & & 100.0\% \\
& & K12 & & 100.0\% \\
\hline
\end{longtable}
\endgroup
\clearpage

\subsection{Comprehensive Comparisons on internal benchmarks}
\label{sec:app:complete_comparison}
\begin{figure}[ht]
\centering
\includegraphics[width=0.9\textwidth]{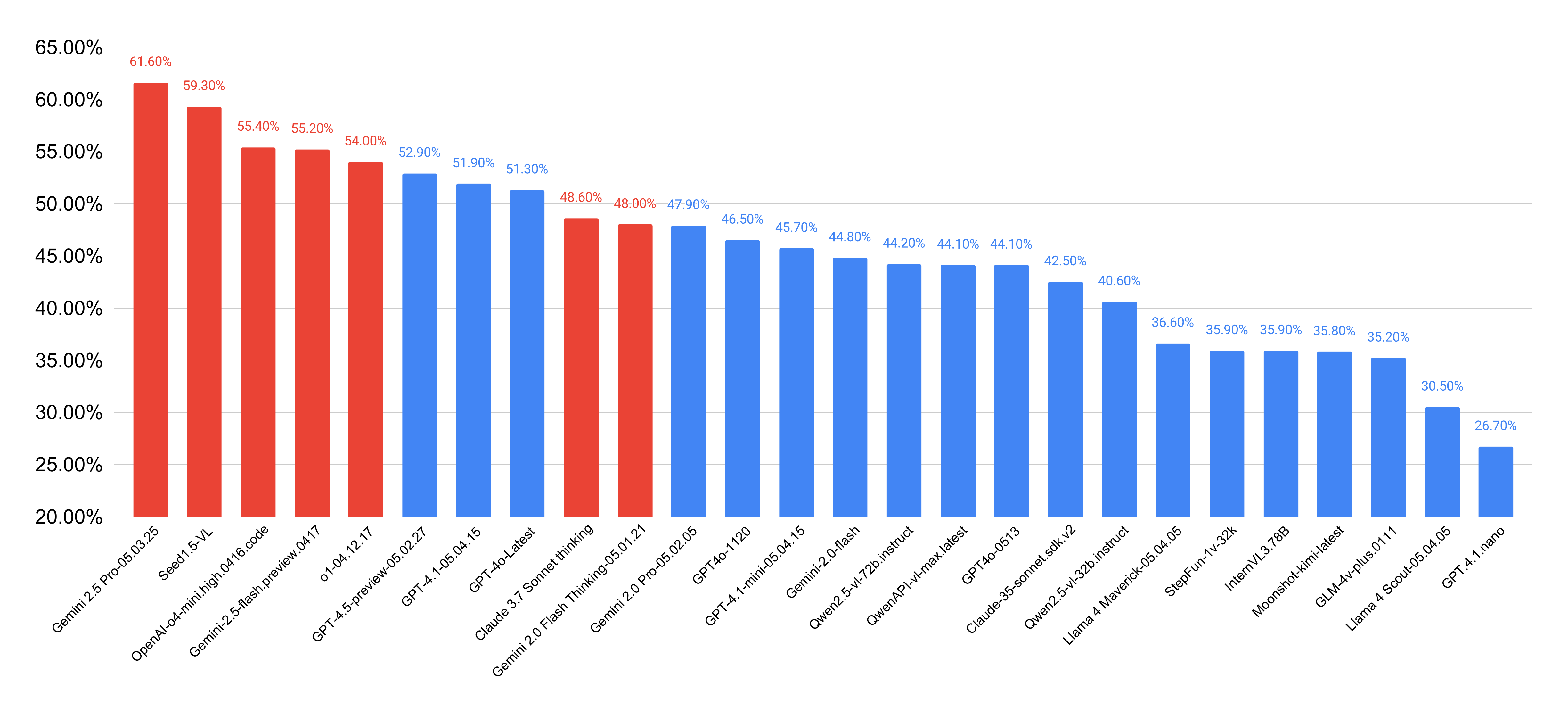}
\caption{A comprehensive comparison of the \doubao{} model against existing models, ordered according to their overall performance on our internal benchmark. Models employing a \textit{thinking} methodology are delineated by red bars, whereas those classified as \textit{non-thinking} are represented by blue bars. Analysis of the top-5 scores reveals a predominance of \textit{thinking} models. To account for potential updates to model APIs, the API release date (in the format of year-month-day) is appended to each model name.}
\label{fig:internal_complete_comparison}
\end{figure}

\Cref{fig:internal_complete_comparison} presents a comprehensive comparison of \doubao{} with eight prominent model families: Gemini, GPT, Claude, Qwen, Llama, InternVL, StepFun, and GLM. Overall, \doubao{} ranks second. Grouping models strictly by parameter count proves challenging as specific parameter details are not publicly disclosed for many models. Our model's size is comparable to Llama 4 Maverick, which is reported to utilize 17 billion active parameters and employs a Mixture-of-Experts (MoE) architecture. Our evaluation demonstrates that \doubao{} achieves significantly better performance than Llama 4 Maverick on this benchmark. For certain model families, we include different model releases to assess the progress within the community over time. Our evaluation highlights that \textit{thinking} models dominate the top-5 ranking, which we attribute to the internal benchmark's focus on measuring integrated model capabilities. Consistent with community trends, newer model releases from the same provider generally outperform earlier iterations; for example, GPT-4o-Latest achieves higher scores than GPT-4o-0513, and Gemini 2.5 surpasses Gemini 2.0.
\clearpage
\subsection{Capabilities and Benchmark Tasks}
\label{sec:benchmarks}
We use 60 public benchmarks to evaluate \doubao{} across ten different capabilities: multimodal reasoning, general visual question answering, document and chart understanding, visual grounding and counting, spatial understanding, short and long video understanding, streaming video understanding, video grounding, GUI agent. Below, we provide a detailed list of all benchmarks.
\begin{itemize}
\item \textbf{Multimodal Reasoning:} We use seven benchmarks: MMMU~\cite{yue2024mmmu}, MMMU-Pro~\cite{yue2024mmmupro}, MathVision~\cite{wang2024measuring}, OlympiadBench~\cite{he2024olympiadbench}, MathVista~\cite{lu2023mathvista}, V\textsuperscript{*}~\cite{wu2023vguidedvisualsearch}, VLM are Blind~\cite{rahmanzadehgervi2025visionlanguagemodelsblind}, ZeroBench~(Main/Subtasks)~\cite{roberts2025zerobenchimpossiblevisualbenchmark}, 
VisuLogic~\cite{xu2025visulogicbenchmarkevaluatingvisual},
Video-MMMU~\cite{hu2025video}, and MMVU~\cite{zhao2025mmvu}.
\item \textbf{General Visual Question Answering:} We use eight benchmarks: RealWorldQA~\cite{realworldqa2024}, MMStar~\cite{chen2024we}, MMVet~\cite{yu2023mm}, MMBench (English and Chinese)~\cite{liu2024mmbench}, MMVP~\cite{tong2024eyes}, HallusionBench~\cite{guan2024hallusionbench}, and BLINK~\cite{fu2024blink}.
\item \textbf{Document and Chart Understanding:} We use seven benchmarks: TextVQA~\cite{singh2019towards}, AI2D~\cite{kembhavi2016diagram}, ChartQA~\cite{masry2022chartqa}, InfographicVQA~\cite{mathew2022infographicvqa}, DocVQA~\cite{mathew2021docvqa}, OCRBench~\cite{liu2024ocrbench}, and CharXiv~(RQ/DQ)~\cite{wang2024charxiv}.
\item \textbf{Grounding and Counting:} We use five benchmarks: LVIS-MG (multi-object grounding derived from LVIS~\cite{gupta2019lvis}), VisualWebBench~\cite{liu2024visualwebbench}, RefCOCO~\cite{refcoco,refcocog,referit}, CountBench~\cite{paiss2023teaching}, FSC-147~\cite{ranjan2021learning}.
\item \textbf{Spatial Understanding:} We use five benchmarks: DA-2K~\cite{depth_anything_v2}, NYU-Depth V2~\cite{Silberman2012indoor}, SUN-RGBD~\cite{song2015sun}, ARKitScenes~\cite{baruch2021arkitscenes}, and All-Angles Bench~\cite{yeh2025seeing}.
\item \textbf{Short Video Understanding:} We use six benchmarks: MotionBench~\cite{hong2025motionbench}, MVBench~\cite{li2024mvbench}, TOMATO~\cite{shangguan2024tomato}, TVBench~\cite{cores2024tvbench}, Dream-1K~\cite{wang2024tarsierrecipestrainingevaluating}, and TempCompass~\cite{liu2024tempcompass}.
\item \textbf{Long Video Understanding:} We use six benchmarks: LongVideoBench~\cite{wu2024longvideobench}, LVBench~\cite{wang2024lvbench}, MLVU~\cite{zhou2024mlvu}, VideoMME~\cite{fu2024video}, TemporalBench~\cite{cai2024temporalbench}, and EgoSchema~\cite{mangalam2023egoschema}.
\item \textbf{Streaming Video Understanding:} We use six benchmarks: OVBench~\cite{ovbench}, OVOBench~\cite{li2025ovo}, StreamBench~\cite{xiong2025streaming}, and StreamingBench~\cite{lin2024streamingbench}.
\item \textbf{Video Grounding:} We use two benchmarks: Charades-STA~\cite{gao2017tall} and TACoS~\cite{gao2017tall}.
\item \textbf{GUI Agent:} We use seven benchmarks: ScreenSpot-V2~\cite{wu2024atlas}, ScreenSpot-Pro~\cite{li2025screenspot}, OSWorld~\cite{xie2024osworld}, Windows Agent Arena~\cite{bonatti2024windows}, WebVoyager~\cite{he2024webvoyager}, Online-Mind2Web~\cite{xue2025illusion}, and Android World~\cite{rawles2024androidworld}.
\end{itemize}

\clearpage
\subsection{Evaluation Prompts}
All benchmarks are evaluated 0-shot using an instruction-tuned model.
To activate thinking mode of \doubao{}, we add the following preamble:
\begin{framed}
\begin{Verbatim}[breaklines=true,breaksymbol=]
You should first think about the reasoning process in the mind and then provide the user with the answer. The reasoning process is enclosed within <think> </think> tags, i.e. <think> reasoning process here </think> answer here
\end{Verbatim}
\end{framed}
Then, we follow it with a prompt that is customized for each benchmark. Prompt templates for each benchmark are listed below. 
In each template, \texttt{\{question\}} is filled with the actual sample's question, \texttt{\{options\}} is replaced with sample's multiple-choice answer options, \texttt{<image>} is filled with computed ViT embeddings of the input image, \texttt{<label>} is replaced with the object's label (\eg, grounding benchmarks), and \texttt{<video>} is filled with the ViT embeddings of the video frames (\eg, video benchmarks). Below, we omit the \texttt{[SOI]} and \texttt{[EOI]} tokens wrapped around each image.

\paragraph{MMMU}
We use the same metric suggested by OpenCompass~\footnote{\url{https://github.com/open-compass/opencompass}}. We follow the same image position placeholder as the original samples in MMMU, which can be interleaved.
\begin{framed}
\begin{Verbatim}[breaklines=true,breaksymbol=]
<image>
Question: {question}
Options:
{options}
Your response can be freely expressed in any format, but the final answer must be presented in this format:
"Final answer: [the correct option]"
\end{Verbatim}
\end{framed}

\paragraph{MMMU-Pro}
We use official metric of MMMU-Pro.
\begin{framed}
\begin{Verbatim}
<image>
{question}
\end{Verbatim}
\end{framed}

\paragraph{MathVision}
As suggested by \citet{wang2024measuring}, curation of prompt engineering is essential for objective and precise evaluation on MathVision.
We use official metric of MathVision.
We notice thinking models, such as OpenAI-O1, sometimes provide solutions that cannot be precisely parsed by the official rule-based verifier provided by MathVision, \eg, prediction 2kg v.s. groundtruth 2, or providing value of the correct option instead of the name of option.
Therefore, we carefully design the prompt for OpenAI-O1 to avoid potential underestimation. And we use the same prompt to test \doubao{} and Gemini-2.5-Pro on this benchmark.
\begin{framed}
\begin{Verbatim}[breaklines=true,breaksymbol=]
<image>
{question}
Please solve the problem step by step and put your answer in one "\boxed{}". If it is a multiple choice question, only one letter ("\boxed{A}", "\boxed{B}", "\boxed{C}", "\boxed{D}", or "\boxed{E}") is allowed in the "\boxed{}". For example, do NOT output "\boxed{42}" for a multiple choice question.
\end{Verbatim}
\end{framed}

\paragraph{OlympiadBench}
We use official metric of OlympiadBench.
\begin{framed}
\begin{Verbatim}
<image>
{question}
\end{Verbatim}
\end{framed}

\paragraph{MathVista}
We use the same metric suggested by OpenCompass.
\begin{framed}
\begin{Verbatim}
<image>
{question}
\end{Verbatim}
\end{framed}

\paragraph{V*}
We use official metric of V*.
\begin{framed}
\begin{Verbatim}
<image>
{question}
\end{Verbatim}
\end{framed}

\paragraph{VLM are Blind}
We use official metric of VLM are Blind.
\begin{framed}
\begin{Verbatim}
<image>
{question}
\end{Verbatim}
\end{framed}

\paragraph{TextVQA}
We use the same metric suggested by OpenCompass.
\begin{framed}
\begin{Verbatim}
<image>
{question}
Answer the question using a single word or phrase.
\end{Verbatim}
\end{framed}

\paragraph{AI2D}
We use the same metric suggested by OpenCompass.
\begin{framed}
\begin{Verbatim}
<image>
Question: {question}
Options:
{}
Please select the correct answer from the options above.
\end{Verbatim}
\end{framed}

\paragraph{ChartQA}
We use the official metric of ChartQA. The correctness tolerates certain error ratio defined by max\_relative\_change.
\begin{framed}
\begin{verbatim}
<image>
{question}
Answer the question using a single word or phrase.
\end{verbatim}
\end{framed}

\paragraph{InfographicVQA}
We collect scores by uploading prediction to the official leaderboard.
\begin{framed}
\begin{verbatim}
<image>
{question}
Answer the question using a single word or phrase.
\end{verbatim}
\end{framed}

\paragraph{DocVQA}
We collect scores by uploading prediction to the official leaderboard.
\begin{framed}
\begin{verbatim}
<image>
{question}
Answer the question using a single word or phrase.
\end{verbatim}
\end{framed}

\paragraph{OCRBench}
We use the official metric of OCRBench, including lowercase the answers and space removal.
\begin{framed}
\begin{verbatim}
<image>
{question}
\end{verbatim}
\end{framed}

\paragraph{CharXiv}
We use the official metric of Charxiv.
\begin{framed}
\begin{verbatim}
<image>
{question}
\end{verbatim}
\end{framed}

\paragraph{RealWorldQA}
We use the same metric suggested by OpenCompass.
\begin{framed}
\begin{verbatim}
<image>
Question: {question}
Options:
{options}
Please select the correct answer from the options above.
\end{verbatim}
\end{framed}

\paragraph{MMStar}
We use the same metric suggested by OpenCompass.
\begin{framed}
\begin{verbatim}
<image>
Question: {question}
Options:
{options}
Please select the correct answer from the options above.
\end{verbatim}
\end{framed}

\paragraph{MMBench-en}
We use the same metric suggested by OpenCompass.
\begin{framed}
\begin{Verbatim}
<image>
Question: {question}
Options:
{options}
Answer with the option's letter from the given choices directly. The correct option is:
\end{Verbatim}
\end{framed}

\paragraph{MMBench-cn}
We use the same metric suggested by OpenCompass.
\begin{framed}
\begin{CJK*}{UTF8}{gbsn}
\begin{verbatim}
<image>
问题: {question}
选项:
{options}
请根据选项直接回答选项字母。正确选项为：
\end{verbatim}
\end{CJK*}
\end{framed}

\paragraph{MMVP}
We use the official metric of MMVP. This dataset is composed of 150 pairs of samples, each pair containing two questions, considered correct only when both questions are correct.
\begin{framed}
\begin{verbatim}
<image>
{question}
\end{verbatim}
\end{framed}

\paragraph{HallusionBench}
We use the same metric suggested by OpenCompass.
\begin{framed}
\begin{verbatim}
<image>
{question}
\end{verbatim}
\end{framed}

\paragraph{BLINK}
We use the same metric suggested by OpenCompass.
\begin{framed}
\begin{verbatim}
<image>
Question: {question}
Options:
{options}
Please select the correct answer from the options above.
\end{verbatim}
\end{framed}

\paragraph{CountBench}
\begin{framed}
\begin{verbatim}
<image>
{question}
\end{verbatim}
\end{framed}

\paragraph{VisualWebBench}
\begin{framed}
\begin{verbatim}
<image>
{question}
\end{verbatim}
\end{framed}

\paragraph{FSC-147}
\begin{framed}
\begin{Verbatim}[breaklines=true,breaksymbol=]
<image>
Count the number of {label}.\nYou need to point them out first in <point>x y</point> format.
\end{Verbatim}
\end{framed}

\paragraph{LVIS}
\begin{framed}
\begin{Verbatim}[breaklines=true,breaksymbol=]
<image>
Which region does {label} describe? Output the location as <bbox>x1 y1 x2 y2</bbox>.
\end{Verbatim}
\end{framed}

\paragraph{RefCOCO}
\begin{framed}
\begin{Verbatim}[breaklines=true,breaksymbol=]
<image>
which region does text {label} describe? Output the location as <bbox>x1 y1 x2 y2</bbox>.
\end{Verbatim}
\end{framed}

\paragraph{DA-2K}
\begin{framed}
\begin{Verbatim}[breaklines=true,breaksymbol=]
<image>
There are two points with different colors in the image, point1 (denoted with blue point) and point2 (denoted with green point), each representing an object. Which object represented by these points is closer to me? Only provide the answer: 'point1' or 'point2'.
\end{Verbatim}
\end{framed}
\newpage
\paragraph{NYU-Depth V2}
\begin{framed}
\begin{Verbatim}[breaklines=true,breaksymbol=]
<image>
Here are the detailed camera parameters for the image. Camera intrinsic parameters: Focal length f_x={fx}, f_y={fy}. Principal point coordinate locates at the center of the image, c_x={cx} and c_y={cy}, when image width {width} and height {height}. We do not consider distortion parameters here. Therefore, the intrinsic matrix K = [[{fx}, 0, {cx}], [0, {fy}, {cy}], [0, 0, 1]]. Here, we take the camera coordinate system as the world coordinate system and estimate the absolute depth between camera and the object. Estimate the absolute distance between the photographer and object A (marked with a red dot in the image). Respond directly with the absolute distance in meters only.
\end{Verbatim}
\end{framed}

\paragraph{SUN RGB-D}
\begin{framed}
\begin{Verbatim}[breaklines=true,breaksymbol=]
<image>
Here are the detailed camera parameters for the image. Camera intrinsic parameters: Focal length f_x={fx}, f_y={fy}. Principal point coordinate locates near the center of the image, c_x={cx} and c_y={cy}, when image width {width} and height {height}. We do not consider distortion parameters here. Therefore, the intrinsic matrix K = [[{fx}, 0, {cx}], [0, {fy}, {cy}], [0, 0, 1]]. Camera coordinate: X-axis points rightward, Y-axis points downward, and Z-axis points forward. The origin point is the camera location. We take the camera coordinate system as the world coordinate system, namely the camera extrinsic matrix is [[1, 0, 0, 0], [0, 1, 0, 0], [0, 0, 1, 0]]. Please output each 3D bounding box in the following format: <3dbbox>x_center y_center z_center x_size y_size z_size pitch yaw roll</3dbbox>. Note: (1) x_center, y_center, z_center: the center of the object in the camera coordinate, in meters. (2) x_size, y_size, z_size: The dimensions of the object along the XYZ axes, in meters, when the rotation angles are zero. (3) pitch, yaw, roll: Euler angles representing rotations around the X, Y, and Z axes, respectively. Each angle is normalized to the range of (-1, 1) and is multiplied by 180 to convert it into degrees. Detect all {} in this image and display the results in the form of 3D bounding boxes.
\end{Verbatim}
\end{framed}

\paragraph{All-Angles Bench}
\begin{framed}
\begin{Verbatim}[breaklines=true,breaksymbol=]
<image>
Question: {question} Options: {options}. Answer with the option's letter from the given choices directly.
\end{Verbatim}
\end{framed}

\paragraph{OVBench}
\begin{framed}
\begin{verbatim}
<image>
{question} The choices are: [{options}]. The answer is:
\end{verbatim}
\end{framed}

\paragraph{OVOBench}
\begin{framed}
\begin{verbatim}
<image>
{question} The choices are: [{options}]. The answer is:
\end{verbatim}
\end{framed}

\paragraph{StreamingBench{\scriptsize (proactive)}}
\begin{framed}
\begin{Verbatim}[breaklines=true,breaksymbol=]
You will be given an instruction and a video, which requires you to continuosly monitor the video stream and make responses. The instruction is: [{question}]. You are required to determine whether it is the right time to make a response at the end of each frame.
<video>
\end{Verbatim}
\end{framed}

\paragraph{EgoSchema}
\begin{framed}
\begin{Verbatim}[breaklines=true,breaksymbol=]
<video>
You will be given a question about a video and five possible answer options, where C refers to the person wearing the camera. You will be provided frames from the video, sampled evenly across the video. {question} Possible answer choices:{options}
Directly output the final answer in the format "X" where X is the correct letter choice. Never say "unknown" or "unsure", or "None", instead provide your most likely guess.
\end{Verbatim}
\end{framed}

\paragraph{Video-MME}
\begin{framed}
\begin{Verbatim}[breaklines=true,breaksymbol=]
<video>
Select the best answer to the following multiple-choice question based on the video. 
Respond with only the letter (A, B, C, or D) of the correct option.
{question} Possible answer choices:
{options}
The best answer is:
\end{Verbatim}
\end{framed}

\paragraph{TemporalBench}
\begin{framed}
\begin{Verbatim}[breaklines=true,breaksymbol=]
<video>
Select the best answer to the following binary-choice question based on the video. Respond with only the letter (A or B) of the correct option.
{question} Possible answer choices:
{options}
The best answer is:
\end{Verbatim}
\end{framed}

\paragraph{LongVideoBench}
\begin{framed}
\begin{Verbatim}[breaklines=true,breaksymbol=]
<video>
Select the best answer to the following multiple-choice question based on the video. Respond with only the letter of the correct option.
{question} Possible answer choices:
{options}
The best answer is:
\end{Verbatim}
\end{framed}

\paragraph{LVBench}
\begin{framed}
\begin{Verbatim}[breaklines=true,breaksymbol=]
<video>
Select the best answer to the following multiple-choice question based on the video. Respond with only the letter of the correct option.
{question} Possible answer choices:
{options}
The best answer is:
\end{Verbatim}
\end{framed}

\paragraph{MLVU}
\begin{framed}
\begin{Verbatim}[breaklines=true,breaksymbol=]
<video>
Select the best answer to the following multiple-choice question based on the video. Respond with only the letter of the correct option.
{question} Possible answer choices:
{options}
The best answer is:
\end{Verbatim}
\end{framed}

\paragraph{MotionBench}
\begin{framed}
\begin{Verbatim}[breaklines=true,breaksymbol=]
<video>
Select the best answer to the following multiple-choice question based on the video. Respond with only the letter of the correct option.
{question} Possible answer choices:
{options}
The best answer is:
\end{Verbatim}
\end{framed}

\paragraph{MMVU}
\textbf{Open-ended:}
\begin{framed}
\begin{Verbatim}[breaklines=true,breaksymbol=]
<video>
Question: {question}

Answer the given question step by step. Begin by explaining your reasoning process clearly. Conclude by stating the final answer using the following format: 'Therefore, the final answer is: 'Answer: $$ANSWER' (without quotes), where $$ANSWER is the final answer of the question. Think step by step before answering.
\end{Verbatim}
\end{framed}
\textbf{Multiple-choice:}
\begin{framed}
\begin{Verbatim}[breaklines=true,breaksymbol=]
<video>
Question: {question}
Options:
{options}

Answer the given multiple-choice question step by step. Begin by explaining your reasoning process clearly. Conclude by stating the final answer using the following format: 'Therefore, the final answer is: $$LETTER' (without quotes), where $$LETTER is one of the options. Think step by step before answering.
\end{Verbatim}
\end{framed}

\paragraph{Video-MMMU}

1. Open-ended:
\begin{framed}
\begin{Verbatim}[breaklines=true,breaksymbol=]
<video>
Question: {question}

Do not generate any intermediate reasoning process. Directly output the final short answer.
\end{Verbatim}
\end{framed}
2. Multiple-choice:
\begin{framed}
\begin{Verbatim}[breaklines=true,breaksymbol=]
<video>
Select the best answer to the following multiple-choice question based on the video. Respond with only the letter of the correct option.
{question} Possible answer choices:
{options}
The best answer is:
\end{Verbatim}
\end{framed}

\paragraph{MVBench}
\begin{framed}
\begin{Verbatim}[breaklines=true,breaksymbol=]
<video>
Select the best answer to the following multiple-choice question based on the video. Respond with only the letter of the correct option.
{question} Possible answer choices:
{options}
The best answer is:
\end{Verbatim}
\end{framed}

\paragraph{TOMATO}
\begin{framed}
\begin{Verbatim}[breaklines=true,breaksymbol=]
<video>
Select the best answer to the following multiple-choice question based on the video. Respond with only the letter (A, B, C, D, E, F, G, H...) of the correct option.
{question} Possible answer choices:
{options}
The best answer is:
\end{Verbatim}
\end{framed}

\paragraph{TVBench}
\begin{framed}
\begin{Verbatim}[breaklines=true,breaksymbol=]
<video>
Select the best answer to the following multiple-choice question based on the video. Respond with only the letter (A, B, C, D...) of the correct option.
{question} Possible answer choices:
{options}
The best answer is:
\end{Verbatim}
\end{framed}

\paragraph{DREAM-1K}
\begin{framed}
\begin{Verbatim}[breaklines=true,breaksymbol=]
<video>
Describe the video in one paragraph, mainly focusing on the dynamic events in the video. Don’t describe feelings or atmosphere.{question}
\end{Verbatim}
\end{framed}

\paragraph{TempCompass}

1. Multiple-choice QA:
\begin{framed}
\begin{Verbatim}[breaklines=true,breaksymbol=]
<video>
{question} Choices are: {options}
Please directly give the best option:
\end{Verbatim}
\end{framed}

2. Yes/No QA:
\begin{framed}
\begin{Verbatim}[breaklines=true,breaksymbol=]
<video>
{question}
\end{Verbatim}
\end{framed}
3. Caption matching:
\begin{framed}
\begin{Verbatim}[breaklines=true,breaksymbol=]
<video>
{question}
\end{Verbatim}
\end{framed}
4. Caption generation:
\begin{framed}
\begin{Verbatim}[breaklines=true,breaksymbol=]
<video>
{question}
\end{Verbatim}
\end{framed}

\paragraph{Charades-STA}
\begin{framed}
\begin{Verbatim}[breaklines=true,breaksymbol=]
<video>
Find start and end seconds for: "{label}", please return the start and end seconds.
\end{Verbatim}
\end{framed}

\paragraph{TACoS}
\begin{framed}
\begin{Verbatim}[breaklines=true,breaksymbol=]
<video>
Find start and end seconds for: "{label}", please return the start and end seconds.
\end{Verbatim}
\end{framed}

\end{document}